\documentclass[onecolumn]{article}
\usepackage{bagrow}

\usepackage{fullpage}
\usepackage{amsmath,amssymb,amsfonts}
\usepackage{graphicx}
\usepackage{adjustbox} %
\usepackage{hyperref}
\usepackage[natbibapa]{apacite}
\usepackage{booktabs}
\usepackage{multirow}
\usepackage{dsfont}
\usepackage{enumitem}
\usepackage{threeparttable}

\usepackage{geometry}
\usepackage{caption}
\usepackage{enumitem}
\setlist[itemize]{itemsep=0pt}
\usepackage{pifont}

\usepackage{xcolor}  %
\newcommand{\methods}{Methods (App.~\ref{sec:methods})}

\newtheorem{definition}{Definition}
\newcommand{\C}{\mathcal{C}}
\newcommand{\T}{\mathcal{T}}
\newcommand{\R}{\mathbb{R}}
\newcommand{\E}{\mathbb{E}}

\newcommand{\cmark}{\textcolor{green!75!black}{\ding{51}}}%
\newcommand{\xmark}{\textcolor{red!90!black}{\ding{55}}}%

\newcommand{\ktot}{\ensuremath{k}}
\title{Softly Symbolifying Kolmogorov--Arnold Networks}

\author[1,2,*]{James Bagrow}
\author[3,2]{Josh Bongard}
\affil[1]{Mathematics \& Statistics, University of Vermont, Burlington, VT, United States }
\affil[2]{Vermont Complex Systems Center, University of Vermont, Burlington, VT, United States}
\affil[3]{Computer Science, University of Vermont, Burlington, VT, United States}
\affil[*]{\corrauthinfo{james.bagrow@uvm.edu}{bagrow.com}
}

\date{October 22, 2026}

\begin{document}

\maketitle

\begin{abstract}\small
Kolmogorov--Arnold Networks (KANs) offer a promising path toward interpretable machine learning: their learnable activations can be studied individually while collectively fitting complex data accurately. In practice, however, trained activations often lack \emph{symbolic fidelity}, learning pathological decompositions with no meaningful correspondence to interpretable forms. We propose Softly Symbolified Kolmogorov--Arnold Networks (S2KAN), which integrate symbolic primitives directly into training. Each activation draws from a dictionary of symbolic and dense terms, with learnable gates that sparsify the representation. Crucially, this sparsification is differentiable, enabling end-to-end optimization, and uses an optimization objective inspired by minimum description length. S2KAN can discover interpretable activation functions while retaining dense splines for flexible approximation. We introduce a quantitative measure of symbolic fidelity, assessing how closely learned activations can be approximated by short expressions from a specified symbolic language. Across real-world prediction tasks, dynamical-system forecasting, and synthetic symbolic benchmarks, S2KAN generally achieves higher symbolic fidelity and often far smaller models, while maintaining competitive predictive accuracy on real-world data and low prediction errors on most symbolic benchmark configurations. We also examine the models' learned responses on scientific data, including the contributions of their symbolic components. Together, these results show that incorporating symbolic structure during training can yield compact, accurate KAN models whose learned responses can be examined through their symbolic components.
\end{abstract}


\section{Introduction}

Neural networks and deep learning are powerful but opaque tools for modeling complex systems~\citep{10.1145/3233231,8466590,rudin2019stop,guidotti2018survey}. 
In scientific problems, accurate predictions are often not enough, and practitioners seek interpretable models that reveal underlying mechanisms, suggest novel hypotheses, or confirm existing theories~\citep{hempel1970aspects,woodward2004making,shmueli2010explain}.
Such interpretability can be supported by parsimony, seeking smaller or less parameterized models~\citep{schwarz1978estimating,hastie2015statistical,sindy2016,louizos2018learning}, or through the use of interpretable building blocks, components such as mathematical functions or symbolic expressions that can be interrogated and understood individually~\citep{schmidt2009distilling,udrescu2020ai,cranmer2020discovering,cranmer2023interpretable,cava2021contemporary}.
These avenues are not mutually exclusive, and in fact the most interpretable models often combine both: sparse combinations of symbolic primitives that are simultaneously compact and semantically meaningful.

Kolmogorov--Arnold Networks (KANs)~\citep{liu2025kan} have emerged as a promising alternative to traditional neural networks, replacing fixed activation functions with learnable univariate functions on each edge. 
Using flexible representations for the activation functions allows KANs to be highly performant~\citep{liu2025kan,PhysRevResearch.7.023037,Bagrow_2025}. 
But this design also lets practitioners examine the internal workings of the KAN by inspecting the learned activations, and possibly infer mathematical expressions (e.g., $\sin(x)$, $e^x$, $x^2$) for those activations~\citep{liu2025kan,liu2024kan2.0}, although interpreting many such functions becomes challenging in large KANs.
This combination of accuracy and interpretability makes KANs attractive for scientific machine learning~\citep{carleo2019machine,xu2021artificial,karniadakis2021physics}.

\begin{figure}
    \centering
    \includegraphics[width=\textwidth]{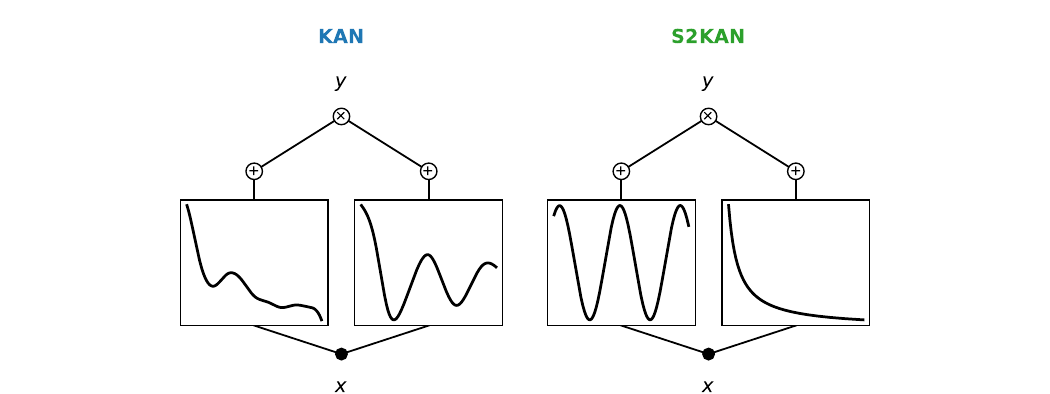}
    \caption{Learning $y = \mathrm{sinc}(x) = \sin(x)/x$ with a multiplicative Kolmogorov--Arnold network (KAN). 
    The standard KAN lacks symbolic fidelity, exhibiting a numerically accurate but otherwise pathological decomposition, whereas S2KAN captures the underlying sinc decomposition.
    The two models have empirical symbolic fidelities of 0\% and 100\%, respectively; see \methods{} for evaluation details.
    \label{fig:sincx}}
\end{figure}

However, the standard KAN reliance on flexible dense representations is at odds with interpretability.
Splines, the typical choice of representation, can fit arbitrary shapes, but the resulting activations may bear no resemblance to recognizable mathematical functions.
The standard remedy---post-hoc symbolification, where each trained activation is independently fitted to a symbolic form~\citep{liu2025kan}---is problematic: splines may learn shapes difficult to fit symbolically, each activation is converted without considering network-level effects, and symbolic forms are not explored during training.
The result: learned activations often lack \emph{symbolic fidelity} (SF): accurate predictions but no meaningful correspondence to interpretable expressions (Fig.~\ref{fig:sincx}).
What is needed is a method that treats symbolic and dense representations on equal footing, automatically selects among them while pruning unnecessary components, and does so continuously during training---yielding compact models with more symbolically expressible activation functions while retaining competitive accuracy.

We propose \textbf{Softly Symbolified KANs (S2KAN)}, which integrate symbolic primitives and sparse bases directly into the training process through a differentiable gating mechanism. 
Our key contributions are:
\begin{itemize}
    \item An activation function dictionary combining dense representations, sparse basis functions, and symbolic primitives;
    \item A differentiable training process that jointly fits, symbolifies, and sparsifies the network;
    \item An objective motivated by minimum description length (MDL), balancing predictive accuracy and model complexity;
    \item A formal definition and quantitative measure of symbolic fidelity, distinguishing predictive accuracy from the simplicity of learned activation functions.
\end{itemize}
Our method combines symbolic and dense representations: S2KAN can select sparse symbolic forms when they suffice, while retaining dense terms to approximate functions outside the supplied symbolic dictionary. 
Rather than maximizing accuracy alone, S2KAN exposes the tradeoff between predictive performance and model complexity---achieving accuracy competitive with baseline KANs, often with far smaller models.

The rest of this paper is organized as follows. 
Section~\ref{sec:background} describes Kolmogorov--Arnold networks and the standard symbolification method.
Section~\ref{sec:symbolic-fidelity} formalizes the notion of symbolic fidelity.
Section~\ref{sec:s2kan} presents the S2KAN method: the activation function dictionary, differentiable gating mechanism, and MDL-inspired loss function. 
Section~\ref{sec:results} evaluates S2KAN on real-world prediction tasks, chaotic dynamical systems, and symbolic benchmarks. 
Section~\ref{sec:learning-dynamics} examines gate dynamics during training. 
We conclude with a discussion in Sec.~\ref{sec:discussion}.

\section{Background}
\label{sec:background}

\subsection{Kolmogorov--Arnold Networks}

Motivated by the Kolmogorov--Arnold Representation Theorem~\citep{kolmogorov1961representation,arnold2009functions,kolmogorov1957representations}, 
a KAN with $L$ layers and shape $[n_0, n_1, \ldots, n_L]$ computes
\begin{equation}
    x^{(\ell+1)}_j = \sum_{i=1}^{n_\ell} \phi_{\ell ij}(x^{(\ell)}_i)
    \label{eqn:activation-update}
\end{equation}
where $\phi_{\ell ij}: \mathbb{R} \to \mathbb{R}$ is a learnable univariate activation function on the edge from neuron $i$ in layer $\ell$ to neuron $j$ in layer $\ell+1$.
The network output results from composing each layer of activation functions in sequence:
\begin{equation}
\mathrm{KAN}(\mathbf{x}) = \left(\Phi_{L-1} \circ \Phi_{L-2} \circ \cdots \circ \Phi_1 \circ \Phi_0\right) \left(\mathbf{x}\right),
\end{equation}
where $\mathbf{x}_{\ell+1} = \Phi_{\ell}\left(\mathbf{x}_\ell\right)$ is the matrix form of~\eqref{eqn:activation-update} and $\Phi_\ell$ is the matrix containing the $\phi$'s connecting layer $\ell$ and $\ell + 1$.

The $\phi_{\ell ij}$ can be parameterized in many ways, including radial basis functions~\citep{li2024fastkan}, Fourier series~\citep{fourierKAN} or sinusoidal functions~\citep{reinhardt2024sinekan}, wavelets~\citep{bozorgasl2405wav}, and Chebyshev polynomials~\citep{sidharth2024chebyshev}. 
The original KAN formulation~\citep{liu2025kan,liu2024kan2.0} used a combination of a B-spline and a fixed base function $b(x)$, typically the sigmoid linear unit (SiLU):
\begin{equation}
    \phi(x) = w_b b(x) + w_s \mathrm{spline}(x),
    \label{eqn:kan-formulation-base+spline}
\end{equation}
where $w_b$ and $w_s$ are learnable scale parameters, and
\begin{equation}
    \mathrm{spline}(x) = \sum_{m=1}^{G+K} c_{m} B_m(x),
\end{equation}
where $B_m$ are B-spline basis functions of order $K$ over $G$ grid intervals, and $c_{m}$ are learnable coefficients. 
The nonpolynomial base function $b(x)$ supplements the piecewise-polynomial spline representation.

When using B-splines, KANs also implement grid updates, where the spline knots are periodically adjusted to better cover the activation function's input range, and grid refinement, where the number of knots is increased during training; see \citet{liu2025kan} for details.
KAN 2.0~\citep{liu2024kan2.0} introduces multiplication nodes (Fig.~\ref{fig:sincx}), which compute products of sums of incoming activations, enabling more compact representations of multiplicative functions.

Standard KANs can be symbolified after training by fitting their learned activations to symbolic functions~\citep{liu2025kan,liu2024kan2.0}.
Predictive accuracy alone does not ensure that these activations admit simple symbolic approximations.
This motivates both our symbolic-fidelity measure and the inclusion of symbolic primitives during training.

\section{Symbolic fidelity}
\label{sec:symbolic-fidelity}

We now formalize the notion of symbolic fidelity introduced informally above.

Let $G=(V,E)$ be the graph of a KAN of shape $[n_0,\ldots,n_L]$: each node sums its incoming edges (a multiplication unit is a product of such summation subnodes), and each edge $e$ carries one learned univariate Borel function $\phi_e$; $\phi:=(\phi_e)_{e\in E}$ is the network's \emph{decomposition}. Let $\mu$ be an input distribution on $\R^{n_0}$ and $\mu^{\phi}_e$ the distribution of the preactivation of edge $e$ (the pushforward of $\mu$ to the input of $\phi_e$); assume $\phi_e\in L^2(\mu^\phi_e)$ (bounded on the support suffices). Let $f=\Phi(\phi)\in L^2(\mu)$ be the network output. Edge $e$ is \emph{functionally active} for the purposes of symbolic fidelity if
\begin{equation}
\operatorname{Var}_{\mu^\phi_e}(\phi_e)>\eta\operatorname{Var}_\mu(z_j)
\quad\text{and}\quad
\operatorname{Var}_{\mu^\phi_e}(\phi_e)>\delta\operatorname{Var}_\mu(f),
\end{equation}
where $z_j$ is the value of the summation (sub)node that $e$ enters, $\eta>0$ is a fixed fraction, and $\delta\ge0$ is a numerical functional-activity floor. Write $E_{\eta,\delta}(\phi)\subseteq E$ for this set. Our computations use $\eta=10^{-2}$ and $\delta=10^{-8}$; setting $\delta=0$ recovers the functional-activity rule without the numerical floor.
This functional-activity criterion is distinct from gate activity: an edge with an open gate need not contribute to the SF denominator.

A \emph{symbolic language} $\Sigma:=(P,\T,J,\epsilon)$ consists of $P$, a finite set of Borel functions, each defined on a subset of $\R$ (e.g.\ $x$, $\sin$, $\exp$, $\log$, $1/x$), with $x\in P$; $\T:=\{x\mapsto a\,g(bx+c)+d:\ a,b\neq0\}$ acting on functions $g$; a budget $J\in\mathbb{N}$; and a tolerance $\epsilon\in[0,1)$. The \emph{expressions} $\C$ form the smallest set containing the constants and $\T P$ and closed under $+$, $\cdot$, and $\circ$, each expression defined where its constituents are; $\C_J\subseteq\C$ are those using at most $J$ occurrences of elements of $P$. %

The evaluation primitives $P$ need not coincide with the symbolic primitives supplied during S2KAN training (Sec.~\ref{subsec:act-fun-dict}).
Within each comparison, both methods are evaluated using the same language, budget, and tolerance; a spline activation can therefore qualify regardless of its parameterization.
An activation consisting of a sum of at most $J$ affinely transformed primitives from the evaluation language is expressible by construction. Overlap with the training dictionary can therefore favor representations built from those primitives; SF measures expressibility in the declared language rather than dictionary-independent interpretability.

\begin{definition}[Expressible edge]
Given the decomposition $\phi$, edge $e$ is \emph{expressible} with respect to $\Sigma$ if some $s\in\C_J$, defined $\mu^\phi_e$-a.e., satisfies
\begin{equation}
\|\phi_e-s\|_{L^2(\mu^{\phi}_e)}\ \le\ \epsilon\,\|\phi_e-\E_{\mu^{\phi}_e}\phi_e\|_{L^2(\mu^{\phi}_e)} .
\end{equation}
The norm on the right is the standard deviation of $\phi_e$, so for a non-constant edge this is $R^2_{\mu^\phi_e}(\phi_e,s)\ge1-\epsilon^2$; the other edges of $\phi$ enter only through $\mu^\phi_e$. Write $\Lambda(\phi)\subseteq E$ for the set of expressible edges.
\end{definition}

\begin{definition}[Symbolic fidelity]
The \emph{symbolic fidelity} of a KAN with respect to $\mu$, $\eta$, $\delta$, and $\Sigma$ is $|\Lambda(\phi)\cap E_{\eta,\delta}(\phi)|/|E_{\eta,\delta}(\phi)|$, the fraction of its functionally active edges that are, on the values that reach them, within tolerance of a short closed form; it is taken as $1$ if $E_{\eta,\delta}(\phi)=\emptyset$ (a vacuous convention, not evidence of symbolic recovery; an empty functional-activity set does not by itself imply a constant network). The network is \emph{fully faithful} if its symbolic fidelity is $1$, i.e.\ $E_{\eta,\delta}(\phi)\subseteq\Lambda(\phi)$.
Constants are included in the symbolic language, but functionally inactive edges do not contribute to the fidelity score; thus fully faithful means that every functionally active edge is expressible.
\end{definition}

SF measures expressibility of individual activation functions, not the readability of their composition: even a fully faithful network can produce a cumbersome overall expression.

To estimate symbolic fidelity empirically, we fit short sums of affinely transformed primitives to sampled responses of each functionally active edge and report the fraction meeting the prescribed approximation tolerance.
This finite search considers additive expressions only and need not find every edge expressible in the broader formal language.
The procedure builds on the affine symbolic fitting used in KAN post-hoc symbolification~\citep{liu2025kan,liu2024kan2.0}, extending it to short sums of primitives for evaluation without replacing the learned activations.
The sampling and fitting procedures are described in \methods{}.

\section{S2KAN: softly symbolified KANs}

\label{sec:s2kan}

Instead of learning activations and converting them post hoc, we propose learning activation functions as combinations of the standard KAN components and other symbolic terms. 
Depending on the choice of terms, the activation functions can form an overcomplete dictionary of functions (Sec.~\ref{subsec:act-fun-dict}) so sparsification will be necessary to avoid overfitting and stabilize training.
Sparse basis selection can be approached with $\ell_1$ regularization, as in the least absolute shrinkage and selection operator (LASSO). However, simply adding an $\ell_1$ penalty to the loss optimized by Adam generally shrinks coefficients without producing exact zeros~\citep{ziyin2023spred}, whereas we seek explicit term selection during end-to-end training across the composed layers of a KAN.
We therefore introduce binary gating variables that select which terms are active, made differentiable via a continuous relaxation of the $\ell_0$ norm (Sec.~\ref{subsec:diff-l0-sparse}), with selection guided by an objective inspired by minimum description length (Sec.~\ref{subsec:loss-function-via-mdl}).

\subsection{Activation function dictionary}
\label{subsec:act-fun-dict}

We organize the activation function using three categories of terms (or atoms):
\begin{equation}
\phi(x) = \sum_{s \in \mathcal{S}} z_s c_s \psi_s(x) + \sum_{f \in \mathcal{F}} z_f c_f \psi_f(x) + \sum_{r \in \mathcal{R}} z_r \phi_r(x; \mathbf{c}_r),
\label{eqn:activation-function-dictionary}
\end{equation}
where the $z \in \{0,1\}$ are binary gates selecting active terms (Sec.~\ref{subsec:diff-l0-sparse}), $c \in \mathbb{R}$ are learnable coefficients, and each $\psi:D_\psi\to\mathbb R$, with $D_\psi\subseteq\mathbb R$, is a candidate univariate function.
Here $\mathcal{S}$ indexes the symbolic library, $\mathcal{F}$ indexes sparse function bases, and $\mathcal{R}$ indexes dense parameterized representations.
Equation~\eqref{eqn:activation-function-dictionary} allows researchers to design problem-specific combinations as needed.
Using the standard KAN B-spline dense representation for $\mathcal{R}$, setting $\mathcal{S} = \mathcal{F} = \emptyset$, and fixing $z_r=1$ for all $r\in\mathcal{R}$, Eq.~\eqref{eqn:activation-function-dictionary} recovers the original KAN formulation.

In this work, we consider the following dictionary:
\begin{description}
\item[Symbolic library ($\mathcal{S}$)]
A collection of elementary functions such as $1, x, \sin(x), 1/x, \log|x|, \ldots$. Each function receives an independent gate and coefficient.
\item[Sparse function bases ($\mathcal{F}$)]
Families of orthogonal or structured functions indexed by degree/frequency:
\begin{itemize}[leftmargin=*]
\item \textit{Chebyshev polynomials:} $P$ terms $T_p(x)$, $p = 0, 1, \ldots, P-1$, computed via recurrence.
\item \textit{Fourier basis:} $\{\sin(qx), \cos(qx)\}$ for mode $q = 1, \ldots, Q$.
\end{itemize}
Setting $P=0$ or $Q=0$ removes the Chebyshev or Fourier family, respectively.
Each term receives an independent gate and coefficient.
\item[Dense representations ($\mathcal{R}$)]
A single gate controls an entire parameterized function class:
\begin{itemize}[leftmargin=*]
\item \textit{B-spline with SiLU residual:} $\phi_{\text{spline}}(x; \mathbf{c}) = w_b \mathrm{SiLU}(x) + w_s \sum_{m=1}^{B} c_m B_m(x)$, where $B=G+K$, the $B_m$ are B-spline basis functions, and $\mathbf{c} = (w_b, w_s, c_1, \ldots, c_B)$ contains the two learnable scales and the spline coefficients.
\end{itemize}
\end{description}

This formulation allows for graceful degradation: when symbolic terms or their composition within the chosen architecture are insufficient, the model can fall back to the dense spline representation.

\subsection{Differentiable sparsity via Hard Concrete Distribution}
\label{subsec:diff-l0-sparse}

We follow the $\ell_0$ regularization approach of \citet{louizos2018learning}, which 
leverages the reparameterization trick to provide a continuous, differentiable relaxation of binary gates while maintaining the ability to produce exact zeros and ones.

For each gate $i$, introduce a learnable parameter $\alpha_i \in \mathbb{R}$ and define a continuous relaxation of a binary variable via the Concrete (Gumbel-Softmax) distribution with temperature $\tau > 0$:
\begin{equation}
s_i = \sigma\left(\frac{\log u - \log(1-u) + \alpha_i}{\tau}\right), \quad u \sim \text{Uniform}(0,1),
\end{equation}
where $\sigma(\cdot)$ is the sigmoid function. This provides a continuous relaxation in $(0, 1)$, but cannot produce exact boundary values.
To assign finite probability mass to exactly 0 and 1, apply a stretched and rectified transformation with parameters $\gamma < 0 < 1 < \zeta$ (typically $\gamma = -0.1, \zeta = 1.1$):
\begin{equation}
\bar{s}_i = s_i(\zeta - \gamma) + \gamma,
\end{equation}
\begin{equation}
\tilde{z}_i = \min(1, \max(0, \bar{s}_i)).
\end{equation}
The stretch extends the range to $(\gamma, \zeta)$ before clipping to $[0, 1]$, allowing the relaxed gate to reach the boundaries exactly during training.

The probability that this gate is nonzero, used for regularization (Sec.~\ref{subsec:loss-function-via-mdl}), has a closed form:
\begin{equation}
p_i := \Pr(\tilde{z}_i>0) = \sigma\left(\alpha_i - \tau \log \frac{-\gamma}{\zeta}\right).
\label{eqn:expected-z}
\end{equation}
This is the expected active-gate indicator, $p_i=\mathbb{E}[\mathds{1}_{\{\tilde{z}_i>0\}}]$, rather than the expected amplitude $\mathbb{E}[\tilde{z}_i]$ of the relaxed gate.
During training, sample $\{\tilde{z}_i\}$ and update their distribution parameters jointly with the other model parameters.
At inference, deterministically threshold: $z_i = \mathds{1}_{\{p_i > 1/2\}}$.

The gradient of the expected $\ell_0$ penalty with respect to gate parameters is
\begin{equation}
\frac{\partial p_i}{\partial \alpha_i} = p_i(1-p_i).
\end{equation}
This gradient is maximal when $p_i=1/2$ and vanishes as $p_i$ approaches zero or one.

\paragraph{Remarks}
An activation function sparsifies as its gates close, and the entire function can be omitted when all its gates close, allowing for architectural sparsification.
Beyond the mechanism of sparsity, these gates offer additional benefits.
Initializing $\alpha_i$ acts as a prior, and one can tune the model towards or away from symbolic entries by, for instance, setting $\alpha_i \approx -1$ for dense terms and $\alpha_i \approx 0$ otherwise.
Gate probabilities $p_i$ also provide a structural convergence criterion: training can terminate, possibly with some patience, once gates become decisive (e.g., when most $p_i < 0.01$ or $p_i > 0.99$).

\subsection{MDL-motivated loss function}
\label{subsec:loss-function-via-mdl}

To balance model accuracy while minimizing complexity, our training objective is motivated by the MDL principle, which seeks to minimize the total bits required to encode both the model and the data given the model: %
\begin{equation}
\mathcal{L}_{\text{MDL}} = \mathcal{L}_{\text{model}} + \mathcal{L}_{\text{data}|\text{model}}.
\end{equation}
The first term captures model complexity while the second encodes residuals.
Assuming independent and identically distributed (iid) scalar residuals $\epsilon_t = y_t - \hat{y}_t \sim \mathcal{N}(0, \sigma^2)$, the data encoding term, $\mathcal{L}_{\text{data}|\text{model}}$, is proportional to the negative log-likelihood:
\begin{equation}
\mathcal{L}_{\text{data}|\text{model}} \propto \frac{n}{2}\log \sigma^2 + \frac{1}{2\sigma^2}\sum_{t=1}^n (y_t - \hat{y}_t)^2.
\end{equation}
The maximum-likelihood estimate is the mean squared error (MSE), $\hat{\sigma}^2=\text{MSE}(y,\hat{y})$. Substituting this estimate gives a per-sample data term $\tfrac12\log\text{MSE}(y,\hat{y})$, up to additive constants.
For $\mathcal{L}_{\text{model}}$, the model complexity term, let 
\begin{equation}
\ktot := \sum_{\ell i j} \sum_{m} p_{\ell ijm}
\end{equation}
be the expected number of active terms over the activation functions in the network, where the nonzero-gate probabilities $p_{\ell ijm}$ are given by Eq.~\eqref{eqn:expected-z}. 
Motivated by the Bayesian information criterion (BIC) \citep{schwarz1978estimating}, we assign the unweighted support penalty $\mathcal{L}_{\text{model}} = \frac{\ktot}{2} \log n$. 
Combining these terms with the complexity weights defined below and a stabilized minibatch data term gives the implemented training objective
\begin{equation}
\mathcal{L} = \frac12\log\!\left(\text{MSE}_{\mathcal{B}} + c_{\mathrm{loss}}\right) + \beta \frac{\ktot_w \log n}{2n},
\label{eqn:loss-function}
\end{equation}
where hyperparameter $\beta\geq0$ controls the sparsity-accuracy tradeoff, $\mathcal{B}$ is the current minibatch, $n$ is the full fitting-set size, and $\ktot_w$ is the weighted expected support defined below. MSE is averaged over observations and output coordinates. The additive constant $c_{\mathrm{loss}}=10^{-12}$ stabilizes the natural logarithm and is expressed in squared units of the training target.

For deterministic, full-data scalar Gaussian regression with a discrete parameter count, the corresponding unstabilized objective at $\beta=1$ equals BIC divided by $2n$, up to additive constants.
Minibatch evaluation, stochastic gates, relaxed support counts, and pooled multioutput errors (for vector-valued models) make the implemented objective an MDL/BIC-motivated regularizer rather than exact BIC model selection or a complete description-length code.
Accordingly, $\beta=1$ serves as a motivated reference setting, while $\beta$ remains tunable and may require problem-specific adjustment.
The iid Gaussian assumption also need not hold for every dataset, particularly temporally dependent dynamical-system observations.
More rigorous coding schemes, such as prequential or normalized maximum likelihood coding~\citep{grunwald2007}, are not implemented here.

\subsubsection{Complexity-weighted sparsity}

The unweighted count $\ktot$ assigns a common cost to each active term. 
However, when selecting among terms with varying structural complexity, we may sometimes wish to account for differential encoding costs, which we can do by incorporating \emph{complexity weights}. 
For term $\psi_m$ with complexity weight $w_m > 0$, the weighted number of active terms is
\begin{equation}
\ktot_w := \sum_{\ell i j} \sum_m w_m p_{\ell ijm},
\label{eqn:effective-parameter-count}
\end{equation}
as used in Eq.~\eqref{eqn:loss-function}.
The gradient $\partial \ktot_{{w}} / \partial \alpha_m = w_m p_m(1-p_m)
$ shows that 
higher complexity weights will induce stronger regularization pressure on the corresponding gates.
Appropriate choices for $w_m$ include uniform weighting ($w_m = 1$, recovering the unweighted support penalty), degree-dependent complexity ($w_m = 1 + \log_2(d + 1)$ for degree $d$), or computational cost ($w_m \propto$ the number of floating-point operations, or FLOPs, required to evaluate $\psi_m$).
Hand-tuning $w_m$ is also common practice in symbolic regression.
We assign weight one to each non-dense term, corresponding to its learned coefficient. A dense branch receives weight equal to its number of spline coefficients plus two scale parameters; see \methods{}. Thus $\ktot_w$ is a differentiable surrogate for retained parameter count, rather than a count of statistically identifiable parameters. 
We do not pursue weighting schemes other than this, but it may be beneficial to consider them in the future.

We discuss further setup and training details in \methods{}.

\section{Results}
\label{sec:results}

We first compare predictive accuracy, model size, and symbolic fidelity on four real-world datasets.
We then examine the scientific consistency of their predicted responses and connect selected responses to learned symbolic activations.
Next, we evaluate forecasting and dynamical behavior in two chaotic systems, followed by the Nguyen symbolic regression benchmarks.

\subsection{Real-world data}

We apply S2KAN to four real-world datasets:

\begin{description}
\item[Airfoil self-noise]
Predict the scaled sound pressure level (in dB) generated by an airfoil from five inputs: frequency (Hz), angle of attack (degrees), chord length (m), free-stream velocity (m/s), and suction-side displacement thickness (m).
The University of California, Irvine (UCI) dataset contains 1503 observations from National Aeronautics and Space Administration (NASA) wind-tunnel tests of National Advisory Committee for Aeronautics (NACA) 0012 airfoils of different sizes and at different flow conditions~\citep{brooks1989airfoil}.
We use the five supplied features without additional derived features.
For modeling, we allocated 1000 observations to fitting and validation (800 and 200, respectively), with the remaining 503 observations used for testing.

\item[Concrete compressive strength]

Predict the compressive strength (in MPa) of concrete based on sample properties. 
The dataset contains 1030 samples with eight features: 
cement, blast furnace slag, fly ash, water, superplasticizer, coarse aggregate, fine aggregate (all in kg/m$^{3}$), and age (days). 
Compressive strength in concrete is well known to depend on the water-to-cement ratio~\citep{neville2011properties}, so we include this as a derived feature. 
We also compute total binder (cement plus slag and fly ash), total aggregate (coarse plus fine), and the water-to-binder ratio, which generalizes the water-to-cement relationship when supplementary cementitious materials are present. 
We include $\log(\text{age}+1)$ as strength gain in concrete follows an approximately logarithmic relationship with curing time~\citep{neville2011properties}, motivating the inclusion of transformed age variables.
For modeling, we allocated 824 observations to fitting and validation (659 and 165, respectively), with the remaining 206 observations used for testing.
Data were collected by \citet{YEH19981797,YEHconcrete2}.

\item[Power plant]
Predict the net hourly electrical power output (in MW) of a combined-cycle power plant from four inputs: ambient temperature (${}^{\circ}$C), exhaust vacuum (cm Hg), ambient pressure (mbar), and relative humidity (\%).
The UCI dataset contains 9568 observations collected during full-load operation over 2006--2011; the inputs are hourly averages of sensor measurements~\citep{tufekci2014prediction}.
We use the four supplied features without additional derived features.
For modeling, we sampled 1000 observations for fitting and validation (800 and 200, respectively) and 1000 separate observations for testing.

\item[Superconductor critical temperature]
Predict critical temperature (in K) of superconductors based on their material properties. 
The original dataset includes many features and derived statistics; we focus on five that capture composition, electronic structure, and bonding: number of elements, weighted mean valence, valence entropy, weighted mean first ionization energy, and mean electron affinity.
These same representative features were used in prior KAN modeling~\citep{Bagrow_2025}.
For modeling, we sampled 1000 observations for fitting and validation (800 and 200, respectively) and 1000 separate observations for testing.
The data come from Japan's National Institute for Materials Science superconductor database~\citep{hamidieh2018data,center-a}.
\end{description}

Details of data preprocessing, model fitting, and validation-based checkpoint selection are given in \methods{}.
Baseline KAN retains the default MSE loss of the authors' implementation~\citep{liu2025kan}, whereas S2KAN uses the log-MSE-based objective in Eq.~\eqref{eqn:loss-function}.

As shown in Table~\ref{tab:real-world-both}, across all four datasets, S2KAN found considerably more compact representations at $\beta=1$, with no spline terms retained for concrete or power plant.
For concrete, median test $R^2$ was 0.894 versus 0.915 for baseline KAN, with only 61 versus 448 active functions.
For superconductivity, median test $R^2$ was 0.682 versus 0.696, with 61.3 versus 450 effective parameters.
S2KAN achieved slightly higher median test $R^2$ with smaller models on airfoil and power plant.

S2KAN also achieved higher median symbolic fidelity on all four datasets at all six settings of expression budget and approximation tolerance.
This fidelity advantage persists when baseline KAN is trained with log MSE (App.~\ref{sec:kan-loss-control}).

Appendix~\ref{sec:beta-sensitivity} examines how predictive accuracy and model size vary with $\beta$; App.~\ref{sec:eql-like-control} compares full S2KAN with no-dense and dense-only variants.

\begin{table*}[t]
\centering
\caption{Prediction, model size, and symbolic fidelity on four real-world datasets. S2KAN yields substantially smaller models and higher symbolic fidelity, with modest differences in predictive accuracy. 
Test accuracy is reported as $R^2$ and root mean squared error (RMSE). Values are medians over ten data splits with three initialization seeds per split; brackets are interquartile ranges (IQRs; omitted for constant values). 
S2KAN uses $\beta=1$ after warmup. ``Active'' counts gated-on activation functions, and $\ktot$ denotes the effective parameter count $\ktot_w$ from Eq.~\eqref{eqn:effective-parameter-count}. Symbolic fidelity (SF, percent; Section~\ref{sec:symbolic-fidelity}) is the fraction of functionally active edges approximated within relative tolerance $\epsilon$ using at most $J$ terms.}
\label{tab:real-world-both}
\begingroup
\small
\setlength{\tabcolsep}{4pt}
\renewcommand{\arraystretch}{1.13}
\begin{tabular}{@{}lrrrrrrrrrr@{}}
\toprule
 & \multicolumn{2}{c}{Test accuracy} & \multicolumn{2}{c}{Model size} & \multicolumn{3}{c}{SF (\%), $\epsilon=0.01$} & \multicolumn{3}{c}{SF (\%), $\epsilon=0.1$} \\
\cmidrule(lr){2-3}\cmidrule(lr){4-5}\cmidrule(lr){6-8}\cmidrule(lr){9-11}
Model & $R^2$ & RMSE & Active & $\ktot$ & $J=1$ & $J=2$ & $J=3$ & $J=1$ & $J=2$ & $J=3$ \\
\midrule
\multicolumn{11}{@{}l}{\textbf{Airfoil}: shape $[5,5,1]$; RMSE in dB} \\
\addlinespace[2pt]
KAN & \shortstack[r]{0.911\\[-1pt]{\fontsize{6}{7}\selectfont [0.902, 0.918]}} & \shortstack[r]{2.05\\[-1pt]{\fontsize{6}{7}\selectfont [1.96, 2.13]}} & \shortstack[r]{30\\[-1pt]{\fontsize{6}{7}\selectfont \vphantom{[0.0, 0.0]}}} & \shortstack[r]{450\\[-1pt]{\fontsize{6}{7}\selectfont \vphantom{[0.0, 0.0]}}} & \shortstack[r]{10.5\\[-1pt]{\fontsize{6}{7}\selectfont [7.2, 13.5]}} & \shortstack[r]{26.8\\[-1pt]{\fontsize{6}{7}\selectfont [24.4, 28.6]}} & \shortstack[r]{28.6\\[-1pt]{\fontsize{6}{7}\selectfont [28.6, 31.0]}} & \shortstack[r]{14.3\\[-1pt]{\fontsize{6}{7}\selectfont [11.1, 17.7]}} & \shortstack[r]{39.3\\[-1pt]{\fontsize{6}{7}\selectfont [36.0, 46.0]}} & \shortstack[r]{59.6\\[-1pt]{\fontsize{6}{7}\selectfont [51.8, 62.1]}} \\
S2KAN & \shortstack[r]{0.912\\[-1pt]{\fontsize{6}{7}\selectfont [0.909, 0.920]}} & \shortstack[r]{2.02\\[-1pt]{\fontsize{6}{7}\selectfont [1.95, 2.05]}} & \shortstack[r]{26\\[-1pt]{\fontsize{6}{7}\selectfont [24.2, 26.0]}} & \shortstack[r]{103.8\\[-1pt]{\fontsize{6}{7}\selectfont [94.7, 114.1]}} & \shortstack[r]{22.4\\[-1pt]{\fontsize{6}{7}\selectfont [20.0, 25.7]}} & \shortstack[r]{50.0\\[-1pt]{\fontsize{6}{7}\selectfont [46.3, 52.2]}} & \shortstack[r]{64.0\\[-1pt]{\fontsize{6}{7}\selectfont [60.0, 69.2]}} & \shortstack[r]{33.3\\[-1pt]{\fontsize{6}{7}\selectfont [29.4, 39.8]}} & \shortstack[r]{84.3\\[-1pt]{\fontsize{6}{7}\selectfont [82.6, 87.5]}} & \shortstack[r]{91.3\\[-1pt]{\fontsize{6}{7}\selectfont [87.5, 92.0]}} \\
\addlinespace[5pt]
\multicolumn{11}{@{}l}{\textbf{Concrete}: shape $[13,32,1]$; RMSE in MPa} \\
\addlinespace[2pt]
KAN & \shortstack[r]{0.915\\[-1pt]{\fontsize{6}{7}\selectfont [0.898, 0.925]}} & \shortstack[r]{4.90\\[-1pt]{\fontsize{6}{7}\selectfont [4.49, 5.23]}} & \shortstack[r]{448\\[-1pt]{\fontsize{6}{7}\selectfont \vphantom{[0.0, 0.0]}}} & \shortstack[r]{6720\\[-1pt]{\fontsize{6}{7}\selectfont \vphantom{[0.0, 0.0]}}} & \shortstack[r]{0.0\\[-1pt]{\fontsize{6}{7}\selectfont [0.0, 0.3]}} & \shortstack[r]{2.6\\[-1pt]{\fontsize{6}{7}\selectfont [1.9, 3.4]}} & \shortstack[r]{12.9\\[-1pt]{\fontsize{6}{7}\selectfont [11.6, 13.9]}} & \shortstack[r]{26.2\\[-1pt]{\fontsize{6}{7}\selectfont [19.3, 29.0]}} & \shortstack[r]{61.5\\[-1pt]{\fontsize{6}{7}\selectfont [47.2, 65.1]}} & \shortstack[r]{83.2\\[-1pt]{\fontsize{6}{7}\selectfont [71.5, 86.9]}} \\
S2KAN & \shortstack[r]{0.894\\[-1pt]{\fontsize{6}{7}\selectfont [0.885, 0.906]}} & \shortstack[r]{5.27\\[-1pt]{\fontsize{6}{7}\selectfont [5.00, 5.61]}} & \shortstack[r]{61\\[-1pt]{\fontsize{6}{7}\selectfont [58.0, 63.8]}} & \shortstack[r]{76.2\\[-1pt]{\fontsize{6}{7}\selectfont [71.7, 80.4]}} & \shortstack[r]{77.8\\[-1pt]{\fontsize{6}{7}\selectfont [73.7, 79.5]}} & \shortstack[r]{85.4\\[-1pt]{\fontsize{6}{7}\selectfont [83.5, 88.6]}} & \shortstack[r]{86.9\\[-1pt]{\fontsize{6}{7}\selectfont [85.1, 90.4]}} & \shortstack[r]{82.1\\[-1pt]{\fontsize{6}{7}\selectfont [78.5, 85.6]}} & \shortstack[r]{90.3\\[-1pt]{\fontsize{6}{7}\selectfont [89.1, 94.2]}} & \shortstack[r]{94.4\\[-1pt]{\fontsize{6}{7}\selectfont [91.9, 96.1]}} \\
\addlinespace[5pt]
\multicolumn{11}{@{}l}{\textbf{Power plant}: shape $[4,5,1]$; RMSE in MW} \\
\addlinespace[2pt]
KAN & \shortstack[r]{0.932\\[-1pt]{\fontsize{6}{7}\selectfont [0.930, 0.936]}} & \shortstack[r]{4.41\\[-1pt]{\fontsize{6}{7}\selectfont [4.27, 4.50]}} & \shortstack[r]{25\\[-1pt]{\fontsize{6}{7}\selectfont \vphantom{[0.0, 0.0]}}} & \shortstack[r]{375\\[-1pt]{\fontsize{6}{7}\selectfont \vphantom{[0.0, 0.0]}}} & \shortstack[r]{0.0\\[-1pt]{\fontsize{6}{7}\selectfont \vphantom{[0.0, 0.0]}}} & \shortstack[r]{0.0\\[-1pt]{\fontsize{6}{7}\selectfont \vphantom{[0.0, 0.0]}}} & \shortstack[r]{0.0\\[-1pt]{\fontsize{6}{7}\selectfont [0.0, 7.2]}} & \shortstack[r]{12.8\\[-1pt]{\fontsize{6}{7}\selectfont [8.7, 20.0]}} & \shortstack[r]{51.0\\[-1pt]{\fontsize{6}{7}\selectfont [43.6, 56.4]}} & \shortstack[r]{74.5\\[-1pt]{\fontsize{6}{7}\selectfont [68.4, 80.3]}} \\
S2KAN & \shortstack[r]{0.937\\[-1pt]{\fontsize{6}{7}\selectfont [0.931, 0.941]}} & \shortstack[r]{4.19\\[-1pt]{\fontsize{6}{7}\selectfont [4.10, 4.47]}} & \shortstack[r]{16\\[-1pt]{\fontsize{6}{7}\selectfont [15.0, 17.0]}} & \shortstack[r]{36.7\\[-1pt]{\fontsize{6}{7}\selectfont [34.3, 39.4]}} & \shortstack[r]{35.5\\[-1pt]{\fontsize{6}{7}\selectfont [30.2, 39.6]}} & \shortstack[r]{64.3\\[-1pt]{\fontsize{6}{7}\selectfont [59.1, 71.4]}} & \shortstack[r]{69.9\\[-1pt]{\fontsize{6}{7}\selectfont [64.3, 79.6]}} & \shortstack[r]{50.0\\[-1pt]{\fontsize{6}{7}\selectfont [46.2, 57.1]}} & \shortstack[r]{92.6\\[-1pt]{\fontsize{6}{7}\selectfont [85.7, 93.9]}} & \shortstack[r]{92.9\\[-1pt]{\fontsize{6}{7}\selectfont [92.3, 100.0]}} \\
\addlinespace[5pt]
\multicolumn{11}{@{}l}{\textbf{Superconductor}: shape $[5,5,1]$; RMSE in K} \\
\addlinespace[2pt]
KAN & \shortstack[r]{0.696\\[-1pt]{\fontsize{6}{7}\selectfont [0.669, 0.727]}} & \shortstack[r]{18.80\\[-1pt]{\fontsize{6}{7}\selectfont [18.22, 19.43]}} & \shortstack[r]{30\\[-1pt]{\fontsize{6}{7}\selectfont \vphantom{[0.0, 0.0]}}} & \shortstack[r]{450\\[-1pt]{\fontsize{6}{7}\selectfont \vphantom{[0.0, 0.0]}}} & \shortstack[r]{0.0\\[-1pt]{\fontsize{6}{7}\selectfont \vphantom{[0.0, 0.0]}}} & \shortstack[r]{0.0\\[-1pt]{\fontsize{6}{7}\selectfont [0.0, 3.7]}} & \shortstack[r]{16.7\\[-1pt]{\fontsize{6}{7}\selectfont [15.3, 16.7]}} & \shortstack[r]{3.3\\[-1pt]{\fontsize{6}{7}\selectfont [0.0, 3.6]}} & \shortstack[r]{19.3\\[-1pt]{\fontsize{6}{7}\selectfont [16.7, 23.3]}} & \shortstack[r]{33.9\\[-1pt]{\fontsize{6}{7}\selectfont [30.0, 36.7]}} \\
S2KAN & \shortstack[r]{0.682\\[-1pt]{\fontsize{6}{7}\selectfont [0.637, 0.698]}} & \shortstack[r]{19.51\\[-1pt]{\fontsize{6}{7}\selectfont [18.85, 20.27]}} & \shortstack[r]{16\\[-1pt]{\fontsize{6}{7}\selectfont [14.2, 19.0]}} & \shortstack[r]{61.3\\[-1pt]{\fontsize{6}{7}\selectfont [51.8, 74.8]}} & \shortstack[r]{19.4\\[-1pt]{\fontsize{6}{7}\selectfont [13.4, 27.5]}} & \shortstack[r]{39.4\\[-1pt]{\fontsize{6}{7}\selectfont [30.1, 49.3]}} & \shortstack[r]{53.3\\[-1pt]{\fontsize{6}{7}\selectfont [44.2, 60.2]}} & \shortstack[r]{25.0\\[-1pt]{\fontsize{6}{7}\selectfont [15.7, 33.3]}} & \shortstack[r]{55.6\\[-1pt]{\fontsize{6}{7}\selectfont [50.0, 65.6]}} & \shortstack[r]{65.9\\[-1pt]{\fontsize{6}{7}\selectfont [57.1, 73.6]}} \\
\bottomrule
\end{tabular}
 \endgroup
\end{table*}

\subsection{Scientific response consistency and symbolic interpretation}
\label{subsec:scientific-insights}

We evaluated selected scientifically meaningful responses and physical admissibility across the four real-world datasets, using the same validation-selected checkpoints as Table~\ref{tab:real-world-both}: ten data splits and three initialization seeds per split.
Unless stated otherwise, percentages below are medians of per-model fractions over the 30 fits per method, and split comparisons average the three initialization seeds.
These exploratory analyses use frozen models without further training or symbolic refitting; the overlapping splits provide descriptive comparisons rather than independent experiments.
Appendix~\ref{sec:scientific-response-diagnostics} gives the diagnostic definitions, secondary comparisons, and sensitivity checks, including the Airfoil peak and speed analyses and the Power Plant pressure analysis.

\paragraph{Airfoil self-noise.}
For fixed airfoil geometry and flow conditions, the measured sound level varies with frequency and can decline after a spectral peak.
Preserving these observed declines provides a response-consistency check: substantial upward reversals would contradict the measured trend.
Because airfoil self-noise comprises several mechanisms with different spectral responses~\citep{brooks1989airfoil}, we restrict this check to recorded conditions exhibiting a clear decline.
Among 106 recorded spectra, we identified 92 with a clear measured decline after their peak and evaluated the models over these observed frequency ranges.
A pass required a predicted net decrease exceeding 1~dB and no upward reversal exceeding 1~dB, so flat predictions failed.
S2KAN passed on a median 94.6\% of spectra, compared with 70.1\% for KAN, with all ten split averages favoring S2KAN.
Held-out centered spectral errors were similar (1.40 versus 1.43~dB for S2KAN and KAN).
The clearest advantage was therefore preservation of measured spectral decline; selection of the tails using observed responses makes this a descriptive diagnostic rather than a wholly held-out test.

\paragraph{Concrete compressive strength.}
As concrete cures, continuing cement hydration generally increases its compressive strength, with gains slowing at later ages.
Substantial predicted strength declines at fixed composition are therefore inconsistent with this expected curing response.
Across 181 fixed-composition recipes, S2KAN predicted strength declines exceeding 1~MPa for 8.8\% of recipes, compared with 63.5\% for KAN.
Positive average growth with smaller late-age than early-age gains occurred for 93.4\% versus 75.4\% of recipes.
Both comparisons favored S2KAN in all ten splits under either mean or median aggregation over seeds, despite higher test RMSE (5.27 versus 4.90~MPa).
These related aging diagnostics sweep each recipe's recorded age range, changing age and the supplied $\log(\mathrm{age}+1)$ feature together; they assess response consistency rather than causal curing laws.

\paragraph{Power plant electrical output.}
For a combined-cycle power plant, warmer ambient air generally reduces the electrical power generated at full load~\citep{poncearrieta2005ambient}.
Preserving this temperature penalty under small changes in ambient conditions provides a check of the model's consistency with the expected plant response.
We evaluated 325 local temperature sweeps spanning $\pm2\,{}^{\circ}\mathrm{C}$ around recorded conditions, holding the other inputs fixed and retaining sweeps within observed joint-input support.
A pass required a net decrease exceeding 0.2~MW and no upward excursion exceeding 1~MW.
The median passing fraction was 100.0\% for S2KAN and 98.2\% for KAN, with all ten split averages favoring S2KAN.
Both methods usually captured the expected direction, making the absolute improvement modest.
All 30 Power Plant S2KAN fits closed their dense branches, allowing their responses to be traced through explicit symbolic and finite-basis activations.
In one fixed example, a decreasing polynomial temperature activation feeds an increasing linear-plus-sine output activation over the visited hidden-input range.
This composition contributes a 0.92~MW decrease over 18.34--22.34~$^{\circ}\mathrm{C}$, within the complete model's 6.25~MW decrease.
It illustrates how a learned expression implements the expected sign over a supported interval, without identifying a thermodynamic mechanism.

\paragraph{Superconductor critical temperature.}
The critical temperature $T_c$ marks the onset of superconductivity and must be nonnegative on the kelvin scale, so nonnegativity provides a basic check of physical admissibility.
S2KAN produced nonphysical $T_c<0$ predictions for 2.65\% of held-out observations, versus 5.10\% for KAN.
After clipping negative predictions to zero for both methods, median test RMSE remained higher for S2KAN (19.44 versus 18.73~K).

The clearest response-consistency improvements are for concrete aging and airfoil spectral decline, with a smaller improvement for Power Plant temperature responses and fewer physically inadmissible Superconductor predictions.
These diagnostics compare the complete training configurations, including their different losses and regularization, and do not isolate symbolic gating as the cause.

\begin{figure}
\centering
\includegraphics[width=\textwidth]{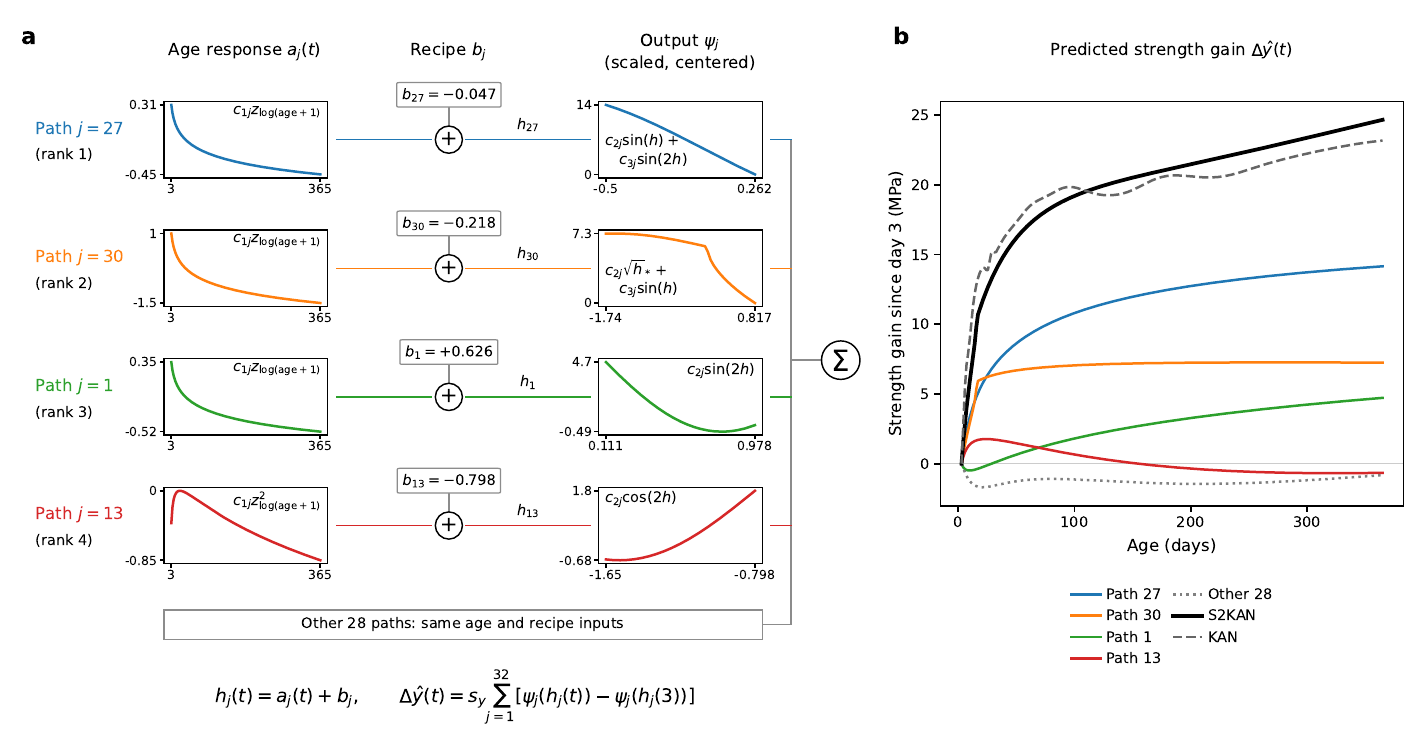}
\caption{\textbf{Illustrative model-response analysis of predicted concrete strength development with age.}
Consistent with expected curing behavior, the complete model predicts strength increasing with age at an overall slowing rate.
The dominant branch exhibits a decreasing rate of gain throughout the evaluated interval.
\textbf{a}, Four hidden branches for a fixed recipe, ranked by total variation of their output contributions over days 3--365.
Each $b_j$ sums the learned input activations evaluated at the recipe's 11 fixed non-age features.
Inset formulas show the learned functions, with branch-specific coefficients $c_{kj}$; $z_{\log(\mathrm{age}+1)}$ denotes standardized log-age, and $\sqrt{h}_*$ denotes the protected square root.
These branches have no active raw-age or dense terms.
Output curves are converted to MPa and centered at day~3.
\textbf{b}, Branch contributions and complete S2KAN and KAN predictions relative to day~3.
All 32 branches remain in the S2KAN total.
Both models use the same data split and initialization seed; the recipe was selected by proximity to the median standardized ingredient composition among 181 eligible recipes, without using strengths or predictions.
Here $s_y$ is the population standard deviation of the fitting-set targets.}
\label{fig:concrete-symbolic-aging}
\end{figure}

To illustrate how learned symbolic activations produce one of these responses, Fig.~\ref{fig:concrete-symbolic-aging} traces the age response of one fitted S2KAN for a fixed concrete recipe.
In the four leading branches, the raw-age edges are inactive and age dependence comes entirely from the supplied log-age feature, demonstrating selection of a domain-informed transformation available to both models.
The dominant branch combines a linear function of standardized log-age with two Fourier terms.
Over the hidden-input range visited here, this composition yields an increasing strength contribution with a progressively slower rate of gain; the trigonometric terms do not imply a periodic age response.
This branch contributes 14.2~MPa of the complete model's predicted 24.7~MPa increase from day~3 to day~365.
This agreement holds over the evaluated age range; the Fourier terms do not enforce monotonic strength development under extrapolation.

\subsection{Data-driven modeling of dynamical systems}
\label{subsec:results:dynamical-systems}

We evaluate S2KAN on two dynamical systems that present challenging forecasting problems due to their chaotic attractors.

The first is the \textit{Ikeda map}~\citep{ikeda1979multiple,hammel1985global}, a discrete-time chaotic system arising from nonlinear optics that resists discovery by sparse regression methods like sparse identification of nonlinear dynamics (SINDy)~\citep{sindy2016}:
\begin{equation}
\begin{gathered}
x_{n+1} = 1 + \mu \left(x_n \cos\left(\phi_n\right)- y_n \sin\left(\phi_n\right)\right),\\
y_{n+1} = \mu\left(x_n \sin(\phi_n) + y_n \cos(\phi_n) \right),
\label{eqn:ikeda}
\end{gathered}
\end{equation}
where $\phi_n = 0.4 - 6\left(1 + x_n^2 + y_n^2\right)^{-1}$ and bifurcation parameter $\mu = 0.9$.
KANs have been shown to model the Ikeda map effectively~\citep{PhysRevResearch.7.023037,Bagrow_2025}.

The second system is a continuous-time \textit{three-species ecosystem}:
\begin{equation}
\begin{gathered}
\frac{dN}{dt} = N\left(1 - \frac{N}{K}\right) - x_p y_p \frac{NP}{N + N_0}, \\
\frac{dP}{dt} = x_p P\left(y_p \frac{N}{N + N_0} - 1\right) - x_q y_q \frac{PQ}{P + P_0}, \\
\frac{dQ}{dt} = x_q Q\left(y_q \frac{P}{P + P_0} - 1\right),
\end{gathered}
\label{eqn:food}
\end{equation}
where $N$, $P$, and $Q$ represent primary producer, herbivore, and carnivore populations, with carrying capacity $K$ serving as the bifurcation parameter.
We set $K = 0.98$, $x_p = 0.4$, $y_p = 2.009$, $x_q = 0.08$, $y_q = 2.876$, $N_0 = 0.16129$, and $P_0 = 0.5$ to produce chaotic dynamics~\citep{mccann1994nonlinear}.

Data for both systems were generated following \citet{PhysRevResearch.7.023037}; the partitions used here are specified in \methods{}.
We compare KAN and S2KAN over ten initialization seeds per configuration. The S2KAN dictionaries combine generic primitives with rational terms motivated by the generating equations; these experiments therefore assess learning with a domain-informed expression language. Architectures, dictionaries, training settings, and Lyapunov-estimation procedures are given in \methods{}.

For the ecosystem only, the small changes between consecutive training observations motivate a ``residual'' formulation, $\hat{\mathbf{x}}_{n+1}=\mathbf{x}_n+f_\theta(\mathbf{x}_n)$, where $f_\theta$ learns the state increment.
This parameterization requires no knowledge of the governing equations; the same fixed identity connection, which adds no trainable parameters, is applied to both KAN and S2KAN.
Ikeda models predict the next state directly.

Table~\ref{tab:ikeda_ecosystem} summarizes the tradeoff between model size, predictive accuracy, and dynamical fidelity.
S2KAN produces smaller models, with median symbolic fidelity meeting or exceeding KAN's in 34 of 36 comparisons (94\%).
Preserving long-run dynamics is more sensitive to regularization for the ecosystem than for Ikeda.
Appendix~\ref{sec:ds-diagnostics} reports full Lyapunov spectra, multi-step prediction errors and forecast horizons, state-distribution overlap, and autocorrelation and power spectral density comparisons.

\begin{table*}[!t]
\centering
\caption{Model compression in chaotic dynamical systems. 
S2KAN yields smaller models with generally higher SF and lower one-step RMSE for Ikeda; stronger compression trades accuracy for size in the ecosystem. 
Entries are medians over ten initialization seeds; brackets are IQRs (omitted for constant values). 
The model-size measure $k$ is defined in Eq.~\eqref{eqn:effective-parameter-count}. 
$\lambda_1$ is the largest Lyapunov exponent (per iterate for Ikeda; per unit time for the ecosystem). 
}
\label{tab:ikeda_ecosystem}
\begingroup
\small
\setlength{\tabcolsep}{3pt}
\renewcommand{\arraystretch}{1.13}
\begin{adjustbox}{max width=\linewidth}
\begin{tabular}{@{}lrrrrrrrrrrr@{}}
\toprule
 & & Test accuracy & Dynamics & \multicolumn{2}{c}{Model size} & \multicolumn{3}{c}{SF (\%), $\epsilon=0.01$} & \multicolumn{3}{c}{SF (\%), $\epsilon=0.1$} \\
\cmidrule(lr){3-3}\cmidrule(lr){4-4}\cmidrule(lr){5-6}\cmidrule(lr){7-9}\cmidrule(lr){10-12}
Model & $\beta$ & RMSE & $\lambda_1$ & Active & $\ktot$ & $J=1$ & $J=2$ & $J=3$ & $J=1$ & $J=2$ & $J=3$ \\
\midrule
\multicolumn{12}{@{}l}{\textbf{Ikeda map}: RMSE $\times10^{-3}$; reference $\lambda_1=0.508$} \\
\addlinespace[2pt]
KAN & --- & \shortstack[r]{14.10\\[-1pt]{\fontsize{6}{7}\selectfont [10.74, 18.42]}} & \shortstack[r]{0.508\\[-1pt]{\fontsize{6}{7}\selectfont [0.505, 0.511]}} & \shortstack[r]{48\\[-1pt]{\fontsize{6}{7}\selectfont \vphantom{[0.0, 0.0]}}} & \shortstack[r]{720.0\\[-1pt]{\fontsize{6}{7}\selectfont \vphantom{[0.0, 0.0]}}} & \shortstack[r]{10.1\\[-1pt]{\fontsize{6}{7}\selectfont [4.9, 12.2]}} & \shortstack[r]{27.7\\[-1pt]{\fontsize{6}{7}\selectfont [26.4, 40.3]}} & \shortstack[r]{49.4\\[-1pt]{\fontsize{6}{7}\selectfont [44.0, 67.5]}} & \shortstack[r]{56.7\\[-1pt]{\fontsize{6}{7}\selectfont [52.7, 62.7]}} & \shortstack[r]{86.1\\[-1pt]{\fontsize{6}{7}\selectfont [81.8, 88.9]}} & \shortstack[r]{97.6\\[-1pt]{\fontsize{6}{7}\selectfont [95.9, 99.4]}} \\
S2KAN & 0 & \shortstack[r]{4.65\\[-1pt]{\fontsize{6}{7}\selectfont [3.83, 5.09]}} & \shortstack[r]{0.508\\[-1pt]{\fontsize{6}{7}\selectfont [0.506, 0.512]}} & \shortstack[r]{45\\[-1pt]{\fontsize{6}{7}\selectfont [43.0, 46.0]}} & \shortstack[r]{652.5\\[-1pt]{\fontsize{6}{7}\selectfont [634.6, 697.6]}} & \shortstack[r]{6.5\\[-1pt]{\fontsize{6}{7}\selectfont [5.1, 10.1]}} & \shortstack[r]{25.7\\[-1pt]{\fontsize{6}{7}\selectfont [21.4, 30.3]}} & \shortstack[r]{59.2\\[-1pt]{\fontsize{6}{7}\selectfont [55.9, 64.6]}} & \shortstack[r]{60.3\\[-1pt]{\fontsize{6}{7}\selectfont [56.9, 62.6]}} & \shortstack[r]{89.6\\[-1pt]{\fontsize{6}{7}\selectfont [82.9, 95.9]}} & \shortstack[r]{100.0\\[-1pt]{\fontsize{6}{7}\selectfont [95.0, 100.0]}} \\
S2KAN & 1 & \shortstack[r]{3.17\\[-1pt]{\fontsize{6}{7}\selectfont [2.60, 3.61]}} & \shortstack[r]{0.508\\[-1pt]{\fontsize{6}{7}\selectfont [0.501, 0.512]}} & \shortstack[r]{41.5\\[-1pt]{\fontsize{6}{7}\selectfont [39.5, 42.0]}} & \shortstack[r]{287.5\\[-1pt]{\fontsize{6}{7}\selectfont [285.5, 313.0]}} & \shortstack[r]{29.7\\[-1pt]{\fontsize{6}{7}\selectfont [25.8, 31.6]}} & \shortstack[r]{66.6\\[-1pt]{\fontsize{6}{7}\selectfont [62.6, 68.6]}} & \shortstack[r]{86.5\\[-1pt]{\fontsize{6}{7}\selectfont [86.1, 90.7]}} & \shortstack[r]{67.9\\[-1pt]{\fontsize{6}{7}\selectfont [66.9, 70.1]}} & \shortstack[r]{95.8\\[-1pt]{\fontsize{6}{7}\selectfont [94.6, 99.3]}} & \shortstack[r]{100.0\\[-1pt]{\fontsize{6}{7}\selectfont [100.0, 100.0]}} \\
S2KAN & 3 & \shortstack[r]{5.18\\[-1pt]{\fontsize{6}{7}\selectfont [3.33, 8.47]}} & \shortstack[r]{0.507\\[-1pt]{\fontsize{6}{7}\selectfont [0.502, 0.508]}} & \shortstack[r]{40\\[-1pt]{\fontsize{6}{7}\selectfont [38.2, 40.8]}} & \shortstack[r]{186.0\\[-1pt]{\fontsize{6}{7}\selectfont [176.5, 212.2]}} & \shortstack[r]{38.3\\[-1pt]{\fontsize{6}{7}\selectfont [36.9, 43.9]}} & \shortstack[r]{78.8\\[-1pt]{\fontsize{6}{7}\selectfont [74.6, 82.4]}} & \shortstack[r]{93.1\\[-1pt]{\fontsize{6}{7}\selectfont [91.2, 94.2]}} & \shortstack[r]{81.6\\[-1pt]{\fontsize{6}{7}\selectfont [77.5, 83.7]}} & \shortstack[r]{100.0\\[-1pt]{\fontsize{6}{7}\selectfont [97.1, 100.0]}} & \shortstack[r]{100.0\\[-1pt]{\fontsize{6}{7}\selectfont \vphantom{[0.0, 0.0]}}} \\
\addlinespace[5pt]
\multicolumn{12}{@{}l}{\textbf{Ecosystem}: RMSE $\times10^{-5}$; reference $\lambda_1=0.00814$} \\
\addlinespace[2pt]
KAN & --- & \shortstack[r]{5.00\\[-1pt]{\fontsize{6}{7}\selectfont [3.68, 6.68]}} & \shortstack[r]{0.00666\\[-1pt]{\fontsize{6}{7}\selectfont [0.00092, 0.00861]}} & \shortstack[r]{27\\[-1pt]{\fontsize{6}{7}\selectfont \vphantom{[0.0, 0.0]}}} & \shortstack[r]{405.0\\[-1pt]{\fontsize{6}{7}\selectfont \vphantom{[0.0, 0.0]}}} & \shortstack[r]{18.9\\[-1pt]{\fontsize{6}{7}\selectfont [11.7, 25.3]}} & \shortstack[r]{66.7\\[-1pt]{\fontsize{6}{7}\selectfont [51.4, 81.7]}} & \shortstack[r]{87.8\\[-1pt]{\fontsize{6}{7}\selectfont [85.2, 91.5]}} & \shortstack[r]{83.3\\[-1pt]{\fontsize{6}{7}\selectfont [79.6, 84.6]}} & \shortstack[r]{100.0\\[-1pt]{\fontsize{6}{7}\selectfont [95.9, 100.0]}} & \shortstack[r]{100.0\\[-1pt]{\fontsize{6}{7}\selectfont [100.0, 100.0]}} \\
S2KAN & 0 & \shortstack[r]{5.08\\[-1pt]{\fontsize{6}{7}\selectfont [4.87, 5.63]}} & \shortstack[r]{0.00688\\[-1pt]{\fontsize{6}{7}\selectfont [0.00002, 0.00926]}} & \shortstack[r]{25\\[-1pt]{\fontsize{6}{7}\selectfont [24.0, 25.8]}} & \shortstack[r]{288.5\\[-1pt]{\fontsize{6}{7}\selectfont [267.5, 312.5]}} & \shortstack[r]{51.2\\[-1pt]{\fontsize{6}{7}\selectfont [47.7, 61.5]}} & \shortstack[r]{86.3\\[-1pt]{\fontsize{6}{7}\selectfont [85.0, 90.7]}} & \shortstack[r]{95.5\\[-1pt]{\fontsize{6}{7}\selectfont [95.0, 100.0]}} & \shortstack[r]{95.0\\[-1pt]{\fontsize{6}{7}\selectfont [90.9, 95.5]}} & \shortstack[r]{100.0\\[-1pt]{\fontsize{6}{7}\selectfont [100.0, 100.0]}} & \shortstack[r]{100.0\\[-1pt]{\fontsize{6}{7}\selectfont \vphantom{[0.0, 0.0]}}} \\
S2KAN & 0.1 & \shortstack[r]{8.79\\[-1pt]{\fontsize{6}{7}\selectfont [7.91, 9.19]}} & \shortstack[r]{0.00664\\[-1pt]{\fontsize{6}{7}\selectfont [-0.00002, 0.00909]}} & \shortstack[r]{23\\[-1pt]{\fontsize{6}{7}\selectfont [22.0, 24.8]}} & \shortstack[r]{177.5\\[-1pt]{\fontsize{6}{7}\selectfont [173.9, 190.7]}} & \shortstack[r]{51.3\\[-1pt]{\fontsize{6}{7}\selectfont [50.0, 54.5]}} & \shortstack[r]{88.6\\[-1pt]{\fontsize{6}{7}\selectfont [75.8, 93.2]}} & \shortstack[r]{94.3\\[-1pt]{\fontsize{6}{7}\selectfont [90.2, 100.0]}} & \shortstack[r]{92.5\\[-1pt]{\fontsize{6}{7}\selectfont [90.2, 94.7]}} & \shortstack[r]{100.0\\[-1pt]{\fontsize{6}{7}\selectfont \vphantom{[0.0, 0.0]}}} & \shortstack[r]{100.0\\[-1pt]{\fontsize{6}{7}\selectfont \vphantom{[0.0, 0.0]}}} \\
S2KAN & 1 & \shortstack[r]{27.14\\[-1pt]{\fontsize{6}{7}\selectfont [23.93, 38.59]}} & \shortstack[r]{0.00001\textsuperscript{a}\\[-1pt]{\fontsize{6}{7}\selectfont [-0.00003, 0.00768]}} & \shortstack[r]{22.5\\[-1pt]{\fontsize{6}{7}\selectfont [22.0, 23.8]}} & \shortstack[r]{87.5\\[-1pt]{\fontsize{6}{7}\selectfont [76.0, 88.7]}} & \shortstack[r]{68.2\\[-1pt]{\fontsize{6}{7}\selectfont [64.8, 71.7]}} & \shortstack[r]{93.0\\[-1pt]{\fontsize{6}{7}\selectfont [90.9, 98.9]}} & \shortstack[r]{100.0\\[-1pt]{\fontsize{6}{7}\selectfont [93.2, 100.0]}} & \shortstack[r]{100.0\\[-1pt]{\fontsize{6}{7}\selectfont [95.5, 100.0]}} & \shortstack[r]{100.0\\[-1pt]{\fontsize{6}{7}\selectfont [100.0, 100.0]}} & \shortstack[r]{100.0\\[-1pt]{\fontsize{6}{7}\selectfont \vphantom{[0.0, 0.0]}}} \\
\bottomrule
\end{tabular}
 \end{adjustbox}
\endgroup
\par\smallskip
\begin{minipage}{\linewidth}
\footnotesize\raggedright{}
\textsuperscript{a} $\lambda_1$ uses 9/10 fits for the ecosystem at $\beta=1$; all other entries use 10/10. Undefined Lyapunov estimates are excluded from medians and IQRs.
 \end{minipage}
\end{table*}

At $\beta=1$, S2KAN reduces the median effective parameter count from 720 to 287.5 for Ikeda and from 405 to 87.5 for the ecosystem, while matching or exceeding KAN's median SF at all six settings.
Median one-step RMSE decreases from $1.41\times10^{-2}$ to $3.17\times10^{-3}$ for Ikeda, but increases from $5.00\times10^{-5}$ to $2.71\times10^{-4}$ for the ecosystem.
For Ikeda, this combines 60\% fewer effective parameters with a median leading exponent close to the reference (0.5076 versus 0.5081), although the model from one of ten seeds converges to incorrect long-run behavior.
For the ecosystem, the median leading Lyapunov exponent falls to approximately zero, substantially below the reference value of $0.00814$.

Weaker regularization improves the ecosystem's prediction accuracy and median leading-exponent estimate, although recovery varies across seeds.
Even without an explicit sparsity penalty ($\beta=0$), S2KAN's median effective parameter count is 9\% below KAN for Ikeda and 29\% below KAN for the ecosystem, with lower one-step error for Ikeda and nearly unchanged error for the ecosystem.
Appendix~\ref{subsec:self-sparsification} shows the corresponding reductions in parameter count during training.

\subsection{Nguyen symbolic regression benchmark}

Next, we turn to the Nguyen symbolic regression benchmark~\citep{uy2011semantically}. 
This benchmark contains a sample of mathematical functions designed to capture a variety of symbolic regression challenges.
We focus in Table~\ref{tab:nguyen-full} on the first 10, the last two of which are bivariate.
For each problem we generate 1024 training and 256 testing points.
We apply three different architectures: small (S), with no hidden units; large (L), with one hidden layer of 3 summation units; and large-mult (LM), with one hidden layer of 3 summation and 1 multiplication units.
Models were trained for up to 10,000 epochs with a batch size of 128. S2KAN's first 200 epochs were warmup ($\beta=0$).
We report medians over ten initialization seeds on a fixed training/test sample; Appendix~\ref{sec:nguyen-variability} reports the corresponding quartiles.
Table~\ref{tab:nguyen-full} reports test $R^2$, RMSE, model size, and symbolic fidelity for both methods, using $\beta=1$ for S2KAN.

Across 180 comparisons of median symbolic fidelity (10 problems $\times$ 3 architectures $\times$ 6 fidelity settings), S2KAN exceeds KAN in 114 (63.3\%), ties in 66 (36.7\%), and loses in none; excluding the small F10 pair, whose S2KAN fits have no functionally active edges and receive 100\% fidelity by convention, gives 108 wins out of 174 comparisons (62.1\%), with 66 ties and no losses. These counts are descriptive, since the six fidelity settings are not independent.

Baseline KAN meets the median-RMSE threshold of 1\% of test-target SD in 29 of 30 problem--architecture configurations, compared with 23 for S2KAN at $\beta=1$. Across the 30 configuration medians, median normalized RMSE is 0.040\% of target SD for KAN and 0.44\% for S2KAN. Both methods fail on F10 with the small architecture; S2KAN additionally exceeds the threshold on F5, F7 and F8 with L and LM. The largest departures occur on F5, where S2KAN's median test $R^2$ is 0.9517 for L and 0.8041 for LM; the latter has an interquartile range of [0.2745, 0.9171], indicating substantial variability across initializations.
Compression is architecture-dependent.
Across the ten problem-level medians, S2KAN reduces the median effective parameter count from 90.0 to 46.5 for L and from 135.0 to 62.8 for LM, but increases it from 15.0 to 19.8 for S.

Appendix~\ref{sec:eql-like-control} tests the contribution of the dense branch by disabling it throughout training and inference.
Appendix~\ref{sec:beta-sensitivity} reports the corresponding accuracy--size sensitivity to $\beta$ across all three Nguyen architectures.

\begin{table}[p]
\centering\singlespacing{}
\caption{Nguyen benchmarks across architectures. 
S2KAN generally retains high predictive accuracy, with the largest departures from baseline on F5 for the large and large-mult architectures. Values are medians over ten initialization seeds; quartiles are reported in Table~\ref{tab:nguyen-uncertainty}. Y and N indicate whether RMSE is at most 1\% of the test-target standard deviation (SD). 
Shapes: S$=[n_0,1]$, L$=[n_0,3,1]$, LM$=[n_0,(3,1),1]$ with multiplicative interactions.
The model-size measure $k$ is defined in Eq.~\eqref{eqn:effective-parameter-count}. }
\label{tab:nguyen-full}
\begingroup\small\setlength{\tabcolsep}{3pt}\renewcommand{\arraystretch}{1.02}
\begin{adjustbox}{max width=\linewidth,max totalheight=0.88\textheight}
\begin{tabular}{clllrrrrrrrrrr}
\toprule
 & & & & \multicolumn{2}{c}{Test accuracy} & \multicolumn{2}{c}{Model size} & \multicolumn{3}{c}{SF (\%), $\epsilon=0.01$} & \multicolumn{3}{c}{SF (\%), $\epsilon=0.1$} \\
\cmidrule(lr){5-6}\cmidrule(lr){7-8}\cmidrule(lr){9-11}\cmidrule(lr){12-14}
ID & Expression / domain & Shape & Model & $R^2$ & RMSE ($\leq1\%$ SD) & Active & $k$ & $J=1$ & $J=2$ & $J=3$ & $J=1$ & $J=2$ & $J=3$ \\
\midrule
F1 & $x^3+x^2+x$ & S & KAN & 1.0000 & $5.41\times10^{-6}$ (Y) & 1 & 15.0 & 0.0 & 100.0 & 100.0 & 0.0 & 100.0 & 100.0 \\
 & $x\in[-1,1]$ &  & S2KAN & 1.0000 & $8.06\times10^{-4}$ (Y) & 1 & 20.1 & 0.0 & 100.0 & 100.0 & 0.0 & 100.0 & 100.0 \\
 &  & L & KAN & 1.0000 & $5.50\times10^{-4}$ (Y) & 6 & 90.0 & 0.0 & 0.0 & 16.7 & 33.3 & 75.0 & 100.0 \\
 &  &  & S2KAN & 1.0000 & $5.46\times10^{-3}$ (Y) & 6 & 42.5 & 40.0 & 80.0 & 100.0 & 60.0 & 100.0 & 100.0 \\
 &  & LM & KAN & 1.0000 & $4.36\times10^{-4}$ (Y) & 9 & 135.0 & 0.0 & 0.0 & 12.5 & 35.4 & 76.4 & 88.2 \\
 &  &  & S2KAN & 1.0000 & $4.18\times10^{-3}$ (Y) & 9 & 65.9 & 38.1 & 100.0 & 100.0 & 57.1 & 100.0 & 100.0 \\
\midrule
F2 & $x^4+x^3+x^2+x$ & S & KAN & 1.0000 & $1.25\times10^{-4}$ (Y) & 1 & 15.0 & 0.0 & 100.0 & 100.0 & 0.0 & 100.0 & 100.0 \\
 & $x\in[-1,1]$ &  & S2KAN & 1.0000 & $8.25\times10^{-4}$ (Y) & 1 & 19.2 & 0.0 & 100.0 & 100.0 & 0.0 & 100.0 & 100.0 \\
 &  & L & KAN & 1.0000 & $4.60\times10^{-4}$ (Y) & 6 & 90.0 & 0.0 & 16.7 & 50.0 & 41.7 & 91.7 & 100.0 \\
 &  &  & S2KAN & 1.0000 & $6.26\times10^{-3}$ (Y) & 6 & 41.4 & 60.0 & 100.0 & 100.0 & 100.0 & 100.0 & 100.0 \\
 &  & LM & KAN & 1.0000 & $4.50\times10^{-4}$ (Y) & 9 & 135.0 & 0.0 & 5.6 & 33.3 & 35.4 & 75.0 & 94.4 \\
 &  &  & S2KAN & 1.0000 & $5.25\times10^{-3}$ (Y) & 9 & 46.6 & 63.3 & 100.0 & 100.0 & 83.3 & 100.0 & 100.0 \\
\midrule
F3 & $x^5+x^4+x^3+x^2+x$ & S & KAN & 1.0000 & $2.43\times10^{-4}$ (Y) & 1 & 15.0 & 0.0 & 0.0 & 100.0 & 0.0 & 100.0 & 100.0 \\
 & $x\in[-1,1]$ &  & S2KAN & 1.0000 & $4.94\times10^{-4}$ (Y) & 1 & 20.1 & 0.0 & 0.0 & 100.0 & 0.0 & 100.0 & 100.0 \\
 &  & L & KAN & 1.0000 & $7.47\times10^{-4}$ (Y) & 6 & 90.0 & 0.0 & 0.0 & 25.0 & 33.3 & 75.0 & 100.0 \\
 &  &  & S2KAN & 1.0000 & $6.16\times10^{-3}$ (Y) & 6 & 55.9 & 40.0 & 80.0 & 80.0 & 60.0 & 100.0 & 100.0 \\
 &  & LM & KAN & 1.0000 & $6.24\times10^{-4}$ (Y) & 9 & 135.0 & 0.0 & 5.6 & 12.5 & 23.6 & 64.6 & 88.9 \\
 &  &  & S2KAN & 1.0000 & $4.97\times10^{-3}$ (Y) & 9 & 66.6 & 35.4 & 92.9 & 100.0 & 50.0 & 100.0 & 100.0 \\
\midrule
F4 & $x^6+x^5+x^4+x^3+x^2+x$ & S & KAN & 1.0000 & $5.61\times10^{-4}$ (Y) & 1 & 15.0 & 0.0 & 0.0 & 100.0 & 0.0 & 100.0 & 100.0 \\
 & $x\in[-1,1]$ &  & S2KAN & 1.0000 & $7.55\times10^{-4}$ (Y) & 1 & 19.5 & 0.0 & 0.0 & 100.0 & 0.0 & 100.0 & 100.0 \\
 &  & L & KAN & 1.0000 & $7.91\times10^{-4}$ (Y) & 6 & 90.0 & 0.0 & 0.0 & 26.7 & 20.0 & 63.3 & 91.7 \\
 &  &  & S2KAN & 1.0000 & $7.81\times10^{-3}$ (Y) & 6 & 50.5 & 50.0 & 80.0 & 91.7 & 77.5 & 100.0 & 100.0 \\
 &  & LM & KAN & 1.0000 & $5.89\times10^{-4}$ (Y) & 9 & 135.0 & 0.0 & 12.5 & 25.0 & 37.5 & 64.6 & 87.5 \\
 &  &  & S2KAN & 1.0000 & $5.54\times10^{-3}$ (Y) & 9 & 62.8 & 28.6 & 100.0 & 100.0 & 57.1 & 100.0 & 100.0 \\
\midrule
F5 & $\sin(x^2)\cos(x)-1$ & S & KAN & 1.0000 & $5.11\times10^{-5}$ (Y) & 1 & 15.0 & 0.0 & 100.0 & 100.0 & 100.0 & 100.0 & 100.0 \\
 & $x\in[-1,1]$ &  & S2KAN & 1.0000 & $9.01\times10^{-4}$ (Y) & 1 & 19.2 & 0.0 & 100.0 & 100.0 & 100.0 & 100.0 & 100.0 \\
 &  & L & KAN & 1.0000 & $1.77\times10^{-4}$ (Y) & 6 & 90.0 & 8.3 & 29.2 & 36.7 & 45.0 & 66.7 & 100.0 \\
 &  &  & S2KAN & 0.9517 & $3.52\times10^{-2}$ (N) & 6 & 58.2 & 75.0 & 100.0 & 100.0 & 100.0 & 100.0 & 100.0 \\
 &  & LM & KAN & 1.0000 & $1.26\times10^{-4}$ (Y) & 9 & 135.0 & 5.6 & 12.5 & 25.0 & 29.2 & 64.6 & 88.2 \\
 &  &  & S2KAN & 0.8041 & $7.34\times10^{-2}$ (N) & 8.5 & 73.2 & 70.8 & 100.0 & 100.0 & 91.7 & 100.0 & 100.0 \\
\midrule
F6 & $\sin(x)+\sin(x+x^2)$ & S & KAN & 1.0000 & $8.28\times10^{-5}$ (Y) & 1 & 15.0 & 0.0 & 100.0 & 100.0 & 100.0 & 100.0 & 100.0 \\
 & $x\in[-1,1]$ &  & S2KAN & 1.0000 & $9.55\times10^{-4}$ (Y) & 1 & 23.6 & 0.0 & 100.0 & 100.0 & 100.0 & 100.0 & 100.0 \\
 &  & L & KAN & 1.0000 & $2.85\times10^{-4}$ (Y) & 6 & 90.0 & 0.0 & 18.3 & 45.0 & 66.7 & 100.0 & 100.0 \\
 &  &  & S2KAN & 1.0000 & $2.95\times10^{-3}$ (Y) & 6 & 40.9 & 80.0 & 100.0 & 100.0 & 100.0 & 100.0 & 100.0 \\
 &  & LM & KAN & 1.0000 & $3.59\times10^{-4}$ (Y) & 9 & 135.0 & 0.0 & 11.8 & 33.3 & 37.5 & 77.8 & 94.4 \\
 &  &  & S2KAN & 1.0000 & $6.08\times10^{-3}$ (Y) & 9 & 62.7 & 71.4 & 100.0 & 100.0 & 85.7 & 100.0 & 100.0 \\
\midrule
F7 & $\log(x+1)+\log(x^2+1)$ & S & KAN & 1.0000 & $1.03\times10^{-5}$ (Y) & 1 & 15.0 & 100.0 & 100.0 & 100.0 & 100.0 & 100.0 & 100.0 \\
 & $x\in[0,2]$ &  & S2KAN & 1.0000 & $2.30\times10^{-3}$ (Y) & 1 & 20.7 & 100.0 & 100.0 & 100.0 & 100.0 & 100.0 & 100.0 \\
 &  & L & KAN & 1.0000 & $3.07\times10^{-4}$ (Y) & 6 & 90.0 & 0.0 & 8.3 & 41.7 & 50.0 & 83.3 & 100.0 \\
 &  &  & S2KAN & 0.9994 & $1.96\times10^{-2}$ (N) & 6 & 56.5 & 55.0 & 100.0 & 100.0 & 80.0 & 100.0 & 100.0 \\
 &  & LM & KAN & 1.0000 & $3.60\times10^{-4}$ (Y) & 9 & 135.0 & 0.0 & 5.6 & 25.0 & 37.5 & 81.9 & 100.0 \\
 &  &  & S2KAN & 0.9963 & $5.02\times10^{-2}$ (N) & 9 & 72.7 & 58.6 & 100.0 & 100.0 & 80.4 & 100.0 & 100.0 \\
\midrule
F8 & $\sqrt{x}$ & S & KAN & 1.0000 & $1.83\times10^{-3}$ (Y) & 1 & 15.0 & 0.0 & 0.0 & 0.0 & 100.0 & 100.0 & 100.0 \\
 & $x\in[0,4]$ &  & S2KAN & 1.0000 & $2.05\times10^{-3}$ (Y) & 1 & 19.1 & 0.0 & 0.0 & 0.0 & 100.0 & 100.0 & 100.0 \\
 &  & L & KAN & 1.0000 & $3.51\times10^{-4}$ (Y) & 6 & 90.0 & 0.0 & 16.7 & 33.3 & 50.0 & 83.3 & 100.0 \\
 &  &  & S2KAN & 0.9991 & $1.41\times10^{-2}$ (N) & 6 & 51.8 & 41.7 & 77.5 & 100.0 & 63.3 & 100.0 & 100.0 \\
 &  & LM & KAN & 1.0000 & $1.90\times10^{-4}$ (Y) & 9 & 135.0 & 0.0 & 5.6 & 19.6 & 35.4 & 76.4 & 88.2 \\
 &  &  & S2KAN & 0.9993 & $1.25\times10^{-2}$ (N) & 9 & 61.5 & 34.3 & 80.0 & 100.0 & 66.7 & 100.0 & 100.0 \\
\midrule
F9 & $\sin(x)+\sin(y^2)$ & S & KAN & 1.0000 & $6.75\times10^{-5}$ (Y) & 2 & 30.0 & 50.0 & 100.0 & 100.0 & 100.0 & 100.0 & 100.0 \\
 & $x,y\in[-1,1]$ &  & S2KAN & 1.0000 & $1.03\times10^{-3}$ (Y) & 2 & 34.4 & 50.0 & 100.0 & 100.0 & 100.0 & 100.0 & 100.0 \\
 &  & L & KAN & 1.0000 & $2.13\times10^{-4}$ (Y) & 9 & 135.0 & 7.1 & 34.3 & 58.6 & 67.0 & 93.8 & 100.0 \\
 &  &  & S2KAN & 1.0000 & $3.73\times10^{-3}$ (Y) & 9 & 41.9 & 93.8 & 100.0 & 100.0 & 100.0 & 100.0 & 100.0 \\
 &  & LM & KAN & 1.0000 & $3.41\times10^{-4}$ (Y) & 14 & 210.0 & 0.0 & 8.7 & 45.8 & 53.8 & 96.2 & 100.0 \\
 &  &  & S2KAN & 1.0000 & $3.08\times10^{-3}$ (Y) & 13.5 & 50.2 & 80.0 & 100.0 & 100.0 & 89.4 & 100.0 & 100.0 \\
\midrule
F10 & $2\sin(x)\cos(y)$ & S & KAN & -0.0372 & $1.01\times10^{0}$ (N) & 2 & 30.0 & 0.0 & 0.0 & 0.0 & 0.0 & 0.0 & 0.0 \\
 & $x,y\in[-\pi,\pi]$ &  & S2KAN\textsuperscript{a} & -0.0006 & $9.90\times10^{-1}$ (N) & 0 & 0.0 & 100.0 & 100.0 & 100.0 & 100.0 & 100.0 & 100.0 \\
 &  & L & KAN & 1.0000 & $5.51\times10^{-3}$ (Y) & 9 & 135.0 & 0.0 & 43.7 & 77.8 & 44.4 & 94.4 & 100.0 \\
 &  &  & S2KAN & 1.0000 & $3.73\times10^{-3}$ (Y) & 9 & 22.0 & 86.6 & 100.0 & 100.0 & 94.4 & 100.0 & 100.0 \\
 &  & LM & KAN & 1.0000 & $3.31\times10^{-3}$ (Y) & 14 & 210.0 & 0.0 & 14.3 & 50.0 & 37.1 & 78.6 & 96.4 \\
 &  &  & S2KAN & 1.0000 & $1.85\times10^{-3}$ (Y) & 10 & 27.2 & 85.7 & 100.0 & 100.0 & 100.0 & 100.0 & 100.0 \\
\midrule
 & Median F1--F10 & S & KAN & 1.0000 & $1.04\times10^{-4}$ & 1 & 15.0 & 0.0 & 100.0 & 100.0 & 50.0 & 100.0 & 100.0 \\
 &  &  & S2KAN & 1.0000 & $9.28\times10^{-4}$ & 1 & 19.8 & 0.0 & 100.0 & 100.0 & 100.0 & 100.0 & 100.0 \\
 &  & L & KAN & 1.0000 & $4.05\times10^{-4}$ & 6 & 90.0 & 0.0 & 16.7 & 39.2 & 44.7 & 83.3 & 100.0 \\
 &  &  & S2KAN & 1.0000 & $6.21\times10^{-3}$ & 6 & 46.5 & 57.5 & 100.0 & 100.0 & 87.2 & 100.0 & 100.0 \\
 &  & LM & KAN & 1.0000 & $3.98\times10^{-4}$ & 9 & 135.0 & 0.0 & 7.1 & 25.0 & 36.3 & 76.4 & 91.7 \\
 &  &  & S2KAN & 1.0000 & $5.40\times10^{-3}$ & 9 & 62.8 & 61.0 & 100.0 & 100.0 & 81.8 & 100.0 & 100.0 \\
\bottomrule
\end{tabular}
 \end{adjustbox}
\endgroup
\par\smallskip
\begin{minipage}{\linewidth}
\footnotesize\raggedright{}
\textsuperscript{a} All ten fits have no functionally active edges; SF is 100\% by convention. This does not imply an accurate model.
 \end{minipage}
\end{table}

\section{Learning dynamics}
\label{sec:learning-dynamics}

To understand how S2KAN discovers sparse representations during training, we performed an experiment to track gate statistics throughout optimization.
For each gate with learnable parameter $\alpha_i$, $p_i$ is the probability that the gate is nonzero (Eq.~\eqref{eqn:expected-z}).
Using these, we compute the total binary entropy across all gates in the network:
\begin{equation}
H = -\sum_{m }\left[p_m \log_2 p_m + (1-p_m) \log_2(1-p_m)\right],
\end{equation}
which measures the total uncertainty in the gate configuration.
When all gates are fully decided ($p_m \approx 0$ or $p_m \approx 1$), entropy is near zero; when gates are maximally uncertain ($p_m \approx 0.5$), entropy is maximized.
We also report \emph{decisiveness}, the fraction of gates with $p_m < 0.01$ or $p_m > 0.99$, indicating convergence to discrete selection.

\begin{figure*}[t]%
    \centering
    \includegraphics[width=\linewidth]{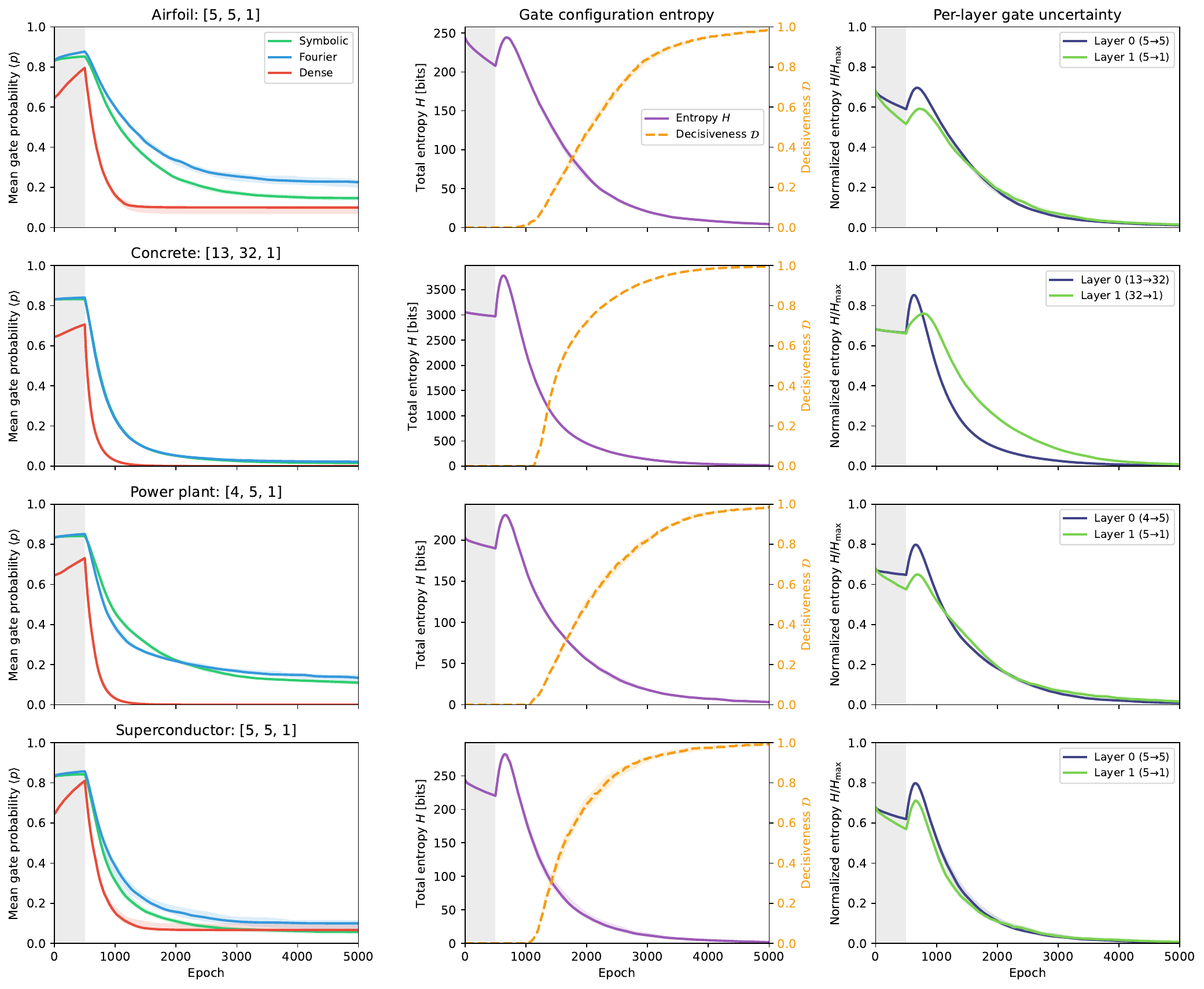}
    \caption{Learning dynamics for S2KAN on four real-world datasets, with $\beta=1$ after 500 warmup epochs. Rows show Airfoil, Concrete, Power plant, and Superconductor. Columns show mean gate-opening probabilities by basis type; total gate entropy and decisiveness; and normalized per-layer entropy $H/H_{\max}$. Dense denotes the spline branch, including its SiLU residual. Curves show medians over ten data splits with three initialization seeds per split; bands show interquartile ranges. Gray shading marks warmup ($\beta=0$). Full 5,000-epoch trajectories are shown\@.
    \label{fig:gate-dynamics}}
\end{figure*}

Figure~\ref{fig:gate-dynamics} shows these statistics for the four real-world datasets using the architectures and dictionaries described in \methods{}.

During warmup (when $\beta=0$), mean dense-gate probabilities increase across all four datasets, while symbolic and Fourier gate probabilities remain high.
After regularization begins ($\beta=1$), gate entropy briefly increases before declining as gates become increasingly decisive.
Dense-gate probabilities decrease more rapidly than those of the symbolic and Fourier terms, approaching zero for Concrete and Power Plant while remaining nonzero for Airfoil and Superconductor.
The per-layer trajectories also show differences in timing: for Concrete, normalized entropy falls more rapidly in the input layer than in the output layer, whereas the two layers follow more similar trajectories on the other datasets.
Together, these observations show a transition from broadly open gates toward sparse, increasingly discrete representations, with the retained activation families and timing of gate decisions varying across datasets.

Model size can also decrease without an explicit sparsity penalty: in the two dynamical systems trained with $\beta=0$ throughout, we observe a decrease in expected effective parameter count during training.
All ten fits per system finish below their first-epoch counts, and the median final count is below baseline KAN in both systems (App.~\ref{subsec:self-sparsification}).
These observations are specific to the tested training configurations; their mechanism and dependence on optimization settings remain unresolved.

\begin{figure*}[tbp]
\begin{center}
\includegraphics[width=\textwidth]{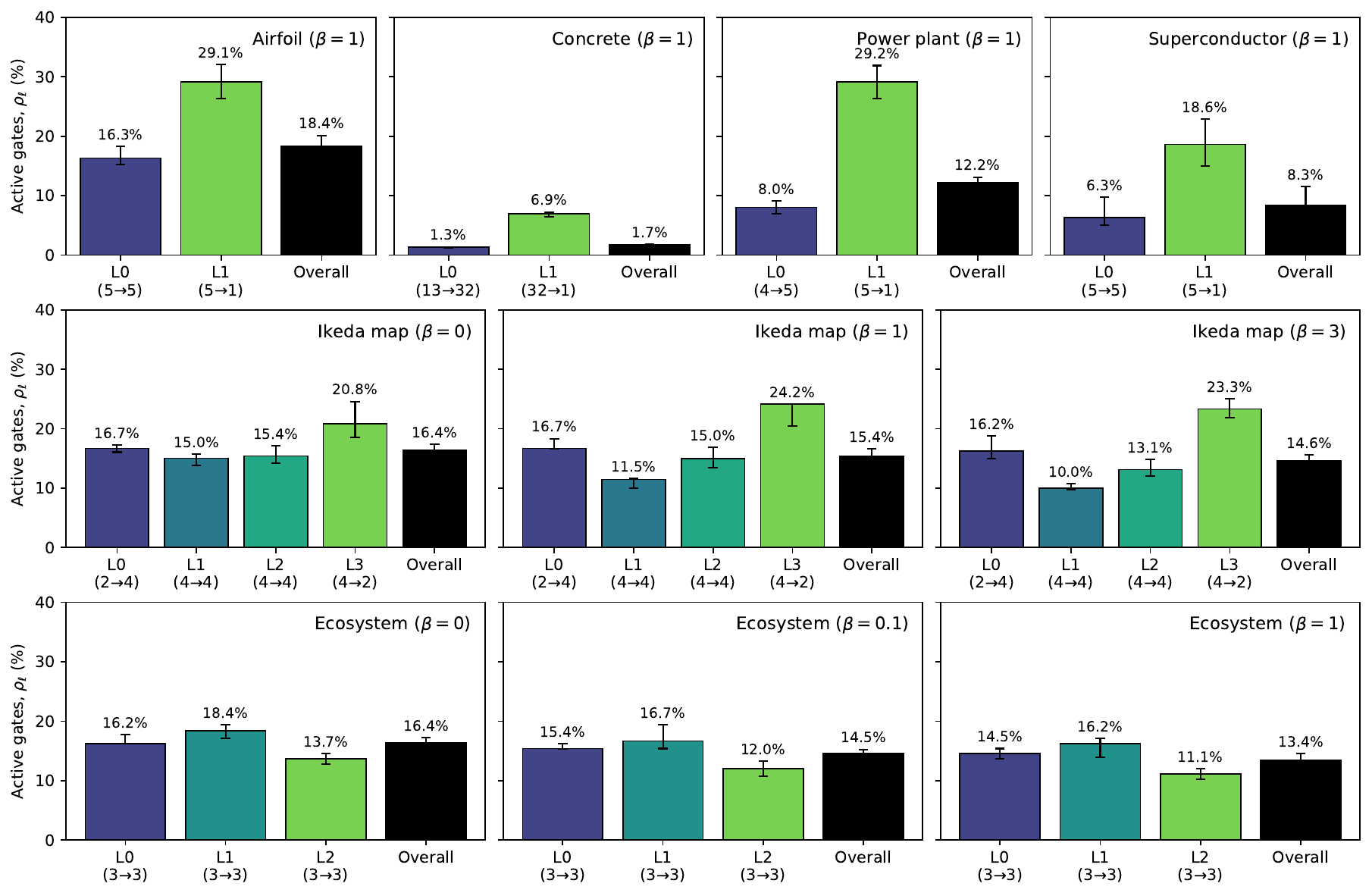}
\caption{Expected fraction of active gates per layer, $\rho_\ell$ (Eq.~\eqref{eqn:active-gate-fraction}), for the real-world and dynamical-systems S2KAN models. 
Bars show medians; whiskers indicate IQRs across ten initialization seeds per dynamical-system configuration and 30 fits per real-world dataset (ten data splits and three initialization seeds per split). 
Overall fractions are computed within each model before aggregation. 
\label{fig:sparsity-bathtub}}
\end{center}
\end{figure*}

To understand how sparsity distributes across network depth, we examined the expected fraction of active gates per layer, $\rho_\ell$, for the multi-layer architectures in our experiments (Fig.~\ref{fig:sparsity-bathtub}):
\begin{equation}
\rho_\ell := \frac{1}{n_\ell n_{\ell+1} M} {\sum_{ijm} p_{\ell ijm}},
\label{eqn:active-gate-fraction}
\end{equation}
where $M = |\mathcal{S}| + |\mathcal{F}| + |\mathcal{R}|$ is the number of dictionary terms per activation, $n_\ell$ and $n_{\ell+1}$ are the input and output widths of layer $\ell$, and $p_{\ell ijm}$ is the probability that the gate is nonzero (Eq.~\eqref{eqn:expected-z}).
Figure~\ref{fig:sparsity-bathtub} compares the four-layer Ikeda, three-layer ecosystem, and two-layer real-world networks.
Ikeda retains a smaller fraction of gates in its interior layers than at its boundaries; the ecosystem instead retains the smallest fraction in its output layer, while the real-world networks retain a larger fraction in the output layer.
The resulting sparsity profiles appear architecture- and dataset-dependent, rather than a universal interior-layer pattern.

\section{Discussion}
\label{sec:discussion}

Across real-world prediction tasks, dynamical-system forecasting, and synthetic symbolic benchmarks, S2KAN generally achieves higher symbolic fidelity and often far smaller models, while maintaining competitive predictive accuracy on real-world data and low prediction errors on most symbolic benchmark configurations.
Differentiable gating and MDL-inspired regularization jointly select symbolic and dense activation components, allowing compact symbolic representations while preserving the option to retain dense terms. 
We now discuss limitations and directions for future work.

Like any symbolic regression method, S2KAN requires a set of candidate functions: PySR~\citep{cranmer2023interpretable}, SINDy~\citep{sindy2016}, and AI Feynman~\citep{udrescu2020ai} likewise require the user to specify operators or candidate terms.
In a KAN, the activation dictionary plays the role of the operator set. The Kolmogorov--Arnold representation shows that addition suffices with sufficiently expressive univariate functions and a suitable architecture; multiplication nodes can provide more compact representations~\citep{liu2024kan2.0}.
When domain knowledge is available, it can guide this choice---physical systems, for instance, suggest trigonometric functions, rational functions, or exponentials depending on the phenomena involved; otherwise a generic dictionary can be used, as in our Nguyen experiments.
Unlike approaches restricted to the supplied symbolic dictionary, S2KAN retains dense representations as a fallback for whatever the dictionary cannot express.
These dense terms may or may not recover the missing symbolic structure: in the sinc ablation (App.~\ref{subsec:misspecification}), a spline learned the omitted $1/x$ almost exactly in three of five seeds, whereas the splines of a baseline KAN never did.
Whether they do can be checked by fitting the learned dense activations with candidate primitives outside the training dictionary, as we did for $1/x$; like symbolic fidelity generally, such a check is relative to the primitives tested.
The fallback typically keeps predictive accuracy high even with a suboptimal library, though not always at baseline level, and strong regularization can suppress it (App.~\ref{subsec:misspecification}).

The symbolic library presents its own practical challenges. 
Terms such as $1/x$, $\log(x)$, or $\exp(x)$ can diverge for certain inputs, destabilizing training. 
Likewise, $\log(x)$ and $\sqrt{x}$ have restricted domains that may be violated during training.
In this light, the standard KAN's reliance on dense representations is a practical strength: its spline and SiLU components avoid the restricted domains and singularities of primitives such as logarithms and reciprocals.
Yet these challenges are not unique to S2KAN: the problem is endemic to symbolic regression methods including neural network approaches like equation learner (EQL) networks~\citep{martius2016extrapolation,pmlr-v80-sahoo18a,werner2021informed}. 
Practical mitigations include input normalization, domain restriction, careful initialization, and replacing problematic terms with protected variants such as $\sqrt{|x|}$ or $\log\left(|x| + \epsilon\right)$. 
The equation learner with division (EQL-div)~\citep{pmlr-v80-sahoo18a} and informed equation learning (iEQL)~\citep{werner2021informed} variants offer a more principled solution, introducing symbolic primitives with learnable cutoffs that prevent divergences and provide smooth out-of-domain behavior; similar modifications could be incorporated into S2KAN.
Beyond content, dictionary size also matters: larger dictionaries increase training overhead. When properly implemented to omit inactive terms, inference evaluates only the retained representation; replacing spline expansions with simple symbolic primitives can therefore reduce evaluation cost.
Appendix~\ref{sec:training-inference-times} reports training and inference times for KAN and S2KAN across the four real-world datasets, with inactive terms omitted during inference.

S2KAN takes an explicit approach to symbolic fidelity, incorporating symbolic primitives directly into the activation function dictionary. 
However, the training instabilities discussed above---divergences, domain violations, the need for protected operators---motivate an alternative: biasing dense representations implicitly toward forms with high symbolic fidelity. 
Rather than embedding symbolic functions explicitly, implicit approaches modify the training process or architecture to encourage learned activations that, while remaining dense, are more amenable to post-hoc symbolic fitting, effectively a symbolic inductive bias. 
Projective KAN~\citep{poole2025projective} exemplifies this direction, using projection-based regularization to bias activations toward higher symbolic fidelity.
Confining ill-behaved terms to the regularization rather than the main computational graph generally makes them easier to isolate and control.
Combining explicit symbolic terms with implicit biases toward symbolic fidelity is a promising direction: such hybrid approaches could encourage symbolic representations while improving stability when explicit symbolic terms are ill-suited.

How does S2KAN fit into the broader context of symbolic regression and related methods? 
Table~\ref{tab:qualitative-comparison} compares S2KAN qualitatively with related methods.
Since our contribution focuses on symbolifying KANs, our empirical evaluation centers on KAN versus S2KAN, although one of our ablations (App.~\ref{sec:eql-like-control}) yields a model resembling EQL.
To our knowledge, S2KAN is the first method to combine gradient-based optimization, compositionality, symbolic learning, and graceful degradation through a trainable spline fallback.
The table also highlights that compositionality is shared by nearly all of these methods, a feature that interacts with depth and readability in ways worth examining.

\begin{table}[t]
\centering
\caption{Qualitative comparison of S2KAN with selected symbolic regression and KAN methods.
\label{tab:qualitative-comparison}}\small
\begin{threeparttable}
\begin{tabular}{lcccc}
\toprule
Method & Gradient-based & Compositional & Symbolic & Graceful degradation \\
\midrule
SINDy \citep{sindy2016}
  & \xmark{} & \xmark{} & \cmark{} & \xmark{} \\
PySR \citep{cranmer2023interpretable}
  & \xmark{} & \cmark{} & \cmark{} & \xmark{} \\
AI Feynman \citep{udrescu2020ai,udrescu2020ai2}
  & \xmark{}\tnote{a} & \cmark{} & \cmark{} & \xmark{} \\
EQL / iEQL \citep{pmlr-v80-sahoo18a,werner2021informed}
  & \cmark{} & \cmark{} & \cmark{} & \xmark{} \\
KAN \citep{liu2025kan}
  & \cmark{} & \cmark{} & \xmark{}\tnote{b} & n/a \\
Projective KAN \citep{poole2025projective}
  & \cmark{} & \cmark{} & \xmark{}\tnote{b} & n/a \\
\midrule
S2KAN (ours), $\mathcal{S}$ or $\mathcal{S}$+$\mathcal{F}$
  & \cmark{} & \cmark{} & \cmark{} & \xmark{} \\
S2KAN (ours), $\mathcal{S}$+$\mathcal{F}$+$\mathcal{R}$
  & \cmark{} & \cmark{} & \cmark{} & \cmark{} \\
\bottomrule
\end{tabular}
\begin{tablenotes}\footnotesize
\item[a] Brute-force search over expressions; uses neural network surrogates internally for decomposition.
\item[b] Post-hoc symbolification available but not part of the training objective.
\end{tablenotes}
\end{threeparttable}
\end{table}

Compositionality is essential for discovering complex expressions, but the number of active paths can grow combinatorially as depth and breadth increase, producing large expressions that may hinder readability.
Sparsity mitigates this to an extent by reducing the number of active paths, but depth (more function compositions) and breadth (more terms per function) interact with sparsity and must be considered together.
S2KAN promotes sparsity, while its distribution across layers appears to depend on both the dataset and architecture (Fig.~\ref{fig:sparsity-bathtub}).
The observed profiles describe how retained capacity is distributed through each model, but do not by themselves establish whether its depth or breadth is appropriate for the task.
These observations raise a broader question for future work: when do interpretable components produce an interpretable model? Understanding how depth, breadth, and sparsity affect whole-model readability and predictive accuracy could help guide architecture selection.

We have presented S2KAN, a method that integrates symbolic primitives directly into Kolmogorov--Arnold networks through differentiable gating and MDL-inspired regularization. 
By placing symbolic and dense representations on equal footing, S2KAN can learn interpretable activation functions while retaining dense components for approximation. 
Exploring whether this differentiable sparse selection mechanism can support learned symbolic structure in other architectures remains a direction for future work.
These results show how incorporating symbolic structure during training can produce compact KAN models whose learned responses can be examined through the contributions of their symbolic and dense components.

\appendix

\section{Methods}
\label{sec:methods}

S2KAN was implemented in PyTorch v2.8.0~\citep{paszke2019pytorch}. 
Optimization was performed using Adam~\citep{kingma2014adam} with default parameters and a constant learning rate of $10^{-3}$ for both coefficients and gates. 
Predictive-benchmark $R^2$ and RMSE values are computed on held-out test sets; the response diagnostics use the evaluation subsets specified in App.~\ref{sec:scientific-response-diagnostics}.
Non-spline coefficients were initialized from $U(-0.05, 0.05)$; spline coefficients were initialized as in~\citet{liu2025kan}.
For $\phi_{\text{spline}}$ we used SiLU residuals and B-splines with $G=10$ grid intervals and order $K=3$ for all experiments. 

We apply gates to individual activation terms before summing edge outputs into subnodes. Multiplication nodes multiply their two designated subnode sums: a closed term contributes zero, and a zero subnode makes the product zero. No factor is dropped from the product. A single dense gate jointly controls the spline and SiLU contributions.

Unless otherwise noted, spline grids were adapted using training activations before optimization and at the start of epochs $0,5,\ldots,45$, then held fixed; grid refinement was not used.
Chebyshev polynomials are defined on $[-1, 1]$, so activation function domains are tracked per activation and updated alongside spline grid updates; inputs are rescaled to this domain before computing the Chebyshev basis. 
Fourier terms use natural frequencies $\sin(qx), \cos(qx)$ without domain rescaling. 
We use the following implementations for selected symbolic terms: $\sqrt{x} \to \sqrt{\max(x, \epsilon)}$, $1/x \to 1/(x + \epsilon)$, and $\log(x) \to \log(\max(x, \epsilon))$, with $\epsilon = 10^{-8}$; $\exp(x)$ is clamped to $\exp(\mathrm{clamp}(x, -10, 10))$ to prevent overflow.

For regularization term $k = \sum_{\ell i j} \sum_{m} w_m p_{\ell ijm}$, we used $w_m = 1$ for all non-dense terms, and $w_m = K + G + 2$ for the dense term (the +2 accounts for the learnable scale parameters; Eq.~\eqref{eqn:kan-formulation-base+spline}). 

For the Hard Concrete distribution, we used temperature $\tau = 2/3$ and stretch parameters $\gamma = -0.1$, $\zeta = 1.1$ throughout. 
Unless otherwise noted, gates were initialized with $\alpha_i \sim \mathcal{N}(0, 0.1)$ except for spline gates which used $\alpha_i = -1$, providing a slight bias toward symbolic representations at initialization. 
No temperature annealing was used.

Unless otherwise noted, models were trained with a batch size of 128. For S2KAN, we set $\beta=0$ for the first 200 epochs, then switch to the stated value for the remainder of training; experiment-specific warmup durations are given below. 
Unless otherwise noted, baseline KAN used $\phi_{\text{spline}}$ only, with gates fixed open, and minimized raw MSE without regularization. S2KAN used the log-MSE objective in Eq.~\eqref{eqn:loss-function}. Appendix~\ref{sec:kan-loss-control} compares raw and log MSE for baseline KAN under otherwise identical settings.

The real-world datasets contain measurement noise, and we observed overfitting in baseline KANs. To limit overfitting, we split each training sample 80/20 into fitting and validation subsets for both KAN and S2KAN and selected the checkpoint with the lowest validation MSE, evaluated every ten epochs. These runs completed the full training budget; validation selected the reported checkpoint without stopping training. The Nguyen and dynamical-system datasets were generated without added observation noise; for these experiments, we report final training checkpoints rather than selecting checkpoints by validation error.

\textit{Real-world data (Table~\ref{tab:real-world-both}, Fig.~\ref{fig:gate-dynamics})}---
For concrete compressive strength, we used UCI dataset~\href{https://archive.ics.uci.edu/dataset/165/concrete+compressive+strength}{165}~\citep{YEH19981797,YEHconcrete2} with 8 raw features augmented by 5 derived features (water-cement ratio, water-binder ratio, total binder, total aggregate, $\log(\mathrm{age}+1)$), for 13 total.
We used an 80/20 train-test split and architecture $[13, 32, 1]$.
For superconductor critical temperature prediction, we used 5 features from UCI dataset~\href{https://archive.ics.uci.edu/dataset/464/superconductivty+data}{464}~\citep{hamidieh2018data,center-a}: number of elements, weighted mean valence, weighted mean first ionization energy, mean electron affinity, and valence entropy.
We sampled 1000 observations for fitting and validation and 1000 separate test observations, and used architecture $[5, 5, 1]$ (previously used in \citet{Bagrow_2025}).
For Airfoil (UCI dataset~\href{https://archive.ics.uci.edu/dataset/291/airfoil+self+noise}{291}) and Power Plant (UCI dataset~\href{https://archive.ics.uci.edu/dataset/294/combined+cycle+power+plant}{294}), we used the five and four supplied inputs, respectively, with architectures $[5, 5, 1]$ and $[4, 5, 1]$ for both methods.
Airfoil used the remaining 503 observations for testing; Power Plant used 1000 separate test observations.
For all four datasets and both methods, each input feature and the target were standardized as $(v-\mu_{\mathrm{fit}})/(\sigma_{\mathrm{fit}}+10^{-8})$, using means and population standard deviations computed only on the fitting subset.
Concrete used 659 fitting and 165 validation observations; Airfoil, Power Plant, and Superconductor each used 800 fitting and 200 validation observations.
The same transformation was applied to the fitting, validation, and test sets.
Predictions were transformed back to the dataset's target units before reporting RMSE: dB for Airfoil's scaled sound pressure level, MPa for Concrete, MW for Power Plant, and K for Superconductor.
For S2KAN, Concrete used symbolic library $\mathcal{S} = \{1, x, x^2, \sqrt{x}, \log(x+1)\}$; Airfoil, Power Plant, and Superconductor additionally included $x^3$ and $\exp(x)$.
S2KAN used Fourier modes $Q=2$ on all four datasets.
Baseline KAN used dense spline and SiLU activations with gates fixed open and no symbolic or Fourier terms; neither method used a Chebyshev basis.
All models were trained for 5000 epochs with batch size 64; S2KAN used 500 warmup epochs before sparsity regularization was activated.
Early stopping was not used.
Appendix~\ref{sec:sparsity-breakdown} provides a full per-layer breakdown of gate counts by term type.

\textit{Scientific consistency checks (Sec.~\ref{subsec:scientific-insights})}---
We evaluate the same validation-selected real-world checkpoints deterministically, using fitting-set preprocessing and reporting responses in physical units.
Native activation decompositions retain all active terms, protected operations, and fixed-input contributions; no further coefficient fitting is performed.
The response protocols, input-support restrictions, illustrative-example selection, and secondary results are given together in App.~\ref{sec:scientific-response-diagnostics}.

\textit{Dynamical systems (Table~\ref{tab:ikeda_ecosystem})}---
Training/test/unseen partitions contain 6,400/1,600/2,000 observations for Ikeda and 3,000/750/1,250 for the ecosystem; the unseen block is reserved for rollout evaluation.
The partitions cover consecutive time periods at the end of each simulated trajectory, in the order listed, and are shared across all configurations and initialization seeds. Both systems were fitted in their original state coordinates without standardization.
Ecosystem observations are separated by $\Delta t=0.5$ time units, as in \citet{PhysRevResearch.7.023037}.
For the Ikeda map, we used architecture $[2, 4, 4, 4, 2]$ with symbolic library $\mathcal{S} = \{\sqrt{x}, 1/(1+x^2)\}$, $P=4$ Chebyshev terms, and Fourier modes $Q=4$.
For the ecosystem, we used architecture $[3, 3, 3, 3]$ with symbolic library $\mathcal{S} = \{1, x, x^2, 1/(1+\max(x,0))\}$, no Chebyshev basis ($P=0$), and Fourier modes $Q=4$.
Both methods used the same architectures, following \citet{Bagrow_2025}.
In the ecosystem's residual formulation, the fixed identity connection is excluded from parameter counts and symbolic fidelity.
Early stopping was not used.
Training used batch size 128 for 20,000 epochs (Ikeda) or 45,000 epochs (ecosystem), with ten initialization seeds per configuration.
We evaluated $\beta\in\{0,1,3\}$ for Ikeda and $\beta\in\{0,0.1,1\}$ for the ecosystem, including $\beta=0.1$ to examine weaker ecosystem regularization.
One-step RMSE pools errors over observations and output coordinates before taking the square root.

To assess dynamical behavior beyond one-step prediction accuracy, we iterate each trained model in original state coordinates, starting from the first unseen state.
For long-orbit diagnostics, we discard 200 iterates and retain 20,000 iterates for Ikeda or 40,000 for the ecosystem, analyzing the second half.
Lyapunov spectra are estimated by the standard QR method \citep{skokos2010lyapunov}, using automatic-differentiation Jacobians along model trajectories; ecosystem rates are divided by $\Delta t$.
Reference spectra use generating maps and the final 100,000 points of 200,000-point trajectories following a 1,000-step transient.
Additional dynamical diagnostic procedures and results are described in App.~\ref{sec:ds-diagnostics}.

\textit{Nguyen benchmark (Table~\ref{tab:nguyen-full})}---
We evaluated the first 10 Nguyen problems~\citep{uy2011semantically}, generating 1024 training and 256 test points per problem.
Training and test inputs were sampled independently and uniformly over the tabulated domains, with independent coordinates for the bivariate problems. Inputs and targets were used without standardization.
Across all Nguyen $\beta$ values, S2KAN training stopped when gate decisiveness remained $\geq0.99$ for 500 consecutive epochs, or at 10,000 epochs; baseline KAN always completed 10,000 epochs. This stopping rule uses only gate probabilities, with no validation or test loss involved (see also App.~\ref{sec:nguyen-beta-sensitivity}). We report, for both baseline and S2KAN, medians over ten initialization seeds. In Table~\ref{tab:nguyen-full}, this stopping criterion was reached only by the ten small-architecture S2KAN fits for Nguyen-10, which closed all gates and stopped after 3,337--4,059 epochs; all other fits completed 10,000 epochs.
Three architectures were tested: small (no hidden layer), large (one hidden layer with 3 summation units), and large-mult (3 summation plus 1 multiplication unit).
S2KAN used symbolic library $\mathcal{S} = \{1, x, x^2, \sin(x), \cos(x)\}$, $P=4$, and $Q=4$. %
\textit{Sinc function (Fig.~\ref{fig:sincx})}---
We generated 1024 training and 256 test points on uniform grids over $x \in [1,15]$, with target $y=\sin(x)/x$ and no standardization.
Both models used a single multiplication unit with no hidden layer, shape $[1,(0,1)]$, trained for 8,000 epochs with batch size 32 and constant learning rate $10^{-3}$ for coefficients, scale parameters, and gates.
S2KAN used symbolic library $\mathcal{S}=\{x,1/(x+10^{-8})\}$, no Chebyshev terms ($P=0$), Fourier modes $Q=2$, regularization $\beta=1$, and 100 warmup epochs.
KAN used raw MSE, while S2KAN used the log-MSE objective. In this experiment, baseline gates were initialized near-open and remained trainable.
Figure~\ref{fig:sincx} shows the final checkpoints for seed 636139, the first seed in the experiment protocol; no coefficient refitting or term removal was applied. Early stopping was not used.
Appendix~\ref{subsec:misspecification} examines the effect of removing the reciprocal term from this dictionary.

\subsection{Measuring symbolic fidelity}
\label{subsec:methods-symbolic-fidelity}

We evaluate frozen models deterministically on all fitting inputs, excluding validation and test inputs, and apply the functional-activity criteria of Sec.~\ref{sec:symbolic-fidelity}.
All functionally active edges count equally across layers; the ecosystem's fixed identity additions are excluded.
For vector outputs, the numerical functional-activity floor uses the variance of all predicted values pooled across observations and output coordinates, about a single common mean.
For each functionally active edge, we fit
\begin{equation}
 s(u)=d+\sum_{m=1}^{j}a_m p_m(b_m u+c_m),\qquad j\leq J,
 \label{eq:sf-empirical-fit}
\end{equation}
with a free intercept and repeated primitives allowed.
We use $P=\{x,x^2,\sin\}$ for Nguyen and real-world data, adding $1/x$ for sinc, $\sqrt{\max(x,10^{-8})}$ and $1/(1+x^2)$ for Ikeda, or $1/(1+\max(x,0))$ for the ecosystem; both methods use the same evaluation dictionary.

The initialization grid uses mean-centered edge inputs and their 99th--1st percentile range $r$ (set to one if zero).
Sines use 40 geometrically spaced frequencies from 0.25 to 64 cycles per $r$ and eight equally spaced phases in $[0,\pi)$. %
Polynomial primitives use unit input scale and nine equally spaced shifts over $[-r,r]$; other primitives use scales $\{\pm\tfrac12,\pm1,\pm2\}$ with the same shifts.
These values initialize the search rather than constrain subsequent fitting.

After standardizing the edge response, greedy forward selection considers the three grid-initialized atoms with largest absolute normalized residual correlations and the best atom of each primitive.
For each candidate addition, SciPy's nonlinear least-squares solver jointly refits all coefficients, scales, shifts, and the intercept, with at most 2,000 function evaluations. Parameter step scales adapt to the column norms of the residual Jacobian. We retain the best fit only if it improves the residual sum of squares.
Fitting and scoring use the same input sample.
We report the fraction of functionally active edges with approximation $R^2\geq1-\epsilon^2$ for $J=1,2,3$ and $\epsilon=0.01,0.1$.
This searches additive expressions only; products and nested compositions are not searched.
For Fig.~\ref{fig:sincx}, $J=1$ and $\epsilon=0.01$ give 0/2 qualifying edges for KAN and 2/2 for S2KAN.

Large, mutually cancelling terms could produce accurate symbolic approximations with unstable coefficients.
As a sensitivity check (App.~\ref{sec:sf-conditioning-control}), we limit each fitted term's standard deviation to three times that of the edge response; 94.7\% of 6,132 SF scores remain unchanged.

\textit{AI assistance.} The authors used OpenAI Codex and Anthropic Claude to assist with code development, data analysis, and manuscript preparation. The authors reviewed and verified the outputs used in this work and take responsibility for the final content.

\section{Sparsity breakdown by term type}
\label{sec:sparsity-breakdown}

Tables~\ref{tab:sparsity-breakdown} and~\ref{tab:sparsity-breakdown-rw} report per-layer counts of active activation functions and open gates for the dynamical-systems and real-world S2KAN models, respectively, including all four real-world datasets. We summarize all ten initialization seeds for each dynamical-system configuration and all thirty fits (ten data splits, three initialization seeds per split) for each real-world dataset. Entries give medians and first and third quartiles. Active counts are defined by gate state; symbolic fidelity uses separate functional-activity and approximation criteria.

\begin{table}[p]
\centering
\caption{Sparsity breakdown for the dynamical-systems S2KAN models. Entries are medians over ten initialization seeds, with first and third quartiles below (omitted for constant counts). Ikeda uses the direct map at 20,000 epochs with $P=Q=4$; Ecosystem uses the residual map at 45,000 epochs with $P=0$, $Q=4$. Spline grids are updated early in training. Active means at least one open gate; $\mathcal{S}$-only excludes both $\mathcal{F}$ and $\mathcal{R}$, while $\mathcal{S}{+}\mathcal{F}$ excludes $\mathcal{R}$. Gates are counted at $p_{\mathrm{open}}>0.5$: $\mathcal{S}$ is the symbolic library, $\mathcal{F}$ is Chebyshev/Fourier, and $\mathcal{R}$ is the spline branch (one gate per edge, not one per spline coefficient). Ecosystem\textquotesingle{}s three fixed identity additions are excluded. Totals are computed within each model before aggregation, so their medians need not equal sums of layer medians.}
\label{tab:sparsity-breakdown}
\small
\begin{tabular}{@{}lrrrrrrr@{}}
\toprule
 & \multicolumn{4}{c}{Edges} & \multicolumn{3}{c}{Open gates} \\
\cmidrule(lr){2-5}\cmidrule(lr){6-8}
Layer & Edges & Active & $\mathcal{S}$-only & $\mathcal{S}{+}\mathcal{F}$ & $\mathcal{S}$ & $\mathcal{F}$ & $\mathcal{R}$ \\
\midrule
\multicolumn{8}{@{}l}{\textbf{Ikeda, $[2,4,4,4,2]$, $\beta=0$}} \\
\addlinespace[2pt]
L0 ($2\!\to\!4$) & \shortstack[r]{8\\[-1pt]{\fontsize{6}{7}\selectfont \vphantom{[0, 0]}}} & \shortstack[r]{8\\[-1pt]{\fontsize{6}{7}\selectfont \vphantom{[0, 0]}}} & \shortstack[r]{0\\[-1pt]{\fontsize{6}{7}\selectfont \vphantom{[0, 0]}}} & \shortstack[r]{0\\[-1pt]{\fontsize{6}{7}\selectfont [0, 0]}} & \shortstack[r]{0\\[-1pt]{\fontsize{6}{7}\selectfont \vphantom{[0, 0]}}} & \shortstack[r]{12.5\\[-1pt]{\fontsize{6}{7}\selectfont [11.25, 13]}} & \shortstack[r]{8\\[-1pt]{\fontsize{6}{7}\selectfont [8, 8]}} \\
L1 ($4\!\to\!4$) & \shortstack[r]{16\\[-1pt]{\fontsize{6}{7}\selectfont \vphantom{[0, 0]}}} & \shortstack[r]{15.5\\[-1pt]{\fontsize{6}{7}\selectfont [15, 16]}} & \shortstack[r]{0\\[-1pt]{\fontsize{6}{7}\selectfont \vphantom{[0, 0]}}} & \shortstack[r]{0\\[-1pt]{\fontsize{6}{7}\selectfont [0, 0]}} & \shortstack[r]{0.5\\[-1pt]{\fontsize{6}{7}\selectfont [0, 1]}} & \shortstack[r]{20.5\\[-1pt]{\fontsize{6}{7}\selectfont [17.25, 21.75]}} & \shortstack[r]{15\\[-1pt]{\fontsize{6}{7}\selectfont [14.25, 16]}} \\
L2 ($4\!\to\!4$) & \shortstack[r]{16\\[-1pt]{\fontsize{6}{7}\selectfont \vphantom{[0, 0]}}} & \shortstack[r]{14\\[-1pt]{\fontsize{6}{7}\selectfont [13.25, 14.75]}} & \shortstack[r]{0\\[-1pt]{\fontsize{6}{7}\selectfont [0, 0]}} & \shortstack[r]{3\\[-1pt]{\fontsize{6}{7}\selectfont [2, 4.75]}} & \shortstack[r]{2.5\\[-1pt]{\fontsize{6}{7}\selectfont [1, 3.75]}} & \shortstack[r]{24\\[-1pt]{\fontsize{6}{7}\selectfont [22, 26.75]}} & \shortstack[r]{10.5\\[-1pt]{\fontsize{6}{7}\selectfont [9, 12]}} \\
L3 ($4\!\to\!2$) & \shortstack[r]{8\\[-1pt]{\fontsize{6}{7}\selectfont \vphantom{[0, 0]}}} & \shortstack[r]{8\\[-1pt]{\fontsize{6}{7}\selectfont [8, 8]}} & \shortstack[r]{0\\[-1pt]{\fontsize{6}{7}\selectfont [0, 0]}} & \shortstack[r]{2\\[-1pt]{\fontsize{6}{7}\selectfont [2, 2]}} & \shortstack[r]{4\\[-1pt]{\fontsize{6}{7}\selectfont [3.25, 4.75]}} & \shortstack[r]{15.5\\[-1pt]{\fontsize{6}{7}\selectfont [13, 16.75]}} & \shortstack[r]{6\\[-1pt]{\fontsize{6}{7}\selectfont [6, 6]}} \\
Total & \shortstack[r]{48\\[-1pt]{\fontsize{6}{7}\selectfont \vphantom{[0, 0]}}} & \shortstack[r]{45\\[-1pt]{\fontsize{6}{7}\selectfont [43, 46]}} & \shortstack[r]{0\\[-1pt]{\fontsize{6}{7}\selectfont [0, 0]}} & \shortstack[r]{5.5\\[-1pt]{\fontsize{6}{7}\selectfont [4, 8.75]}} & \shortstack[r]{7\\[-1pt]{\fontsize{6}{7}\selectfont [6, 8.75]}} & \shortstack[r]{72.5\\[-1pt]{\fontsize{6}{7}\selectfont [67, 77.25]}} & \shortstack[r]{38\\[-1pt]{\fontsize{6}{7}\selectfont [37, 41]}} \\
\addlinespace[5pt]
\multicolumn{8}{@{}l}{\textbf{Ikeda, $[2,4,4,4,2]$, $\beta=1$}} \\
\addlinespace[2pt]
L0 ($2\!\to\!4$) & \shortstack[r]{8\\[-1pt]{\fontsize{6}{7}\selectfont \vphantom{[0, 0]}}} & \shortstack[r]{8\\[-1pt]{\fontsize{6}{7}\selectfont \vphantom{[0, 0]}}} & \shortstack[r]{0\\[-1pt]{\fontsize{6}{7}\selectfont \vphantom{[0, 0]}}} & \shortstack[r]{2\\[-1pt]{\fontsize{6}{7}\selectfont [1, 2]}} & \shortstack[r]{0\\[-1pt]{\fontsize{6}{7}\selectfont [0, 0]}} & \shortstack[r]{13.5\\[-1pt]{\fontsize{6}{7}\selectfont [13, 16]}} & \shortstack[r]{6\\[-1pt]{\fontsize{6}{7}\selectfont [6, 7]}} \\
L1 ($4\!\to\!4$) & \shortstack[r]{16\\[-1pt]{\fontsize{6}{7}\selectfont \vphantom{[0, 0]}}} & \shortstack[r]{12\\[-1pt]{\fontsize{6}{7}\selectfont [10, 13]}} & \shortstack[r]{0\\[-1pt]{\fontsize{6}{7}\selectfont [0, 0]}} & \shortstack[r]{8.5\\[-1pt]{\fontsize{6}{7}\selectfont [6.25, 10]}} & \shortstack[r]{1\\[-1pt]{\fontsize{6}{7}\selectfont [0.25, 2]}} & \shortstack[r]{23\\[-1pt]{\fontsize{6}{7}\selectfont [19.25, 25]}} & \shortstack[r]{3\\[-1pt]{\fontsize{6}{7}\selectfont [2, 4]}} \\
L2 ($4\!\to\!4$) & \shortstack[r]{16\\[-1pt]{\fontsize{6}{7}\selectfont \vphantom{[0, 0]}}} & \shortstack[r]{13.5\\[-1pt]{\fontsize{6}{7}\selectfont [13, 15.75]}} & \shortstack[r]{0\\[-1pt]{\fontsize{6}{7}\selectfont \vphantom{[0, 0]}}} & \shortstack[r]{11.5\\[-1pt]{\fontsize{6}{7}\selectfont [10, 14]}} & \shortstack[r]{2\\[-1pt]{\fontsize{6}{7}\selectfont [1, 3.75]}} & \shortstack[r]{31\\[-1pt]{\fontsize{6}{7}\selectfont [28.25, 38.5]}} & \shortstack[r]{2\\[-1pt]{\fontsize{6}{7}\selectfont [1, 2]}} \\
L3 ($4\!\to\!2$) & \shortstack[r]{8\\[-1pt]{\fontsize{6}{7}\selectfont \vphantom{[0, 0]}}} & \shortstack[r]{8\\[-1pt]{\fontsize{6}{7}\selectfont [8, 8]}} & \shortstack[r]{0\\[-1pt]{\fontsize{6}{7}\selectfont [0, 0]}} & \shortstack[r]{5.5\\[-1pt]{\fontsize{6}{7}\selectfont [4.25, 6]}} & \shortstack[r]{4.5\\[-1pt]{\fontsize{6}{7}\selectfont [4, 5]}} & \shortstack[r]{21\\[-1pt]{\fontsize{6}{7}\selectfont [18, 21.75]}} & \shortstack[r]{2.5\\[-1pt]{\fontsize{6}{7}\selectfont [2, 3]}} \\
Total & \shortstack[r]{48\\[-1pt]{\fontsize{6}{7}\selectfont \vphantom{[0, 0]}}} & \shortstack[r]{41.5\\[-1pt]{\fontsize{6}{7}\selectfont [39.5, 42]}} & \shortstack[r]{0\\[-1pt]{\fontsize{6}{7}\selectfont [0, 1]}} & \shortstack[r]{27.5\\[-1pt]{\fontsize{6}{7}\selectfont [25.5, 30]}} & \shortstack[r]{8.5\\[-1pt]{\fontsize{6}{7}\selectfont [7, 10.5]}} & \shortstack[r]{90.5\\[-1pt]{\fontsize{6}{7}\selectfont [83.5, 96]}} & \shortstack[r]{13\\[-1pt]{\fontsize{6}{7}\selectfont [12.25, 14.5]}} \\
\addlinespace[5pt]
\multicolumn{8}{@{}l}{\textbf{Ikeda, $[2,4,4,4,2]$, $\beta=3$}} \\
\addlinespace[2pt]
L0 ($2\!\to\!4$) & \shortstack[r]{8\\[-1pt]{\fontsize{6}{7}\selectfont \vphantom{[0, 0]}}} & \shortstack[r]{8\\[-1pt]{\fontsize{6}{7}\selectfont \vphantom{[0, 0]}}} & \shortstack[r]{0\\[-1pt]{\fontsize{6}{7}\selectfont \vphantom{[0, 0]}}} & \shortstack[r]{3\\[-1pt]{\fontsize{6}{7}\selectfont [2.25, 3.75]}} & \shortstack[r]{0\\[-1pt]{\fontsize{6}{7}\selectfont [0, 0.75]}} & \shortstack[r]{15\\[-1pt]{\fontsize{6}{7}\selectfont [13, 16]}} & \shortstack[r]{5\\[-1pt]{\fontsize{6}{7}\selectfont [4.25, 5.75]}} \\
L1 ($4\!\to\!4$) & \shortstack[r]{16\\[-1pt]{\fontsize{6}{7}\selectfont \vphantom{[0, 0]}}} & \shortstack[r]{11.5\\[-1pt]{\fontsize{6}{7}\selectfont [10.25, 12]}} & \shortstack[r]{0\\[-1pt]{\fontsize{6}{7}\selectfont [0, 0]}} & \shortstack[r]{11.5\\[-1pt]{\fontsize{6}{7}\selectfont [9.25, 12]}} & \shortstack[r]{2\\[-1pt]{\fontsize{6}{7}\selectfont [0.25, 3]}} & \shortstack[r]{21\\[-1pt]{\fontsize{6}{7}\selectfont [21, 23]}} & \shortstack[r]{0.5\\[-1pt]{\fontsize{6}{7}\selectfont [0, 1]}} \\
L2 ($4\!\to\!4$) & \shortstack[r]{16\\[-1pt]{\fontsize{6}{7}\selectfont \vphantom{[0, 0]}}} & \shortstack[r]{14\\[-1pt]{\fontsize{6}{7}\selectfont [12.25, 14]}} & \shortstack[r]{0\\[-1pt]{\fontsize{6}{7}\selectfont [0, 0]}} & \shortstack[r]{13.5\\[-1pt]{\fontsize{6}{7}\selectfont [12.25, 14]}} & \shortstack[r]{2.5\\[-1pt]{\fontsize{6}{7}\selectfont [1, 4.5]}} & \shortstack[r]{28\\[-1pt]{\fontsize{6}{7}\selectfont [26.25, 32.5]}} & \shortstack[r]{0\\[-1pt]{\fontsize{6}{7}\selectfont [0, 0]}} \\
L3 ($4\!\to\!2$) & \shortstack[r]{8\\[-1pt]{\fontsize{6}{7}\selectfont \vphantom{[0, 0]}}} & \shortstack[r]{8\\[-1pt]{\fontsize{6}{7}\selectfont [8, 8]}} & \shortstack[r]{0\\[-1pt]{\fontsize{6}{7}\selectfont [0, 0]}} & \shortstack[r]{7\\[-1pt]{\fontsize{6}{7}\selectfont [7, 8]}} & \shortstack[r]{6.5\\[-1pt]{\fontsize{6}{7}\selectfont [5.25, 7]}} & \shortstack[r]{21.5\\[-1pt]{\fontsize{6}{7}\selectfont [19, 24.75]}} & \shortstack[r]{0.5\\[-1pt]{\fontsize{6}{7}\selectfont [0, 1]}} \\
Total & \shortstack[r]{48\\[-1pt]{\fontsize{6}{7}\selectfont \vphantom{[0, 0]}}} & \shortstack[r]{40\\[-1pt]{\fontsize{6}{7}\selectfont [38.25, 40.75]}} & \shortstack[r]{0\\[-1pt]{\fontsize{6}{7}\selectfont [0, 0.75]}} & \shortstack[r]{33\\[-1pt]{\fontsize{6}{7}\selectfont [32.25, 36.25]}} & \shortstack[r]{11\\[-1pt]{\fontsize{6}{7}\selectfont [10.25, 12.75]}} & \shortstack[r]{88\\[-1pt]{\fontsize{6}{7}\selectfont [83.5, 93]}} & \shortstack[r]{6\\[-1pt]{\fontsize{6}{7}\selectfont [5, 7.75]}} \\
\addlinespace[5pt]
\multicolumn{8}{@{}l}{\textbf{Ecosystem, $[3,3,3,3]$, $\beta=0$}} \\
\addlinespace[2pt]
L0 ($3\!\to\!3$) & \shortstack[r]{9\\[-1pt]{\fontsize{6}{7}\selectfont \vphantom{[0, 0]}}} & \shortstack[r]{9\\[-1pt]{\fontsize{6}{7}\selectfont [8.25, 9]}} & \shortstack[r]{0\\[-1pt]{\fontsize{6}{7}\selectfont \vphantom{[0, 0]}}} & \shortstack[r]{1\\[-1pt]{\fontsize{6}{7}\selectfont [0.25, 1]}} & \shortstack[r]{2\\[-1pt]{\fontsize{6}{7}\selectfont [2, 2]}} & \shortstack[r]{10\\[-1pt]{\fontsize{6}{7}\selectfont [9.25, 10]}} & \shortstack[r]{8\\[-1pt]{\fontsize{6}{7}\selectfont [7.25, 8.75]}} \\
L1 ($3\!\to\!3$) & \shortstack[r]{9\\[-1pt]{\fontsize{6}{7}\selectfont \vphantom{[0, 0]}}} & \shortstack[r]{7\\[-1pt]{\fontsize{6}{7}\selectfont [7, 8]}} & \shortstack[r]{0\\[-1pt]{\fontsize{6}{7}\selectfont [0, 1]}} & \shortstack[r]{3\\[-1pt]{\fontsize{6}{7}\selectfont [2.25, 3]}} & \shortstack[r]{8.5\\[-1pt]{\fontsize{6}{7}\selectfont [7.25, 10]}} & \shortstack[r]{8.5\\[-1pt]{\fontsize{6}{7}\selectfont [6.25, 9.75]}} & \shortstack[r]{4.5\\[-1pt]{\fontsize{6}{7}\selectfont [4, 5]}} \\
L2 ($3\!\to\!3$) & \shortstack[r]{9\\[-1pt]{\fontsize{6}{7}\selectfont \vphantom{[0, 0]}}} & \shortstack[r]{8.5\\[-1pt]{\fontsize{6}{7}\selectfont [8, 9]}} & \shortstack[r]{3\\[-1pt]{\fontsize{6}{7}\selectfont [2, 4]}} & \shortstack[r]{4.5\\[-1pt]{\fontsize{6}{7}\selectfont [4, 5]}} & \shortstack[r]{8\\[-1pt]{\fontsize{6}{7}\selectfont [7.25, 9]}} & \shortstack[r]{4\\[-1pt]{\fontsize{6}{7}\selectfont [3, 4]}} & \shortstack[r]{4\\[-1pt]{\fontsize{6}{7}\selectfont [3.25, 5]}} \\
Total & \shortstack[r]{27\\[-1pt]{\fontsize{6}{7}\selectfont \vphantom{[0, 0]}}} & \shortstack[r]{25\\[-1pt]{\fontsize{6}{7}\selectfont [24, 25.75]}} & \shortstack[r]{4\\[-1pt]{\fontsize{6}{7}\selectfont [2.25, 4]}} & \shortstack[r]{8\\[-1pt]{\fontsize{6}{7}\selectfont [7.25, 8.75]}} & \shortstack[r]{18.5\\[-1pt]{\fontsize{6}{7}\selectfont [16.25, 20.5]}} & \shortstack[r]{21.5\\[-1pt]{\fontsize{6}{7}\selectfont [20, 23]}} & \shortstack[r]{16.5\\[-1pt]{\fontsize{6}{7}\selectfont [15.25, 18]}} \\
\addlinespace[5pt]
\multicolumn{8}{@{}l}{\textbf{Ecosystem, $[3,3,3,3]$, $\beta=0.1$}} \\
\addlinespace[2pt]
L0 ($3\!\to\!3$) & \shortstack[r]{9\\[-1pt]{\fontsize{6}{7}\selectfont \vphantom{[0, 0]}}} & \shortstack[r]{8\\[-1pt]{\fontsize{6}{7}\selectfont [8, 8.75]}} & \shortstack[r]{0\\[-1pt]{\fontsize{6}{7}\selectfont \vphantom{[0, 0]}}} & \shortstack[r]{1\\[-1pt]{\fontsize{6}{7}\selectfont [1, 2]}} & \shortstack[r]{2\\[-1pt]{\fontsize{6}{7}\selectfont [2, 2]}} & \shortstack[r]{10\\[-1pt]{\fontsize{6}{7}\selectfont [10, 10]}} & \shortstack[r]{6.5\\[-1pt]{\fontsize{6}{7}\selectfont [6, 7.75]}} \\
L1 ($3\!\to\!3$) & \shortstack[r]{9\\[-1pt]{\fontsize{6}{7}\selectfont \vphantom{[0, 0]}}} & \shortstack[r]{7\\[-1pt]{\fontsize{6}{7}\selectfont [7, 8]}} & \shortstack[r]{1\\[-1pt]{\fontsize{6}{7}\selectfont [0.25, 2]}} & \shortstack[r]{6\\[-1pt]{\fontsize{6}{7}\selectfont [6, 7]}} & \shortstack[r]{9.5\\[-1pt]{\fontsize{6}{7}\selectfont [8.25, 10]}} & \shortstack[r]{10\\[-1pt]{\fontsize{6}{7}\selectfont [8.25, 11]}} & \shortstack[r]{1\\[-1pt]{\fontsize{6}{7}\selectfont [1, 1]}} \\
L2 ($3\!\to\!3$) & \shortstack[r]{9\\[-1pt]{\fontsize{6}{7}\selectfont \vphantom{[0, 0]}}} & \shortstack[r]{8\\[-1pt]{\fontsize{6}{7}\selectfont [7, 8]}} & \shortstack[r]{3\\[-1pt]{\fontsize{6}{7}\selectfont [3, 4]}} & \shortstack[r]{6\\[-1pt]{\fontsize{6}{7}\selectfont [5.25, 6.75]}} & \shortstack[r]{8\\[-1pt]{\fontsize{6}{7}\selectfont [7.25, 8]}} & \shortstack[r]{4\\[-1pt]{\fontsize{6}{7}\selectfont [3, 4.75]}} & \shortstack[r]{2\\[-1pt]{\fontsize{6}{7}\selectfont [2, 2]}} \\
Total & \shortstack[r]{27\\[-1pt]{\fontsize{6}{7}\selectfont \vphantom{[0, 0]}}} & \shortstack[r]{23\\[-1pt]{\fontsize{6}{7}\selectfont [22, 24.75]}} & \shortstack[r]{4\\[-1pt]{\fontsize{6}{7}\selectfont [4, 5.75]}} & \shortstack[r]{13\\[-1pt]{\fontsize{6}{7}\selectfont [12.25, 13.75]}} & \shortstack[r]{19.5\\[-1pt]{\fontsize{6}{7}\selectfont [17, 20]}} & \shortstack[r]{23\\[-1pt]{\fontsize{6}{7}\selectfont [22.25, 24.75]}} & \shortstack[r]{9\\[-1pt]{\fontsize{6}{7}\selectfont [9, 10]}} \\
\addlinespace[5pt]
\multicolumn{8}{@{}l}{\textbf{Ecosystem, $[3,3,3,3]$, $\beta=1$}} \\
\addlinespace[2pt]
L0 ($3\!\to\!3$) & \shortstack[r]{9\\[-1pt]{\fontsize{6}{7}\selectfont \vphantom{[0, 0]}}} & \shortstack[r]{7\\[-1pt]{\fontsize{6}{7}\selectfont [7, 7.75]}} & \shortstack[r]{0\\[-1pt]{\fontsize{6}{7}\selectfont \vphantom{[0, 0]}}} & \shortstack[r]{4.5\\[-1pt]{\fontsize{6}{7}\selectfont [4, 5]}} & \shortstack[r]{2.5\\[-1pt]{\fontsize{6}{7}\selectfont [2, 3]}} & \shortstack[r]{12\\[-1pt]{\fontsize{6}{7}\selectfont [11, 12.75]}} & \shortstack[r]{3\\[-1pt]{\fontsize{6}{7}\selectfont [2, 3]}} \\
L1 ($3\!\to\!3$) & \shortstack[r]{9\\[-1pt]{\fontsize{6}{7}\selectfont \vphantom{[0, 0]}}} & \shortstack[r]{7\\[-1pt]{\fontsize{6}{7}\selectfont [7, 8]}} & \shortstack[r]{1\\[-1pt]{\fontsize{6}{7}\selectfont [1, 2]}} & \shortstack[r]{7\\[-1pt]{\fontsize{6}{7}\selectfont [7, 8]}} & \shortstack[r]{9\\[-1pt]{\fontsize{6}{7}\selectfont [8.25, 10]}} & \shortstack[r]{9.5\\[-1pt]{\fontsize{6}{7}\selectfont [8.25, 10.75]}} & \shortstack[r]{0\\[-1pt]{\fontsize{6}{7}\selectfont \vphantom{[0, 0]}}} \\
L2 ($3\!\to\!3$) & \shortstack[r]{9\\[-1pt]{\fontsize{6}{7}\selectfont \vphantom{[0, 0]}}} & \shortstack[r]{8\\[-1pt]{\fontsize{6}{7}\selectfont [8, 8.75]}} & \shortstack[r]{5\\[-1pt]{\fontsize{6}{7}\selectfont [3.25, 5]}} & \shortstack[r]{8\\[-1pt]{\fontsize{6}{7}\selectfont [8, 8.75]}} & \shortstack[r]{8.5\\[-1pt]{\fontsize{6}{7}\selectfont [8, 9.75]}} & \shortstack[r]{4.5\\[-1pt]{\fontsize{6}{7}\selectfont [3, 5]}} & \shortstack[r]{0\\[-1pt]{\fontsize{6}{7}\selectfont \vphantom{[0, 0]}}} \\
Total & \shortstack[r]{27\\[-1pt]{\fontsize{6}{7}\selectfont \vphantom{[0, 0]}}} & \shortstack[r]{22.5\\[-1pt]{\fontsize{6}{7}\selectfont [22, 23.75]}} & \shortstack[r]{6\\[-1pt]{\fontsize{6}{7}\selectfont [5, 6.75]}} & \shortstack[r]{20\\[-1pt]{\fontsize{6}{7}\selectfont [19, 21]}} & \shortstack[r]{19.5\\[-1pt]{\fontsize{6}{7}\selectfont [19, 20.75]}} & \shortstack[r]{25.5\\[-1pt]{\fontsize{6}{7}\selectfont [24.25, 27.5]}} & \shortstack[r]{3\\[-1pt]{\fontsize{6}{7}\selectfont [2, 3]}} \\
\addlinespace[5pt]
\bottomrule
\end{tabular}
\end{table}

\begin{table}%
\centering
\caption{Sparsity breakdown for the real-world S2KAN models. All use $\beta=1$, $P=0$, $Q=2$. Entries are medians over ten data splits with three initialization seeds per split (30 fits per dataset); brackets below give the first and third quartiles, omitted for constant counts. These are descriptive intervals. Gate definitions and within-model totals follow the dynamical-systems table (Table~\ref{tab:sparsity-breakdown}).}
\label{tab:sparsity-breakdown-rw}
\small
\begin{tabular}{@{}lrrrrrrr@{}}
\toprule
 & \multicolumn{4}{c}{Edges} & \multicolumn{3}{c}{Open gates} \\
\cmidrule(lr){2-5}\cmidrule(lr){6-8}
Layer & Edges & Active & $\mathcal{S}$-only & $\mathcal{S}{+}\mathcal{F}$ & $\mathcal{S}$ & $\mathcal{F}$ & $\mathcal{R}$ \\
\midrule
\multicolumn{8}{@{}l}{\textbf{Airfoil, $[5,5,1]$, $\beta=1$}} \\
\addlinespace[2pt]
L0 ($5\!\to\!5$) & \shortstack[r]{25\\[-1pt]{\fontsize{6}{7}\selectfont \vphantom{[0, 0]}}} & \shortstack[r]{21\\[-1pt]{\fontsize{6}{7}\selectfont [19.25, 21]}} & \shortstack[r]{6\\[-1pt]{\fontsize{6}{7}\selectfont [5, 7]}} & \shortstack[r]{18\\[-1pt]{\fontsize{6}{7}\selectfont [17.25, 19]}} & \shortstack[r]{27\\[-1pt]{\fontsize{6}{7}\selectfont [25, 31]}} & \shortstack[r]{20\\[-1pt]{\fontsize{6}{7}\selectfont [17, 21]}} & \shortstack[r]{2\\[-1pt]{\fontsize{6}{7}\selectfont [2, 2]}} \\
L1 ($5\!\to\!1$) & \shortstack[r]{5\\[-1pt]{\fontsize{6}{7}\selectfont \vphantom{[0, 0]}}} & \shortstack[r]{5\\[-1pt]{\fontsize{6}{7}\selectfont \vphantom{[0, 0]}}} & \shortstack[r]{0\\[-1pt]{\fontsize{6}{7}\selectfont \vphantom{[0, 0]}}} & \shortstack[r]{4\\[-1pt]{\fontsize{6}{7}\selectfont [4, 5]}} & \shortstack[r]{7\\[-1pt]{\fontsize{6}{7}\selectfont [5, 7.75]}} & \shortstack[r]{10\\[-1pt]{\fontsize{6}{7}\selectfont [9, 11]}} & \shortstack[r]{1\\[-1pt]{\fontsize{6}{7}\selectfont [0, 1]}} \\
Total & \shortstack[r]{30\\[-1pt]{\fontsize{6}{7}\selectfont \vphantom{[0, 0]}}} & \shortstack[r]{26\\[-1pt]{\fontsize{6}{7}\selectfont [24.25, 26]}} & \shortstack[r]{6\\[-1pt]{\fontsize{6}{7}\selectfont [5, 7]}} & \shortstack[r]{23\\[-1pt]{\fontsize{6}{7}\selectfont [22, 23.75]}} & \shortstack[r]{33.5\\[-1pt]{\fontsize{6}{7}\selectfont [31, 36]}} & \shortstack[r]{30.5\\[-1pt]{\fontsize{6}{7}\selectfont [27, 32]}} & \shortstack[r]{3\\[-1pt]{\fontsize{6}{7}\selectfont [2, 3]}} \\
\addlinespace[5pt]
\multicolumn{8}{@{}l}{\textbf{Concrete, $[13,32,1]$, $\beta=1$}} \\
\addlinespace[2pt]
L0 ($13\!\to\!32$) & \shortstack[r]{416\\[-1pt]{\fontsize{6}{7}\selectfont \vphantom{[0, 0]}}} & \shortstack[r]{45\\[-1pt]{\fontsize{6}{7}\selectfont [43.25, 48]}} & \shortstack[r]{23\\[-1pt]{\fontsize{6}{7}\selectfont [19.25, 25.75]}} & \shortstack[r]{45\\[-1pt]{\fontsize{6}{7}\selectfont [43.25, 48]}} & \shortstack[r]{27\\[-1pt]{\fontsize{6}{7}\selectfont [24.25, 30]}} & \shortstack[r]{25\\[-1pt]{\fontsize{6}{7}\selectfont [23, 28.75]}} & \shortstack[r]{0\\[-1pt]{\fontsize{6}{7}\selectfont \vphantom{[0, 0]}}} \\
L1 ($32\!\to\!1$) & \shortstack[r]{32\\[-1pt]{\fontsize{6}{7}\selectfont \vphantom{[0, 0]}}} & \shortstack[r]{16\\[-1pt]{\fontsize{6}{7}\selectfont [14.25, 16]}} & \shortstack[r]{3\\[-1pt]{\fontsize{6}{7}\selectfont [2, 4]}} & \shortstack[r]{16\\[-1pt]{\fontsize{6}{7}\selectfont [14.25, 16]}} & \shortstack[r]{7\\[-1pt]{\fontsize{6}{7}\selectfont [6.25, 8.75]}} & \shortstack[r]{14\\[-1pt]{\fontsize{6}{7}\selectfont [14, 15]}} & \shortstack[r]{0\\[-1pt]{\fontsize{6}{7}\selectfont \vphantom{[0, 0]}}} \\
Total & \shortstack[r]{448\\[-1pt]{\fontsize{6}{7}\selectfont \vphantom{[0, 0]}}} & \shortstack[r]{61\\[-1pt]{\fontsize{6}{7}\selectfont [58, 63.75]}} & \shortstack[r]{25\\[-1pt]{\fontsize{6}{7}\selectfont [22.25, 28.75]}} & \shortstack[r]{61\\[-1pt]{\fontsize{6}{7}\selectfont [58, 63.75]}} & \shortstack[r]{35\\[-1pt]{\fontsize{6}{7}\selectfont [32, 37]}} & \shortstack[r]{40\\[-1pt]{\fontsize{6}{7}\selectfont [37, 43]}} & \shortstack[r]{0\\[-1pt]{\fontsize{6}{7}\selectfont \vphantom{[0, 0]}}} \\
\addlinespace[5pt]
\multicolumn{8}{@{}l}{\textbf{Power Plant, $[4,5,1]$, $\beta=1$}} \\
\addlinespace[2pt]
L0 ($4\!\to\!5$) & \shortstack[r]{20\\[-1pt]{\fontsize{6}{7}\selectfont \vphantom{[0, 0]}}} & \shortstack[r]{11\\[-1pt]{\fontsize{6}{7}\selectfont [10, 12]}} & \shortstack[r]{6\\[-1pt]{\fontsize{6}{7}\selectfont [4.25, 7]}} & \shortstack[r]{11\\[-1pt]{\fontsize{6}{7}\selectfont [10, 12]}} & \shortstack[r]{14\\[-1pt]{\fontsize{6}{7}\selectfont [12, 15.75]}} & \shortstack[r]{5\\[-1pt]{\fontsize{6}{7}\selectfont [4.25, 6]}} & \shortstack[r]{0\\[-1pt]{\fontsize{6}{7}\selectfont \vphantom{[0, 0]}}} \\
L1 ($5\!\to\!1$) & \shortstack[r]{5\\[-1pt]{\fontsize{6}{7}\selectfont \vphantom{[0, 0]}}} & \shortstack[r]{5\\[-1pt]{\fontsize{6}{7}\selectfont \vphantom{[0, 0]}}} & \shortstack[r]{0\\[-1pt]{\fontsize{6}{7}\selectfont [0, 0]}} & \shortstack[r]{5\\[-1pt]{\fontsize{6}{7}\selectfont \vphantom{[0, 0]}}} & \shortstack[r]{8\\[-1pt]{\fontsize{6}{7}\selectfont [7, 9]}} & \shortstack[r]{10\\[-1pt]{\fontsize{6}{7}\selectfont [9, 10]}} & \shortstack[r]{0\\[-1pt]{\fontsize{6}{7}\selectfont \vphantom{[0, 0]}}} \\
Total & \shortstack[r]{25\\[-1pt]{\fontsize{6}{7}\selectfont \vphantom{[0, 0]}}} & \shortstack[r]{16\\[-1pt]{\fontsize{6}{7}\selectfont [15, 17]}} & \shortstack[r]{6\\[-1pt]{\fontsize{6}{7}\selectfont [5, 7]}} & \shortstack[r]{16\\[-1pt]{\fontsize{6}{7}\selectfont [15, 17]}} & \shortstack[r]{22\\[-1pt]{\fontsize{6}{7}\selectfont [19.25, 23]}} & \shortstack[r]{15\\[-1pt]{\fontsize{6}{7}\selectfont [13.25, 16]}} & \shortstack[r]{0\\[-1pt]{\fontsize{6}{7}\selectfont \vphantom{[0, 0]}}} \\
\addlinespace[5pt]
\multicolumn{8}{@{}l}{\textbf{Superconductor, $[5,5,1]$, $\beta=1$}} \\
\addlinespace[2pt]
L0 ($5\!\to\!5$) & \shortstack[r]{25\\[-1pt]{\fontsize{6}{7}\selectfont \vphantom{[0, 0]}}} & \shortstack[r]{12\\[-1pt]{\fontsize{6}{7}\selectfont [10, 14]}} & \shortstack[r]{5\\[-1pt]{\fontsize{6}{7}\selectfont [4, 6]}} & \shortstack[r]{10\\[-1pt]{\fontsize{6}{7}\selectfont [8.25, 12]}} & \shortstack[r]{9.5\\[-1pt]{\fontsize{6}{7}\selectfont [7, 12]}} & \shortstack[r]{7\\[-1pt]{\fontsize{6}{7}\selectfont [5, 10.75]}} & \shortstack[r]{2\\[-1pt]{\fontsize{6}{7}\selectfont [1.25, 3]}} \\
L1 ($5\!\to\!1$) & \shortstack[r]{5\\[-1pt]{\fontsize{6}{7}\selectfont \vphantom{[0, 0]}}} & \shortstack[r]{4\\[-1pt]{\fontsize{6}{7}\selectfont [4, 5]}} & \shortstack[r]{0\\[-1pt]{\fontsize{6}{7}\selectfont [0, 1]}} & \shortstack[r]{4\\[-1pt]{\fontsize{6}{7}\selectfont [4, 5]}} & \shortstack[r]{4\\[-1pt]{\fontsize{6}{7}\selectfont [4, 5]}} & \shortstack[r]{6\\[-1pt]{\fontsize{6}{7}\selectfont [5, 8]}} & \shortstack[r]{0\\[-1pt]{\fontsize{6}{7}\selectfont [0, 0]}} \\
Total & \shortstack[r]{30\\[-1pt]{\fontsize{6}{7}\selectfont \vphantom{[0, 0]}}} & \shortstack[r]{16\\[-1pt]{\fontsize{6}{7}\selectfont [14.25, 19]}} & \shortstack[r]{6\\[-1pt]{\fontsize{6}{7}\selectfont [4, 7]}} & \shortstack[r]{14\\[-1pt]{\fontsize{6}{7}\selectfont [12.25, 16.75]}} & \shortstack[r]{14\\[-1pt]{\fontsize{6}{7}\selectfont [11, 19.25]}} & \shortstack[r]{13\\[-1pt]{\fontsize{6}{7}\selectfont [9.5, 18.75]}} & \shortstack[r]{2\\[-1pt]{\fontsize{6}{7}\selectfont [2, 3]}} \\
\addlinespace[5pt]
\bottomrule
\end{tabular}
\end{table}

\section{Training and inference times}
\label{sec:training-inference-times}

Table~\ref{tab:real-world-timings} compares training and inference times for the four real-world datasets on an AMD EPYC 9654 (single-threaded). Entries are medians and first and third quartiles over thirty fits per dataset and method (ten data splits and three initialization seeds). Training times cover the complete 5,000-epoch fitting pipeline, including validation checks and checkpoint saving. Prior to inference, both models are exported to an evaluator that omits computations for inactive terms wherever possible. Times are per entire batch of one or 256 observations, excluding preprocessing, model loading, and export. Each model's latency is the median of eight warmed timing blocks, with paired methods timed sequentially on the same pinned core.

S2KAN's additional training cost is accompanied by lower inference latency in seven of eight dataset--batch-size settings: batch-256 inference is approximately 1.5--30 times faster, while single-observation inference is faster except on Airfoil.

\begin{table*}%
\centering
\caption{Training and inference wall times for the models in Table~\ref{tab:real-world-both}, on an AMD EPYC 9654 (single-threaded). Entries are medians [first quartile, third quartile] over 30 fits per row. 
S2KAN takes about 1.6 times as long to train but generally provides faster inference.
}
\label{tab:real-world-timings}
\small
\begin{tabular}{@{}lr@{\hspace{0.4em}}rr@{\hspace{0.4em}}rr@{\hspace{0.4em}}r@{}}
\toprule
 & \multicolumn{2}{c}{} & \multicolumn{4}{c}{Inference} \\
\cmidrule(lr){4-7}
Model & \multicolumn{2}{c}{Training (min)} & \multicolumn{2}{c}{Batch 1 ($\mu\mathrm{s}$)} & \multicolumn{2}{c}{Batch 256 ($\mu\mathrm{s}$)} \\
\midrule
\multicolumn{7}{@{}l}{\textbf{Airfoil}} \\
\addlinespace[2pt]
KAN & 2.77 & {\scriptsize [2.73, 2.80]} & 452.7 & {\scriptsize [442.0, 472.1]} & 1128.0 & {\scriptsize [1106.5, 1172.0]} \\
S2KAN & 4.34 & {\scriptsize [4.24, 4.42]} & 514.0 & {\scriptsize [318.4, 531.0]} & 731.6 & {\scriptsize [558.3, 842.8]} \\
\addlinespace[5pt]
\multicolumn{7}{@{}l}{\textbf{Concrete}} \\
\addlinespace[2pt]
KAN & 4.71 & {\scriptsize [3.93, 4.94]} & 484.7 & {\scriptsize [473.0, 490.2]} & 3427.9 & {\scriptsize [3352.8, 3806.8]} \\
S2KAN & 7.49 & {\scriptsize [6.10, 7.75]} & 66.4 & {\scriptsize [63.0, 70.2]} & 113.1 & {\scriptsize [106.2, 116.8]} \\
\addlinespace[5pt]
\multicolumn{7}{@{}l}{\textbf{Power plant}} \\
\addlinespace[2pt]
KAN & 2.74 & {\scriptsize [2.70, 2.80]} & 439.7 & {\scriptsize [423.4, 454.8]} & 1039.7 & {\scriptsize [992.0, 1083.3]} \\
S2KAN & 4.31 & {\scriptsize [4.26, 4.34]} & 58.2 & {\scriptsize [52.3, 60.5]} & 83.8 & {\scriptsize [78.7, 88.4]} \\
\addlinespace[5pt]
\multicolumn{7}{@{}l}{\textbf{Superconductor}} \\
\addlinespace[2pt]
KAN & 2.73 & {\scriptsize [2.47, 2.76]} & 440.9 & {\scriptsize [424.2, 452.9]} & 1089.6 & {\scriptsize [1050.0, 1101.2]} \\
S2KAN & 4.24 & {\scriptsize [3.89, 4.29]} & 300.5 & {\scriptsize [288.4, 376.2]} & 521.7 & {\scriptsize [404.5, 556.7]} \\
\bottomrule
\end{tabular}
\end{table*}

\clearpage

\section{Scientific consistency checks}
\label{sec:scientific-response-diagnostics}

This appendix gives the response definitions, secondary comparisons, and sensitivity checks supporting Sec.~\ref{subsec:scientific-insights}.
All analyses use the validation-selected models in Table~\ref{tab:real-world-both}: 30 fits per method and dataset, comprising ten data splits and three initialization seeds per split.
Both methods receive the same physical-input queries, transformed using each fit's preprocessing, and outputs are returned to the original units.
Checkpoints remain frozen, with no further training, checkpoint selection, or symbolic refitting.
We summarize each fit separately before reporting medians and, where given, first and third quartiles in brackets; split comparisons average the three seeds within each method.
These are descriptive summaries across overlapping splits, not confidence intervals from independent experiments.
The diagnostics are exploratory, evaluate selected response patterns and physical admissibility, and compare the complete training configurations.
Response thresholds specify the diagnostic tolerances and are not universal physical or engineering limits.

\subsection{Airfoil self-noise}
\label{sec:response-airfoil}

\paragraph{Post-peak decline and sensitivity.}
The 1,503 observations form 106 spectra defined by identical angle, chord, speed, and displacement thickness.
A spectrum is eligible for the decline diagnostic when it has at least four measurements from its observed maximum onward, a terminal decrease exceeding 3~dB, and no later-minus-earlier increase exceeding 1~dB.
These criteria select 92 spectra; the other spectra are excluded from this diagnostic without being classified as physically invalid.
Models are evaluated at 257 logarithmically spaced frequencies between the observed peak and the largest recorded frequency, holding the four other inputs fixed.
A pass requires an endpoint decrease exceeding 1~dB and a maximum later-minus-earlier increase no greater than 1~dB, with $10^{-6}$~dB numerical slack for the reversal comparison.
Flat responses fail the endpoint requirement.
The passing fractions are 70.1\% [41.0, 81.0] for KAN and 94.6\% [92.4, 95.7] for S2KAN.
Replacing both the required decrease and permitted reversal by 0.5~dB gives 44.6\% versus 88.0\%; using 2~dB gives 91.3\% versus 98.9\%, respectively.
All ten split averages favor S2KAN at each threshold.
Increasing the grid to 513 points changes no classifications at the primary 1~dB threshold or at 2~dB, and changes three of 5,520 model--spectrum classifications at 0.5~dB.
These checks establish numerical stability at the tested resolutions, not monotonicity between samples.
Eligibility uses observed target values across the corpus, including training observations.

\paragraph{Spectral shape and peak localization.}
For held-out spectral shape, we retain spectra with at least three test frequencies (89--99 spectra per split), subtract each predicted and observed spectrum's own test-row mean, and compute RMSE with equal weight per spectrum.
Median centered errors are 1.427 [1.380, 1.505]~dB for KAN and 1.404 [1.361, 1.480]~dB for S2KAN, with S2KAN lower in six splits.
The peak comparison uses 86 spectra whose observed maximum is interior and at least 3~dB above both endpoints.
Predicted peaks are located on a 257-point logarithmic grid spanning each spectrum's lowest to highest recorded frequency; exact ties select the lowest frequency.
Median mean absolute peak-frequency errors are 0.250 [0.238, 0.270] and 0.250 [0.230, 0.263] octaves for KAN and S2KAN.
The fractions within one-third octave of the measured peak are 72.7\% [68.0, 76.7] and 74.4\% [70.9, 77.9], respectively, favoring S2KAN in six splits.
Thus peak localization is similar.
Peak eligibility uses the full measured spectra; even the held-out shape errors assess interpolation among recorded conditions, since other frequencies at the same condition can occur in fitting data.

\paragraph{Contrasts between recorded speeds.}
Within identical angle/chord families, 73 adjacent-speed pairs have at least four shared measured frequency bands.
Each speed uses the displacement thickness supplied with the dataset, which is computed from the source's empirical relations for chord, speed, and angle of attack~\citep{brooks1989airfoil}.
The contrast is the arithmetic mean difference in the dataset's scaled sound pressure level (dB) across common frequency bands. It evaluates prediction of the supplied response across speeds; it does not directly quantify the change in unscaled sound pressure level or identify a velocity-scaling exponent.
For held-out evaluation, at least two common bands must have both speed endpoints in the test set, leaving 29--34 eligible pairs per split.
Median contrast mean absolute error (MAE) is 1.208 [1.128, 1.331]~dB for KAN and 1.023 [0.940, 1.292]~dB for S2KAN, with S2KAN lower in nine splits.
Direction agreement on held-out contrasts exceeding 1~dB in observed absolute magnitude is 89.7\% [84.6, 91.5] versus 94.7\% [84.4, 98.9], favoring S2KAN in seven splits.
Across the full corpus, however, direction agreement at the same magnitude threshold favors KAN: 95.7\% [91.3, 95.7] versus 91.3\% [89.7, 91.3], with KAN higher in eight splits, S2KAN higher in one, and one tie.
The speed comparisons therefore provide mixed evidence and do not establish generally better velocity dependence.

\subsection{Concrete compressive strength}
\label{sec:response-concrete}

The aging diagnostics use the 181 exact-composition recipes with at least two observed ages.
Each recipe receives equal weight and is evaluated on 257 equally spaced ages within its recorded range, varying raw age and the supplied $\log(\mathrm{age}+1)$ feature together.
The decline score is the largest earlier-to-later decrease anywhere on that grid; a recipe violates the criterion when this exceeds 1~MPa.
The slowing-growth criterion compares endpoint secant slopes over the first and last thirds of the grid: indices 0--85 and 171--256, respectively, with zero-based indexing.
A pass requires $0<s_{\mathrm{late}}<s_{\mathrm{early}}$.
This excludes a flat late interval and does not require the complete curve to be everywhere increasing or concave; a physical plateau can therefore fail this particular criterion.
The two scores describe related properties of the same model curves and do not constitute independent discoveries or held-out aging-accuracy tests.

On a one-day grid, the median fraction of recipes with declines exceeding 1~MPa is 8.3\% for S2KAN versus 62.7\% for KAN, favoring S2KAN in all ten splits under both mean and median aggregation.
Grids finer than the dataset's one-day age-recording resolution detect additional declines in both methods, including sub-hour dips from protected logarithmic activations in some S2KAN fits; the overall median decline frequency remains lower for S2KAN.

The concrete illustration in Fig.~\ref{fig:concrete-symbolic-aging} uses split and initialization seed 42.
The recipe was selected by proximity to the median standardized ingredient composition among the 181 eligible recipes, without strengths or model predictions.
All native activations, fixed-input offsets, target scaling, and protected operations are retained; the complete prediction includes all 32 branches.
The four displayed branches were ranked by total variation of their output contributions after extraction.
For the dominant branch, the increasing and concave contribution was verified throughout the displayed 3--365-day interval from its native coefficients.
This property of one contribution does not imply concavity of every branch or every fitted model.
Contributions from other branches introduce local increases in the rate of gain, including a corner near day~16 associated with a protected square root; the complete prediction nevertheless remains increasing on the evaluated grid.
This corner is an implementation feature, not an identified curing transition.

\subsection{Power plant electrical output}
\label{sec:response-powerplant}

\paragraph{Local support and directional criteria.}
We sampled 512 anchors from distinct recorded input vectors using inputs alone.
Temperature sweeps vary ambient temperature by $\pm2\,{}^{\circ}\mathrm{C}$, and pressure sweeps vary atmospheric pressure by $\pm4$~mbar, with the other coordinates fixed.
Each sweep has 65 equally spaced points.
We reject a sweep if any point leaves the corpus marginal ranges or its nearest-neighbor distance to a recorded input, in corpus-standardized four-dimensional input space, exceeds the 95th percentile of corpus leave-one-out nearest-neighbor distances.
This retains 325 temperature and 148 pressure sweeps.
A stricter 90th-percentile screen retains 212 and 71 sweeps, respectively.
Support uses all corpus inputs and is a geometric restriction, not proof of training support or an achievable controlled intervention.
In particular, holding relative humidity fixed while changing temperature does not hold absolute humidity fixed.

The expected directions are decreasing power with temperature and increasing power with pressure.
For either direction, a pass requires a net change exceeding 0.2~MW in that direction and no earlier-to-later excursion exceeding 1~MW in the opposite direction.
Temperature passing fractions are 98.2\% [93.8, 99.4] for KAN and 100.0\% [99.7, 100.0] for S2KAN.
The corresponding fractions with an upward excursion exceeding 1~MW are 1.23\% [0.00, 5.15] and 0.00\% [0.00, 0.00].
For pressure, passing fractions are 35.8\% [25.5, 45.6] and 54.1\% [47.8, 57.8], while wrong-way excursions exceeding 1~MW occur in 61.5\% [51.5, 71.6] and 24.7\% [5.4, 30.4] of sweeps, respectively.
Both primary endpoints favor S2KAN in all ten split averages for each variable, but substantial pressure-response failures remain.

\paragraph{Observed-data checks and sensitivity.}
To assess the directional evidence in recorded observations, we match each input to its nearest eligible partner in the other three standardized coordinates, with each coordinate difference at most 0.2 standard deviations and a focal difference of 1--4~$^{\circ}\mathrm{C}$ or 2--8~mbar.
Duplicate input vectors are averaged, and a 0.1-standard-deviation matching box provides a sensitivity check.
Temperature pairs show the expected decrease in 76.5--78.2\% of comparisons, whereas pressure pairs show the expected increase in only 51.3--53.6\%.
Pairs share observations and describe associations, not independent interventions.
Under the stricter support screen, temperature passing fractions are 98.35\% for KAN and 100.00\% for S2KAN; pressure passing fractions are 30.99\% and 46.48\%.
All ten split averages still favor S2KAN, but its median desired pressure net change in this subset is only 0.051~MW over the 8~mbar interval.
Fewer reversals therefore do not establish recovery of a nontrivial pressure law.
Doubling the evaluation grid to 129 points changes three KAN temperature classifications for excursions exceeding 1~MW, and none for S2KAN or pressure.
Across excursion thresholds of 0.1, 0.5, 1, and 2~MW and net-change thresholds of 0, 0.2, and 0.5~MW, favorable-split counts for the directional pass criteria range from eight to ten; the advantage is not universal for every sensitivity and split.

The symbolic temperature example uses split and initialization seed 42 at fixed exhaust vacuum 46.63~cm~Hg, atmospheric pressure 1013.16~mbar, and relative humidity 79.08\%.
The context was selected using inputs before inspecting the native activations; the polynomial-to-sine branch was subsequently chosen for readability.
All five branches remain in the prediction, including a small opposing contribution.
The reported branch and full-model decreases are exact endpoint contributions in MW, not measured intervention effects or evidence that a larger predicted decrease is more accurate.

\subsection{Superconductor critical temperature}
\label{sec:response-superconductor}

The physical-admissibility check measures the fraction of negative predicted critical temperatures among each fit's 1,000 untouched test observations, without clipping.
The clipping control replaces predictions by $\max(\hat T_c,0)$ without retraining, and computes test RMSE on the same observations.

\suppressfloats[t]
\section{Dynamical-system diagnostics}
\label{sec:ds-diagnostics}

These diagnostics assess recovery of dynamical behavior beyond one-step prediction accuracy for the models in Table~\ref{tab:ikeda_ecosystem}. Lyapunov estimation is described in \methods{}; additional evaluation procedures are given below.

\subsection{Lyapunov spectra}

\begin{table}
\centering
\caption{Full Lyapunov spectra of the models in Table~\ref{tab:ikeda_ecosystem}. Entries are medians [first quartile, third quartile] over defined spectra; ``Defined'' gives their count out of ten fits. The Ikeda map has two exponents, so $\lambda_3$ is inapplicable. Reference entries are finite-orbit estimates for the generating map. $|\Delta\lambda_1|$ is the absolute difference between leading-exponent estimates from the first and second halves of the evaluated orbit.}
\label{tab:ds-spectra}
\small
\begin{tabular}{@{}lrrrrrr@{}}
\toprule
Model & $\beta$ & Defined & $\lambda_1$ & $\lambda_2$ & $\lambda_3$ & $|\Delta\lambda_1|$ \\
\midrule
\multicolumn{7}{@{}l}{\textbf{Ikeda map}: exponents per iterate} \\
\addlinespace[2pt]
Reference & --- & --- & 0.508 & -0.719 & --- & --- \\
KAN & --- & \shortstack[r]{10/10\\[-1pt]{\fontsize{6}{7}\selectfont\vphantom{[0.0, 0.0]}}} & \shortstack[r]{0.508\\[-1pt]{\fontsize{6}{7}\selectfont [0.505, 0.511]}} & \shortstack[r]{-0.727\\[-1pt]{\fontsize{6}{7}\selectfont [-0.731, -0.722]}} & \shortstack[r]{---\\[-1pt]{\fontsize{6}{7}\selectfont\vphantom{[0.0, 0.0]}}} & \shortstack[r]{0.0059\\[-1pt]{\fontsize{6}{7}\selectfont [0.0040, 0.0094]}} \\
S2KAN & 0 & \shortstack[r]{10/10\\[-1pt]{\fontsize{6}{7}\selectfont\vphantom{[0.0, 0.0]}}} & \shortstack[r]{0.508\\[-1pt]{\fontsize{6}{7}\selectfont [0.506, 0.512]}} & \shortstack[r]{-0.725\\[-1pt]{\fontsize{6}{7}\selectfont [-0.731, -0.723]}} & \shortstack[r]{---\\[-1pt]{\fontsize{6}{7}\selectfont\vphantom{[0.0, 0.0]}}} & \shortstack[r]{0.0074\\[-1pt]{\fontsize{6}{7}\selectfont [0.0046, 0.0108]}} \\
S2KAN & 1 & \shortstack[r]{10/10\\[-1pt]{\fontsize{6}{7}\selectfont\vphantom{[0.0, 0.0]}}} & \shortstack[r]{0.508\\[-1pt]{\fontsize{6}{7}\selectfont [0.501, 0.512]}} & \shortstack[r]{-0.727\\[-1pt]{\fontsize{6}{7}\selectfont [-0.729, -0.725]}} & \shortstack[r]{---\\[-1pt]{\fontsize{6}{7}\selectfont\vphantom{[0.0, 0.0]}}} & \shortstack[r]{0.0068\\[-1pt]{\fontsize{6}{7}\selectfont [0.0037, 0.0143]}} \\
S2KAN & 3 & \shortstack[r]{10/10\\[-1pt]{\fontsize{6}{7}\selectfont\vphantom{[0.0, 0.0]}}} & \shortstack[r]{0.507\\[-1pt]{\fontsize{6}{7}\selectfont [0.502, 0.508]}} & \shortstack[r]{-0.742\\[-1pt]{\fontsize{6}{7}\selectfont [-0.756, -0.730]}} & \shortstack[r]{---\\[-1pt]{\fontsize{6}{7}\selectfont\vphantom{[0.0, 0.0]}}} & \shortstack[r]{0.0044\\[-1pt]{\fontsize{6}{7}\selectfont [0.0009, 0.0115]}} \\
\addlinespace[5pt]
\multicolumn{7}{@{}l}{\textbf{Ecosystem}: exponents per unit time} \\
\addlinespace[2pt]
Reference & --- & --- & 0.00814 & 0.00007 & -0.39132 & --- \\
KAN & --- & \shortstack[r]{10/10\\[-1pt]{\fontsize{6}{7}\selectfont\vphantom{[0.0, 0.0]}}} & \shortstack[r]{0.00666\\[-1pt]{\fontsize{6}{7}\selectfont [0.00092, 0.00861]}} & \shortstack[r]{-0.00013\\[-1pt]{\fontsize{6}{7}\selectfont [-0.00189, 0.00002]}} & \shortstack[r]{-0.29512\\[-1pt]{\fontsize{6}{7}\selectfont [-0.33723, -0.27225]}} & \shortstack[r]{0.00115\\[-1pt]{\fontsize{6}{7}\selectfont [0.00045, 0.00262]}} \\
S2KAN & 0 & \shortstack[r]{10/10\\[-1pt]{\fontsize{6}{7}\selectfont\vphantom{[0.0, 0.0]}}} & \shortstack[r]{0.00688\\[-1pt]{\fontsize{6}{7}\selectfont [0.00002, 0.00926]}} & \shortstack[r]{-0.00019\\[-1pt]{\fontsize{6}{7}\selectfont [-0.00099, 0.00002]}} & \shortstack[r]{-0.37177\\[-1pt]{\fontsize{6}{7}\selectfont [-0.38882, -0.33182]}} & \shortstack[r]{0.00081\\[-1pt]{\fontsize{6}{7}\selectfont [0.00052, 0.00193]}} \\
S2KAN & 0.1 & \shortstack[r]{10/10\\[-1pt]{\fontsize{6}{7}\selectfont\vphantom{[0.0, 0.0]}}} & \shortstack[r]{0.00664\\[-1pt]{\fontsize{6}{7}\selectfont [-0.00002, 0.00909]}} & \shortstack[r]{-0.00022\\[-1pt]{\fontsize{6}{7}\selectfont [-0.00316, -0.00001]}} & \shortstack[r]{-0.36529\\[-1pt]{\fontsize{6}{7}\selectfont [-0.38974, -0.35080]}} & \shortstack[r]{0.00065\\[-1pt]{\fontsize{6}{7}\selectfont [0.00046, 0.00072]}} \\
S2KAN & 1 & \shortstack[r]{9/10\\[-1pt]{\fontsize{6}{7}\selectfont\vphantom{[0.0, 0.0]}}} & \shortstack[r]{0.00001\\[-1pt]{\fontsize{6}{7}\selectfont [-0.00003, 0.00768]}} & \shortstack[r]{-0.00392\\[-1pt]{\fontsize{6}{7}\selectfont [-0.01525, -0.00031]}} & \shortstack[r]{-0.31852\\[-1pt]{\fontsize{6}{7}\selectfont [-0.35507, -0.10222]}} & \shortstack[r]{0.00041\\[-1pt]{\fontsize{6}{7}\selectfont [0.00023, 0.00050]}} \\
\bottomrule
\end{tabular}
\end{table}

Table~\ref{tab:ds-spectra} shows similar median leading Lyapunov exponents for all Ikeda configurations, close to the reference value 0.5081. All spectra are defined, but one S2KAN seed at $\beta=1$ has a negative leading exponent and incorrect long-run behavior. For the ecosystem, KAN and unregularized S2KAN have median leading exponents 0.00666 and 0.00688, respectively, versus the reference 0.00814; each has three nonpositive estimates. At $\beta=1$, S2KAN's median falls near zero and one trajectory diverges. Agreement of medians therefore does not imply recovery in every seed.

\subsection{Temporal structure}
To assess whether the learned models reproduce the reference systems\textquoteright{} temporal dependence and oscillatory behavior, we compare autocorrelation functions (ACFs) and power spectral densities (PSDs) on the model and reference orbits described in \methods{}.

Coordinate-wise PSDs use Welch's method with segment lengths 1,024 (Ikeda) or 4,096 (ecosystem), segment-mean subtraction, a periodic Hann window, 50\% overlap, and one-sided density scaling. ACFs use the demeaned sample covariance with denominator $N$, the number of samples in the analysis window (10,000 for Ikeda and 20,000 for the ecosystem), normalized at lag zero, through 100 iterates or 1,000 time units, respectively. Each diagnostic uses the second half of a completed model orbit; pointwise medians and IQRs summarize seeds, with PSD quantiles computed in linear power. Constant-coordinate ACFs are undefined and excluded from ACF summaries; this affects one ecosystem S2KAN fit at $\beta=1$, whose PSD remains included. Reference curves average diagnostics over ten 10,000-point Ikeda blocks or five 20,000-point ecosystem blocks from the reference orbit's final 100,000 points, matching each model's analysis-window length. Frequencies use sampling intervals 1 and $\Delta t=0.5$, respectively; only the zero-frequency bin is omitted from the logarithmic PSD display.

\begin{figure*}%
\centering
\includegraphics[width=\linewidth,height=0.78\textheight,keepaspectratio]{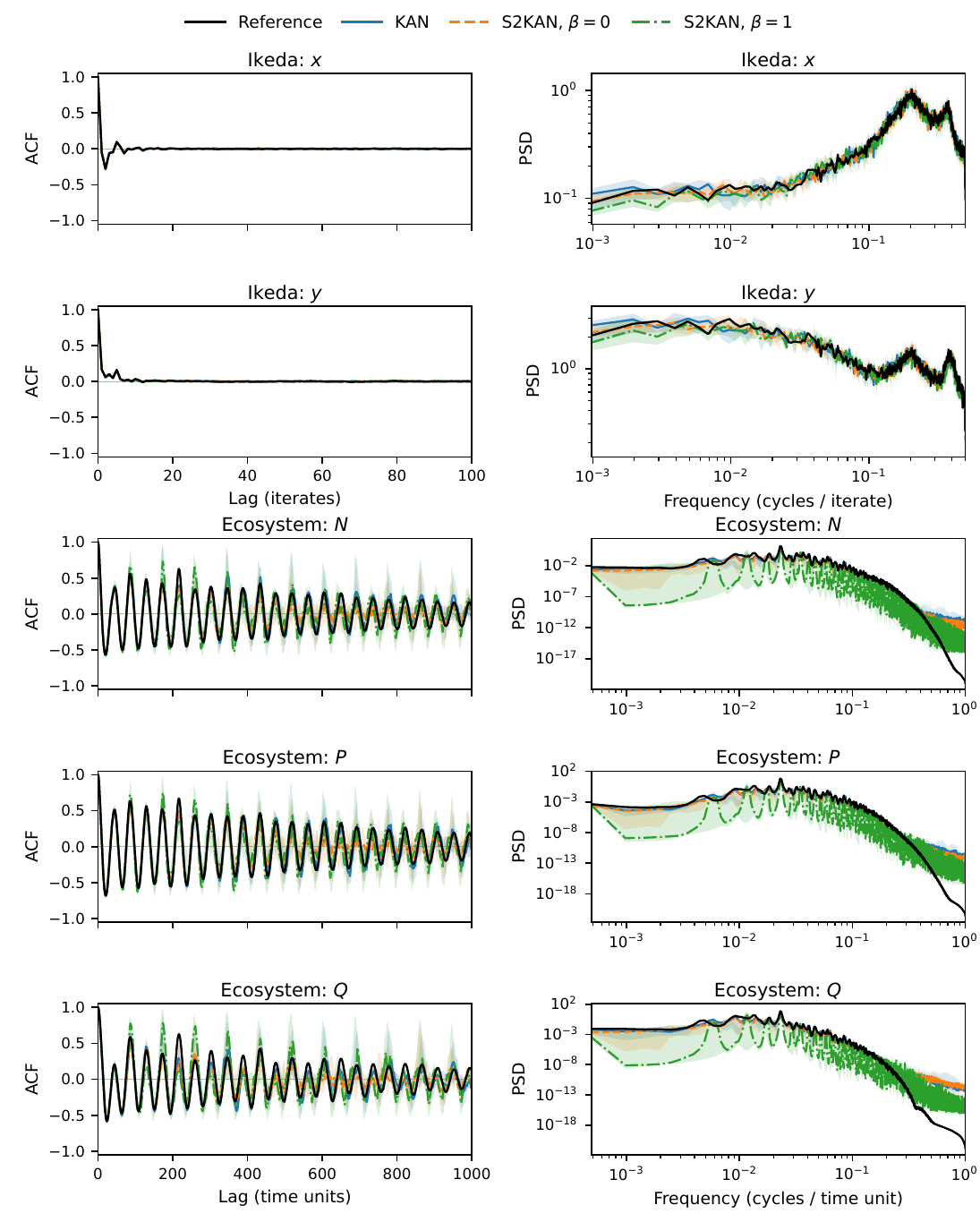}
\caption{Temporal structure of the learned dynamical systems measured using autocorrelation functions (left) and power spectral densities (right), for each state coordinate. Colored curves and shading show pointwise medians and IQRs across completed fits; black curves are reference block means. All shown Ikeda, KAN, and $\beta=0$ configurations contribute 10/10 fits. Ecosystem S2KAN at $\beta=1$ contributes nine PSDs and eight ACFs: one trajectory diverges and another is constant, making its ACF undefined. Other completed trajectories remain included.}
\label{fig:ds-temporal}
\end{figure*}

Figure~\ref{fig:ds-temporal} shows broadly similar median ACFs and PSDs for Ikeda. Ecosystem KAN and unregularized S2KAN retain the broad reference oscillatory patterns, whereas S2KAN at $\beta=1$ has more sharply peaked spectra and less broadband low-frequency power. Of the 60 trajectories represented in Fig.~\ref{fig:ds-temporal}, one ecosystem trajectory at $\beta=1$ diverges and another converges to a constant state. Completed trajectories remain included wherever the diagnostic is defined, so these median curves should be read alongside the seed variation and exceptions above.

\subsection{Forecasts and long-run state distributions}
Multi-step RMSE averages coordinate-wise RMSEs over 2,000 Ikeda steps or 1,250 ecosystem steps from the first unseen state; divergent rollouts yield undefined RMSE. This differs from pooling coordinates before taking the square root for one-step RMSE. Forecast horizons use 20 evenly spaced initial states with a full evaluation window in the unseen trajectory. State errors are divided coordinate-wise by reference standard deviations over the same 2,000- or 1,250-step unseen window (plus $10^{-12}$). Each horizon counts the steps before the root-mean-square standardized error first exceeds 0.2 or becomes nonfinite, capped at 200 steps (Ikeda) or 500 (ecosystem); the fit's horizon is the median over the 20 starts.

Late density overlap pools samples from the final third of 20 rollouts started across the unseen block, each lasting 6,000 steps for Ikeda or 3,750 for the ecosystem. The reference ``seen block'' comprises the training and test partitions. Overlap is $\sum_b\min(p_b,q_b)$ for model and seen-block reference joint histogram masses, using 25 or 10 bins per coordinate, respectively, on seen-block bounds padded by 5\%. Normalization retains all requested samples; nonfinite or out-of-box mass is lost. Nonfinite states or magnitudes above $10^6$ terminate trajectories. Initial states are not independent model replicates.

Table~\ref{tab:ds-rollouts} complements pointwise errors with forecast horizons and late density overlap. Reference-map overlaps are 0.916 for Ikeda and 0.903 for the ecosystem, illustrating finite-sample differences even between reference trajectories. Multi-step RMSE is defined for nine ecosystem S2KAN fits at $\beta=1$ and all ten fits otherwise; all fits contribute to horizon and overlap. Nine of the 200 late-evaluation rollouts diverge for ecosystem S2KAN at $\beta=1$, with none elsewhere. These counts differ from the spectral evaluation because the initial states and evaluation horizons differ.

\begin{table}
\centering
\caption{Rollout diagnostics for the models in Table~\ref{tab:ikeda_ecosystem}. Entries are medians [first quartile, third quartile] over ten fits, with undefined multi-step RMSEs excluded as described in the text. Horizon is in discrete steps and late overlap lies in $[0,1]$.}
\label{tab:ds-rollouts}
\begingroup
\small
\setlength{\tabcolsep}{4pt}
\renewcommand{\arraystretch}{1.13}
\begin{tabular}{@{}lrrrr@{}}
\toprule
Model & $\beta$ & Multi-step RMSE & Horizon & Late overlap \\
\midrule
\multicolumn{5}{@{}l}{\textbf{Ikeda map}} \\
\addlinespace[2pt]
KAN & --- & \shortstack[r]{0.858\\[-1pt]{\fontsize{6}{7}\selectfont [0.851, 0.871]}} & \shortstack[r]{9.0\\[-1pt]{\fontsize{6}{7}\selectfont [8.5, 9.0]}} & \shortstack[r]{0.913\\[-1pt]{\fontsize{6}{7}\selectfont [0.910, 0.914]}} \\
S2KAN & 0 & \shortstack[r]{0.866\\[-1pt]{\fontsize{6}{7}\selectfont [0.856, 0.876]}} & \shortstack[r]{9.0\\[-1pt]{\fontsize{6}{7}\selectfont [8.0, 10.0]}} & \shortstack[r]{0.913\\[-1pt]{\fontsize{6}{7}\selectfont [0.909, 0.914]}} \\
S2KAN & 1 & \shortstack[r]{0.861\\[-1pt]{\fontsize{6}{7}\selectfont [0.855, 0.871]}} & \shortstack[r]{8.5\\[-1pt]{\fontsize{6}{7}\selectfont [7.6, 9.0]}} & \shortstack[r]{0.909\\[-1pt]{\fontsize{6}{7}\selectfont [0.907, 0.912]}} \\
S2KAN & 3 & \shortstack[r]{0.873\\[-1pt]{\fontsize{6}{7}\selectfont [0.863, 0.875]}} & \shortstack[r]{7.2\\[-1pt]{\fontsize{6}{7}\selectfont [6.5, 7.5]}} & \shortstack[r]{0.904\\[-1pt]{\fontsize{6}{7}\selectfont [0.897, 0.910]}} \\
\addlinespace[5pt]
\multicolumn{5}{@{}l}{\textbf{Ecosystem}} \\
\addlinespace[2pt]
KAN & --- & \shortstack[r]{0.130\\[-1pt]{\fontsize{6}{7}\selectfont [0.120, 0.147]}} & \shortstack[r]{270.8\\[-1pt]{\fontsize{6}{7}\selectfont [261.9, 408.5]}} & \shortstack[r]{0.831\\[-1pt]{\fontsize{6}{7}\selectfont [0.784, 0.911]}} \\
S2KAN & 0 & \shortstack[r]{0.093\\[-1pt]{\fontsize{6}{7}\selectfont [0.060, 0.122]}} & \shortstack[r]{396.2\\[-1pt]{\fontsize{6}{7}\selectfont [308.6, 464.4]}} & \shortstack[r]{0.818\\[-1pt]{\fontsize{6}{7}\selectfont [0.745, 0.869]}} \\
S2KAN & 0.1 & \shortstack[r]{0.080\\[-1pt]{\fontsize{6}{7}\selectfont [0.075, 0.134]}} & \shortstack[r]{313.8\\[-1pt]{\fontsize{6}{7}\selectfont [294.6, 385.9]}} & \shortstack[r]{0.841\\[-1pt]{\fontsize{6}{7}\selectfont [0.784, 0.871]}} \\
S2KAN & 1 & \shortstack[r]{0.138\\[-1pt]{\fontsize{6}{7}\selectfont [0.131, 0.171]}} & \shortstack[r]{177.5\\[-1pt]{\fontsize{6}{7}\selectfont [94.5, 222.5]}} & \shortstack[r]{0.674\\[-1pt]{\fontsize{6}{7}\selectfont [0.309, 0.729]}} \\
\bottomrule
\end{tabular}
 \endgroup
\end{table}

For Ikeda, S2KAN at $\beta=1$ has a slightly shorter median forecast horizon than KAN (8.5 versus 9 steps) and similar late overlap (0.909 versus 0.913). For the ecosystem, unregularized S2KAN lowers multi-step RMSE in nine of ten paired seeds and lengthens the horizon in seven, although its median late overlap is slightly lower (0.818 versus 0.831). At $\beta=1$, ecosystem overlap falls to 0.674. Thus smaller models can retain useful dynamical behavior, but the appropriate regularization strength depends on the system and the property being evaluated.

\subsection{Self-sparsification under the tested training configurations}
\label{subsec:self-sparsification}

\begin{figure*}[t]%
\begin{center}
\includegraphics[width=0.47\textwidth]{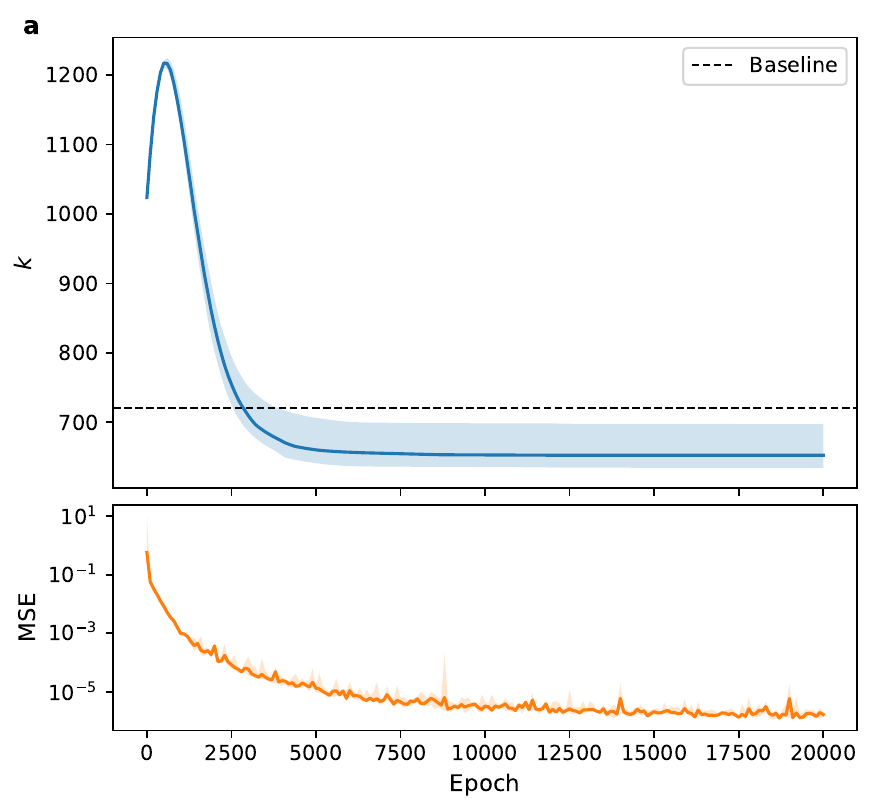}%
\includegraphics[width=0.47\textwidth]{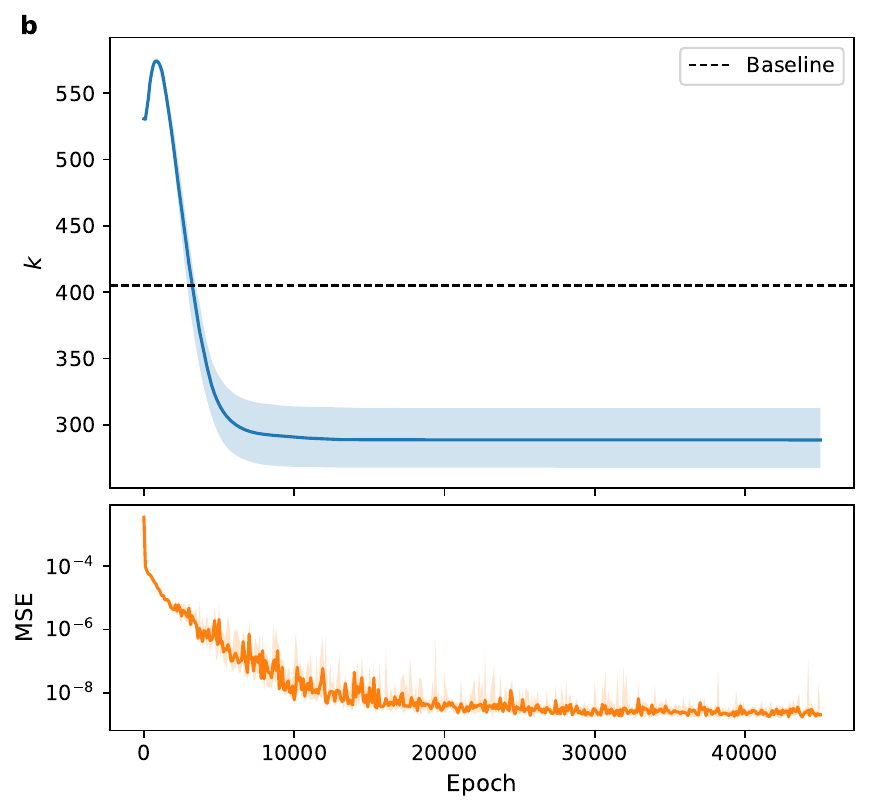}
\caption{Self-sparsification of the Ikeda map (a) and Ecosystem networks (b), both at $\beta=0$.
Solid curves show medians across ten seeds; shading indicates the interquartile range.
Horizontal dashed lines indicate the effective parameter count $\ktot$ of baseline KAN. Counts are expected effective parameter counts, rather than hard gate or active-edge counts. The first point follows the first completed epoch. Training MSEs are epoch-averaged stochastic minibatch values; shading shows descriptive interquartile ranges, not confidence intervals.
\label{fig:self-sparsifying}}
\end{center}
\end{figure*}

For the $\beta=0$ fits in Table~\ref{tab:ikeda_ecosystem}, we tracked the expected effective parameter count $k$ throughout training.
In both systems, its median rises early and then falls below the baseline KAN count (Fig.~\ref{fig:self-sparsifying}).
Median counts decrease from 1,023.5 after the first epoch to 652.5 at the fixed 20,000-epoch endpoint for Ikeda, compared with 720 for baseline KAN.
For the ecosystem, the corresponding medians are 530.7 and 288.5 at the fixed 45,000-epoch endpoint, compared with 405 for baseline KAN.
The first-epoch count follows one completed epoch and is not the untrained initialization.
All ten fits per system finish below their first-epoch counts; nine of ten Ikeda fits and all ten ecosystem fits finish below baseline KAN.
The remaining Ikeda fit ends at $k=720.049$, slightly above the baseline count.
Thus shrinkage during training and a final size below baseline KAN are distinct observations; the latter holds at the median for both systems.
Training continued to fixed endpoints without early stopping.
These results document self-sparsification under the tested configurations, without establishing its mechanism or its robustness to changes in optimizer, gate parameterization, dictionary, architecture, or data.

\section{Nguyen benchmark variability}
\label{sec:nguyen-variability}

For completeness, Table~\ref{tab:nguyen-uncertainty} reports seed variability for the full Nguyen comparison in Table~\ref{tab:nguyen-full}, using the same fits for KAN and S2KAN across architectures S, L and LM.

\begin{table}[p]
\centering\singlespacing{}
\caption{Seed variability for Table~\ref{tab:nguyen-full}. Entries are [first, third quartiles] over ten initialization seeds; S2KAN uses $\beta=1$.}
\label{tab:nguyen-uncertainty}
\begingroup\fontsize{8}{9}\selectfont
\setlength{\tabcolsep}{2pt}\renewcommand{\arraystretch}{1.02}
\begin{adjustbox}{max width=\linewidth}
\begin{tabular}{@{}lllrrrrrrrrrr@{}}
\toprule
 & & & \multicolumn{2}{c}{Test accuracy} & \multicolumn{2}{c}{Model size} & \multicolumn{3}{c}{SF (\%), $\epsilon=0.01$} & \multicolumn{3}{c}{SF (\%), $\epsilon=0.1$} \\
\cmidrule(lr){4-5}\cmidrule(lr){6-7}\cmidrule(lr){8-10}\cmidrule(lr){11-13}
ID & Shape & Model & $R^2$ & RMSE & Active & $k$ & $J=1$ & $J=2$ & $J=3$ & $J=1$ & $J=2$ & $J=3$ \\
\midrule
F1 & S & KAN & $[1.0000,1.0000]$ & $[0.20,1.99]\!\times\!10^{-5}$ & $[1,1]$ & $[15,15]$ & $[0,0]$ & $[100,100]$ & $[100,100]$ & $[0,0]$ & $[100,100]$ & $[100,100]$ \\
 &  & S2KAN & $[1.0000,1.0000]$ & $[0.58,1.24]\!\times\!10^{-3}$ & $[1,1]$ & $[20.1,21.1]$ & $[0,0]$ & $[100,100]$ & $[100,100]$ & $[0,0]$ & $[100,100]$ & $[100,100]$ \\
 & L & KAN & $[1.0000,1.0000]$ & $[4.27,7.10]\!\times\!10^{-4}$ & $[6,6]$ & $[90,90]$ & $[0,12.5]$ & $[0,12.5]$ & $[4.2,31.2]$ & $[27.1,33.3]$ & $[61.7,95.8]$ & $[100,100]$ \\
 &  & S2KAN & $[0.9999,1.0000]$ & $[3.30,8.20]\!\times\!10^{-3}$ & $[6,6]$ & $[38.4,51.3]$ & $[21.3,40]$ & $[76.2,95]$ & $[100,100]$ & $[50,60]$ & $[100,100]$ & $[100,100]$ \\
 & LM & KAN & $[1.0000,1.0000]$ & $[3.36,5.99]\!\times\!10^{-4}$ & $[9,9]$ & $[135,135]$ & $[0,0]$ & $[0,0]$ & $[11.5,24.3]$ & $[25,44.4]$ & $[65.6,77.8]$ & $[87.5,100]$ \\
 &  & S2KAN & $[1.0000,1.0000]$ & $[2.96,5.78]\!\times\!10^{-3}$ & $[9,9]$ & $[57.2,67.9]$ & $[28.6,42.9]$ & $[100,100]$ & $[100,100]$ & $[57.1,69.2]$ & $[100,100]$ & $[100,100]$ \\
\addlinespace[2pt]
F2 & S & KAN & $[1.0000,1.0000]$ & $[1.00,1.59]\!\times\!10^{-4}$ & $[1,1]$ & $[15,15]$ & $[0,0]$ & $[100,100]$ & $[100,100]$ & $[0,0]$ & $[100,100]$ & $[100,100]$ \\
 &  & S2KAN & $[1.0000,1.0000]$ & $[0.44,1.63]\!\times\!10^{-3}$ & $[1,1]$ & $[18.5,19.5]$ & $[0,0]$ & $[100,100]$ & $[100,100]$ & $[0,0]$ & $[100,100]$ & $[100,100]$ \\
 & L & KAN & $[1.0000,1.0000]$ & $[3.63,6.54]\!\times\!10^{-4}$ & $[6,6]$ & $[90,90]$ & $[0,12.5]$ & $[16.7,33.3]$ & $[33.3,50]$ & $[33.3,50]$ & $[70.8,100]$ & $[100,100]$ \\
 &  & S2KAN & $[0.9999,1.0000]$ & $[4.50,9.98]\!\times\!10^{-3}$ & $[6,6]$ & $[40,45.4]$ & $[52.5,72.9]$ & $[80,100]$ & $[100,100]$ & $[76.2,100]$ & $[100,100]$ & $[100,100]$ \\
 & LM & KAN & $[1.0000,1.0000]$ & $[2.96,6.12]\!\times\!10^{-4}$ & $[9,9]$ & $[135,135]$ & $[0,0]$ & $[0,11.1]$ & $[16.5,33.3]$ & $[29.8,42.7]$ & $[67.9,77.8]$ & $[78.1,100]$ \\
 &  & S2KAN & $[1.0000,1.0000]$ & $[4.73,5.59]\!\times\!10^{-3}$ & $[6.2,9]$ & $[39.5,58.3]$ & $[42.5,66.7]$ & $[100,100]$ & $[100,100]$ & $[80.8,100]$ & $[100,100]$ & $[100,100]$ \\
\addlinespace[2pt]
F3 & S & KAN & $[1.0000,1.0000]$ & $[2.23,2.74]\!\times\!10^{-4}$ & $[1,1]$ & $[15,15]$ & $[0,0]$ & $[0,0]$ & $[100,100]$ & $[0,0]$ & $[100,100]$ & $[100,100]$ \\
 &  & S2KAN & $[1.0000,1.0000]$ & $[4.35,7.44]\!\times\!10^{-4}$ & $[1,1]$ & $[19.3,20.1]$ & $[0,0]$ & $[0,0]$ & $[100,100]$ & $[0,0]$ & $[100,100]$ & $[100,100]$ \\
 & L & KAN & $[1.0000,1.0000]$ & $[0.47,1.21]\!\times\!10^{-3}$ & $[6,6]$ & $[90,90]$ & $[0,0]$ & $[0,16.7]$ & $[16.7,33.3]$ & $[33.3,33.3]$ & $[66.7,83.3]$ & $[100,100]$ \\
 &  & S2KAN & $[0.9999,1.0000]$ & $[0.43,1.20]\!\times\!10^{-2}$ & $[6,6]$ & $[53.2,59.5]$ & $[28.7,40]$ & $[75,80]$ & $[80,95.8]$ & $[42.5,60]$ & $[85,100]$ & $[100,100]$ \\
 & LM & KAN & $[1.0000,1.0000]$ & $[2.35,7.37]\!\times\!10^{-4}$ & $[9,9]$ & $[135,135]$ & $[0,0]$ & $[0,12.2]$ & $[2.8,24.3]$ & $[12.5,31.2]$ & $[57.3,75]$ & $[78.1,100]$ \\
 &  & S2KAN & $[1.0000,1.0000]$ & $[3.19,8.40]\!\times\!10^{-3}$ & $[9,9]$ & $[63.9,70]$ & $[20.8,42.9]$ & $[83.9,100]$ & $[100,100]$ & $[42.9,55.4]$ & $[100,100]$ & $[100,100]$ \\
\addlinespace[2pt]
F4 & S & KAN & $[1.0000,1.0000]$ & $[5.46,6.05]\!\times\!10^{-4}$ & $[1,1]$ & $[15,15]$ & $[0,0]$ & $[0,0]$ & $[100,100]$ & $[0,0]$ & $[100,100]$ & $[100,100]$ \\
 &  & S2KAN & $[1.0000,1.0000]$ & $[6.66,9.18]\!\times\!10^{-4}$ & $[1,1]$ & $[19.1,20]$ & $[0,0]$ & $[0,0]$ & $[100,100]$ & $[0,0]$ & $[100,100]$ & $[100,100]$ \\
 & L & KAN & $[1.0000,1.0000]$ & $[0.74,1.78]\!\times\!10^{-3}$ & $[6,6]$ & $[90,90]$ & $[0,0]$ & $[0,12.5]$ & $[4.2,33.3]$ & $[16.7,33.3]$ & $[52.5,79.2]$ & $[70,100]$ \\
 &  & S2KAN & $[1.0000,1.0000]$ & $[4.35,9.11]\!\times\!10^{-3}$ & $[5.2,6]$ & $[46.6,59]$ & $[50,60]$ & $[76.2,80]$ & $[80,100]$ & $[75,80]$ & $[100,100]$ & $[100,100]$ \\
 & LM & KAN & $[1.0000,1.0000]$ & $[4.88,7.92]\!\times\!10^{-4}$ & $[9,9]$ & $[135,135]$ & $[0,0]$ & $[0,12.5]$ & $[15.6,31.2]$ & $[22.9,37.5]$ & $[62.5,77.1]$ & $[75.7,96.9]$ \\
 &  & S2KAN & $[0.9999,1.0000]$ & $[0.41,1.24]\!\times\!10^{-2}$ & $[9,9]$ & $[55.1,64.4]$ & $[7.1,53.6]$ & $[89.3,100]$ & $[100,100]$ & $[57.1,68.6]$ & $[100,100]$ & $[100,100]$ \\
\addlinespace[2pt]
F5 & S & KAN & $[1.0000,1.0000]$ & $[4.50,7.09]\!\times\!10^{-5}$ & $[1,1]$ & $[15,15]$ & $[0,0]$ & $[100,100]$ & $[100,100]$ & $[100,100]$ & $[100,100]$ & $[100,100]$ \\
 &  & S2KAN & $[0.9998,1.0000]$ & $[0.54,2.35]\!\times\!10^{-3}$ & $[1,1]$ & $[19.1,19.9]$ & $[0,0]$ & $[100,100]$ & $[100,100]$ & $[100,100]$ & $[100,100]$ & $[100,100]$ \\
 & L & KAN & $[1.0000,1.0000]$ & $[0.92,2.43]\!\times\!10^{-4}$ & $[6,6]$ & $[90,90]$ & $[0,19.2]$ & $[5,38.3]$ & $[17.5,50]$ & $[23.3,50]$ & $[61.7,82.5]$ & $[100,100]$ \\
 &  & S2KAN & $[0.8682,0.9856]$ & $[1.99,6.02]\!\times\!10^{-2}$ & $[6,6]$ & $[57.5,58.7]$ & $[52.5,75]$ & $[100,100]$ & $[100,100]$ & $[75,100]$ & $[100,100]$ & $[100,100]$ \\
 & LM & KAN & $[1.0000,1.0000]$ & $[0.72,1.87]\!\times\!10^{-4}$ & $[9,9]$ & $[135,135]$ & $[0,12.5]$ & $[11.5,21.9]$ & $[14.9,34.4]$ & $[15.6,37.5]$ & $[53.1,66.7]$ & $[78.1,100]$ \\
 &  & S2KAN & $[0.2745,0.9171]$ & $[0.47,1.37]\!\times\!10^{-1}$ & $[7,9]$ & $[65.9,77.3]$ & $[66.7,84.3]$ & $[100,100]$ & $[100,100]$ & $[83.3,100]$ & $[100,100]$ & $[100,100]$ \\
\addlinespace[2pt]
F6 & S & KAN & $[1.0000,1.0000]$ & $[6.37,9.51]\!\times\!10^{-5}$ & $[1,1]$ & $[15,15]$ & $[0,0]$ & $[100,100]$ & $[100,100]$ & $[100,100]$ & $[100,100]$ & $[100,100]$ \\
 &  & S2KAN & $[1.0000,1.0000]$ & $[0.81,1.97]\!\times\!10^{-3}$ & $[1,1]$ & $[23.2,24.1]$ & $[0,0]$ & $[100,100]$ & $[100,100]$ & $[100,100]$ & $[100,100]$ & $[100,100]$ \\
 & L & KAN & $[1.0000,1.0000]$ & $[2.16,5.42]\!\times\!10^{-4}$ & $[6,6]$ & $[90,90]$ & $[0,0]$ & $[16.7,33.3]$ & $[35,50]$ & $[52.5,76.7]$ & $[87.5,100]$ & $[100,100]$ \\
 &  & S2KAN & $[1.0000,1.0000]$ & $[2.06,4.10]\!\times\!10^{-3}$ & $[6,6]$ & $[37.1,44.2]$ & $[65,95]$ & $[100,100]$ & $[100,100]$ & $[85,100]$ & $[100,100]$ & $[100,100]$ \\
 & LM & KAN & $[1.0000,1.0000]$ & $[2.13,6.04]\!\times\!10^{-4}$ & $[9,9]$ & $[135,135]$ & $[0,0]$ & $[11.1,12.5]$ & $[22.9,41.7]$ & $[34.4,42.7]$ & $[75,85.1]$ & $[87.8,100]$ \\
 &  & S2KAN & $[0.9999,1.0000]$ & $[5.15,9.13]\!\times\!10^{-3}$ & $[9,9]$ & $[55.2,63.7]$ & $[58.5,74.1]$ & $[100,100]$ & $[100,100]$ & $[85.7,87.5]$ & $[100,100]$ & $[100,100]$ \\
\addlinespace[2pt]
F7 & S & KAN & $[1.0000,1.0000]$ & $[0.91,3.49]\!\times\!10^{-5}$ & $[1,1]$ & $[15,15]$ & $[100,100]$ & $[100,100]$ & $[100,100]$ & $[100,100]$ & $[100,100]$ & $[100,100]$ \\
 &  & S2KAN & $[0.9999,1.0000]$ & $[1.15,6.63]\!\times\!10^{-3}$ & $[1,1]$ & $[11.3,21.2]$ & $[100,100]$ & $[100,100]$ & $[100,100]$ & $[100,100]$ & $[100,100]$ & $[100,100]$ \\
 & L & KAN & $[1.0000,1.0000]$ & $[2.59,9.16]\!\times\!10^{-4}$ & $[6,6]$ & $[90,90]$ & $[0,0]$ & $[0,16.7]$ & $[23.3,57.5]$ & $[42.5,65]$ & $[80,83.3]$ & $[87.5,100]$ \\
 &  & S2KAN & $[0.9988,0.9999]$ & $[0.88,2.87]\!\times\!10^{-2}$ & $[6,6]$ & $[55.5,57.4]$ & $[50,60]$ & $[100,100]$ & $[100,100]$ & $[63.7,100]$ & $[100,100]$ & $[100,100]$ \\
 & LM & KAN & $[1.0000,1.0000]$ & $[2.57,5.04]\!\times\!10^{-4}$ & $[9,9]$ & $[135,135]$ & $[0,0]$ & $[0,12.2]$ & $[22.2,46.9]$ & $[27.1,58]$ & $[75,88.9]$ & $[100,100]$ \\
 &  & S2KAN & $[0.9947,0.9994]$ & $[1.69,5.99]\!\times\!10^{-2}$ & $[7.5,9]$ & $[69.3,76.1]$ & $[51.8,71.4]$ & $[100,100]$ & $[100,100]$ & $[71.4,85.7]$ & $[100,100]$ & $[100,100]$ \\
\addlinespace[2pt]
F8 & S & KAN & $[1.0000,1.0000]$ & $[1.82,1.84]\!\times\!10^{-3}$ & $[1,1]$ & $[15,15]$ & $[0,0]$ & $[0,0]$ & $[0,0]$ & $[100,100]$ & $[100,100]$ & $[100,100]$ \\
 &  & S2KAN & $[1.0000,1.0000]$ & $[1.87,2.32]\!\times\!10^{-3}$ & $[1,1]$ & $[19.1,19.9]$ & $[0,0]$ & $[0,0]$ & $[0,0]$ & $[100,100]$ & $[100,100]$ & $[100,100]$ \\
 & L & KAN & $[1.0000,1.0000]$ & $[2.39,5.70]\!\times\!10^{-4}$ & $[6,6]$ & $[90,90]$ & $[0,0]$ & $[0,22.9]$ & $[17.5,40]$ & $[42.5,57.5]$ & $[80,95.8]$ & $[87.5,100]$ \\
 &  & S2KAN & $[0.9981,0.9997]$ & $[0.74,2.03]\!\times\!10^{-2}$ & $[6,6]$ & $[48.6,54.2]$ & $[20,57.5]$ & $[63.7,80]$ & $[100,100]$ & $[52.5,95]$ & $[100,100]$ & $[100,100]$ \\
 & LM & KAN & $[1.0000,1.0000]$ & $[1.68,2.85]\!\times\!10^{-4}$ & $[9,9]$ & $[135,135]$ & $[0,8.3]$ & $[0,12.2]$ & $[12.9,25]$ & $[29.8,48.6]$ & $[75,88.5]$ & $[87.5,100]$ \\
 &  & S2KAN & $[0.9985,0.9997]$ & $[0.87,1.79]\!\times\!10^{-2}$ & $[7.2,9]$ & $[52.7,63.1]$ & $[21.3,48.2]$ & $[72.3,95.8]$ & $[100,100]$ & $[61.7,71.4]$ & $[100,100]$ & $[100,100]$ \\
\addlinespace[2pt]
F9 & S & KAN & $[1.0000,1.0000]$ & $[5.20,9.67]\!\times\!10^{-5}$ & $[2,2]$ & $[30,30]$ & $[50,50]$ & $[100,100]$ & $[100,100]$ & $[100,100]$ & $[100,100]$ & $[100,100]$ \\
 &  & S2KAN & $[1.0000,1.0000]$ & $[0.59,1.84]\!\times\!10^{-3}$ & $[2,2]$ & $[22.2,36.2]$ & $[50,50]$ & $[100,100]$ & $[100,100]$ & $[100,100]$ & $[100,100]$ & $[100,100]$ \\
 & L & KAN & $[1.0000,1.0000]$ & $[1.46,4.07]\!\times\!10^{-4}$ & $[9,9]$ & $[135,135]$ & $[0,27.7]$ & $[28.6,62.5]$ & $[57.1,90.6]$ & $[57.9,96.4]$ & $[85.7,100]$ & $[100,100]$ \\
 &  & S2KAN & $[0.9999,1.0000]$ & $[2.89,4.43]\!\times\!10^{-3}$ & $[9,9]$ & $[39.8,43.7]$ & $[70.8,100]$ & $[100,100]$ & $[100,100]$ & $[87.5,100]$ & $[100,100]$ & $[100,100]$ \\
 & LM & KAN & $[1.0000,1.0000]$ & $[2.67,6.28]\!\times\!10^{-4}$ & $[14,14]$ & $[210,210]$ & $[0,5.8]$ & $[7.9,16.3]$ & $[40,49]$ & $[44.7,73.6]$ & $[91.7,100]$ & $[100,100]$ \\
 &  & S2KAN & $[0.9999,1.0000]$ & $[2.23,5.18]\!\times\!10^{-3}$ & $[13,14]$ & $[49.5,53.3]$ & $[68.8,88.1]$ & $[100,100]$ & $[100,100]$ & $[80.2,97.5]$ & $[100,100]$ & $[100,100]$ \\
\addlinespace[2pt]
F10 & S & KAN & $[-0.0374,-0.0372]$ & $[1.01,1.01]$ & $[2,2]$ & $[30,30]$ & $[0,0]$ & $[0,0]$ & $[0,0]$ & $[0,0]$ & $[0,0]$ & $[0,0]$ \\
 &  & S2KAN\textsuperscript{a} & $[-0.0006,-0.0006]$ & $[9.90,9.90]\!\times\!10^{-1}$ & $[0,0]$ & $[0,0]$ & $[100,100]$ & $[100,100]$ & $[100,100]$ & $[100,100]$ & $[100,100]$ & $[100,100]$ \\
 & L & KAN & $[0.9999,1.0000]$ & $[0.40,1.01]\!\times\!10^{-2}$ & $[9,9]$ & $[135,135]$ & $[0,0]$ & $[25,52.8]$ & $[68.8,88.1]$ & $[25,62.5]$ & $[80.6,100]$ & $[100,100]$ \\
 &  & S2KAN & $[0.9999,1.0000]$ & $[3.12,6.96]\!\times\!10^{-3}$ & $[9,9]$ & $[20.1,24.1]$ & $[55.6,96.9]$ & $[100,100]$ & $[100,100]$ & $[87.5,100]$ & $[100,100]$ & $[100,100]$ \\
 & LM & KAN & $[1.0000,1.0000]$ & $[3.07,5.05]\!\times\!10^{-3}$ & $[14,14]$ & $[210,210]$ & $[0,0]$ & $[7.3,21.4]$ & $[37.5,68]$ & $[21.4,48.2]$ & $[73.2,90.7]$ & $[92.9,100]$ \\
 &  & S2KAN & $[1.0000,1.0000]$ & $[1.39,2.40]\!\times\!10^{-3}$ & $[8,10]$ & $[24.5,29.3]$ & $[68.8,100]$ & $[100,100]$ & $[100,100]$ & $[100,100]$ & $[100,100]$ & $[100,100]$ \\
\bottomrule
\end{tabular}
 \end{adjustbox}
\endgroup
\par\smallskip
\begin{minipage}{\linewidth}
\footnotesize\raggedright{}
\textsuperscript{a} All ten S2KAN fits for F10 with shape S have no functionally active edges; SF is 100\% by convention. Metric and shape definitions follow Table~\ref{tab:nguyen-full}.
\end{minipage}
\end{table}

\section{Ablations and controls}
\label{sec:ablations-controls}

\subsection{Baseline KAN loss control}
\label{sec:kan-loss-control}

To separate the choice of training loss from the model comparison, we retrained baseline KAN using either raw MSE or $\tfrac12\log(\mathrm{MSE}+c_{\mathrm{loss}})$ on all four real-world datasets.
Each loss uses the ten data splits and three initialization seeds of Table~\ref{tab:real-world-both}, giving 120 paired comparisons.
Within each pair, initialization, data, preprocessing, architecture, optimizer, and training schedule are identical; both fits run on the same AMD EPYC 9654 host with one thread each.
All gates remain fixed at one, with no complexity penalty.
Both arms train for 5,000 epochs and select checkpoints by minimum validation MSE, evaluated every ten epochs.
SF uses the same procedure and dictionaries as the main results.

We restrict this loss control to unregularized KAN. For S2KAN, replacing log MSE with raw MSE makes the balance with the complexity penalty depend on the scale of the squared error, so retaining $\beta=1$ would not preserve the regularization strength; a raw-loss comparison would require separate calibration of $\beta$.

Log loss reduces median test RMSE from 2.047 to 2.021 dB on airfoil, 4.900 to 4.686 MPa on concrete, and 4.408 to 4.330 MW on power plant, while increasing it from 18.803 to 18.874 K on superconductors (Table~\ref{tab:kan-loss-control}).
Log loss changes KAN's SF in both directions; S2KAN's median SF remains higher in all 24 dataset--setting comparisons. The unchanged S2KAN models in Table~\ref{tab:real-world-both} also retain substantially smaller parameter counts.
Thus the SF and size advantages persist when both methods use a log-MSE data term; predictive accuracy remains dataset dependent.

\begin{table}%
\centering
\caption{Baseline KAN trained with raw or log MSE. Values are medians over thirty fits; smaller brackets give IQRs, omitted for constant values. All checkpoints are validation-selected. Parameter counts are unchanged by the loss choice.}
\label{tab:kan-loss-control}
\begingroup\small\setlength{\tabcolsep}{5pt}
\begin{tabular}{@{}lrrrrrrr@{}}
\toprule
 & & \multicolumn{3}{c}{SF (\%), $\epsilon=0.01$} & \multicolumn{3}{c}{SF (\%), $\epsilon=0.1$} \\
\cmidrule(lr){3-5}\cmidrule(lr){6-8}
KAN loss & RMSE & $J=1$ & $J=2$ & $J=3$ & $J=1$ & $J=2$ & $J=3$ \\
\midrule
\multicolumn{8}{@{}l}{\textbf{Airfoil}: RMSE in dB} \\
Raw MSE & \shortstack[r]{2.047\\[-1pt]{\fontsize{6}{7}\selectfont [1.963, 2.132]}} & \shortstack[r]{10.5\\[-1pt]{\fontsize{6}{7}\selectfont [7.2, 13.5]}} & \shortstack[r]{26.8\\[-1pt]{\fontsize{6}{7}\selectfont [24.4, 28.6]}} & \shortstack[r]{28.6\\[-1pt]{\fontsize{6}{7}\selectfont [28.6, 31.0]}} & \shortstack[r]{14.3\\[-1pt]{\fontsize{6}{7}\selectfont [11.1, 17.7]}} & \shortstack[r]{39.3\\[-1pt]{\fontsize{6}{7}\selectfont [36.0, 46.0]}} & \shortstack[r]{59.6\\[-1pt]{\fontsize{6}{7}\selectfont [51.8, 62.1]}} \\
Log MSE & \shortstack[r]{2.021\\[-1pt]{\fontsize{6}{7}\selectfont [1.972, 2.110]}} & \shortstack[r]{11.1\\[-1pt]{\fontsize{6}{7}\selectfont [10.3, 14.2]}} & \shortstack[r]{25.9\\[-1pt]{\fontsize{6}{7}\selectfont [25.0, 28.6]}} & \shortstack[r]{28.6\\[-1pt]{\fontsize{6}{7}\selectfont [28.6, 31.0]}} & \shortstack[r]{14.8\\[-1pt]{\fontsize{6}{7}\selectfont [13.8, 17.9]}} & \shortstack[r]{43.7\\[-1pt]{\fontsize{6}{7}\selectfont [37.6, 44.8]}} & \shortstack[r]{58.2\\[-1pt]{\fontsize{6}{7}\selectfont [54.0, 65.2]}} \\
\addlinespace[4pt]
\multicolumn{8}{@{}l}{\textbf{Concrete}: RMSE in MPa} \\
Raw MSE & \shortstack[r]{4.900\\[-1pt]{\fontsize{6}{7}\selectfont [4.488, 5.234]}} & \shortstack[r]{0.0\\[-1pt]{\fontsize{6}{7}\selectfont [0.0, 0.3]}} & \shortstack[r]{2.6\\[-1pt]{\fontsize{6}{7}\selectfont [1.9, 3.4]}} & \shortstack[r]{12.9\\[-1pt]{\fontsize{6}{7}\selectfont [11.6, 13.9]}} & \shortstack[r]{26.2\\[-1pt]{\fontsize{6}{7}\selectfont [19.3, 29.0]}} & \shortstack[r]{61.5\\[-1pt]{\fontsize{6}{7}\selectfont [47.2, 65.1]}} & \shortstack[r]{83.2\\[-1pt]{\fontsize{6}{7}\selectfont [71.5, 86.9]}} \\
Log MSE & \shortstack[r]{4.686\\[-1pt]{\fontsize{6}{7}\selectfont [4.495, 5.067]}} & \shortstack[r]{0.3\\[-1pt]{\fontsize{6}{7}\selectfont [0.0, 0.6]}} & \shortstack[r]{2.7\\[-1pt]{\fontsize{6}{7}\selectfont [1.9, 3.6]}} & \shortstack[r]{13.3\\[-1pt]{\fontsize{6}{7}\selectfont [12.3, 14.6]}} & \shortstack[r]{23.1\\[-1pt]{\fontsize{6}{7}\selectfont [19.5, 28.9]}} & \shortstack[r]{56.1\\[-1pt]{\fontsize{6}{7}\selectfont [50.0, 64.2]}} & \shortstack[r]{78.9\\[-1pt]{\fontsize{6}{7}\selectfont [73.7, 87.7]}} \\
\addlinespace[4pt]
\multicolumn{8}{@{}l}{\textbf{Power plant}: RMSE in MW} \\
Raw MSE & \shortstack[r]{4.408\\[-1pt]{\fontsize{6}{7}\selectfont [4.269, 4.496]}} & \shortstack[r]{0.0\\[-1pt]{\fontsize{6}{7}\selectfont \vphantom{[0.0, 0.0]}}} & \shortstack[r]{0.0\\[-1pt]{\fontsize{6}{7}\selectfont \vphantom{[0.0, 0.0]}}} & \shortstack[r]{0.0\\[-1pt]{\fontsize{6}{7}\selectfont [0.0, 7.2]}} & \shortstack[r]{12.8\\[-1pt]{\fontsize{6}{7}\selectfont [8.7, 20.0]}} & \shortstack[r]{51.0\\[-1pt]{\fontsize{6}{7}\selectfont [43.6, 56.4]}} & \shortstack[r]{74.5\\[-1pt]{\fontsize{6}{7}\selectfont [68.4, 80.3]}} \\
Log MSE & \shortstack[r]{4.330\\[-1pt]{\fontsize{6}{7}\selectfont [4.232, 4.514]}} & \shortstack[r]{0.0\\[-1pt]{\fontsize{6}{7}\selectfont \vphantom{[0.0, 0.0]}}} & \shortstack[r]{0.0\\[-1pt]{\fontsize{6}{7}\selectfont \vphantom{[0.0, 0.0]}}} & \shortstack[r]{4.1\\[-1pt]{\fontsize{6}{7}\selectfont [0.0, 4.5]}} & \shortstack[r]{13.3\\[-1pt]{\fontsize{6}{7}\selectfont [8.7, 23.4]}} & \shortstack[r]{54.4\\[-1pt]{\fontsize{6}{7}\selectfont [47.9, 59.6]}} & \shortstack[r]{82.2\\[-1pt]{\fontsize{6}{7}\selectfont [72.7, 86.8]}} \\
\addlinespace[4pt]
\multicolumn{8}{@{}l}{\textbf{Superconductor}: RMSE in K} \\
Raw MSE & \shortstack[r]{18.803\\[-1pt]{\fontsize{6}{7}\selectfont [18.222, 19.431]}} & \shortstack[r]{0.0\\[-1pt]{\fontsize{6}{7}\selectfont \vphantom{[0.0, 0.0]}}} & \shortstack[r]{0.0\\[-1pt]{\fontsize{6}{7}\selectfont [0.0, 3.7]}} & \shortstack[r]{16.7\\[-1pt]{\fontsize{6}{7}\selectfont [15.3, 16.7]}} & \shortstack[r]{3.3\\[-1pt]{\fontsize{6}{7}\selectfont [0.0, 3.6]}} & \shortstack[r]{19.3\\[-1pt]{\fontsize{6}{7}\selectfont [16.7, 23.3]}} & \shortstack[r]{33.9\\[-1pt]{\fontsize{6}{7}\selectfont [30.0, 36.7]}} \\
Log MSE & \shortstack[r]{18.874\\[-1pt]{\fontsize{6}{7}\selectfont [18.009, 19.381]}} & \shortstack[r]{0.0\\[-1pt]{\fontsize{6}{7}\selectfont \vphantom{[0.0, 0.0]}}} & \shortstack[r]{3.3\\[-1pt]{\fontsize{6}{7}\selectfont [0.0, 3.5]}} & \shortstack[r]{16.7\\[-1pt]{\fontsize{6}{7}\selectfont [15.3, 16.7]}} & \shortstack[r]{3.3\\[-1pt]{\fontsize{6}{7}\selectfont [0.0, 3.6]}} & \shortstack[r]{18.9\\[-1pt]{\fontsize{6}{7}\selectfont [14.5, 24.6]}} & \shortstack[r]{33.3\\[-1pt]{\fontsize{6}{7}\selectfont [26.7, 39.8]}} \\
\addlinespace[4pt]
\bottomrule
\end{tabular}
 \endgroup
\end{table}

\subsection{Activation-family ablations}
\label{sec:eql-like-control}

We test two restrictions of the activation dictionary. No-dense S2KAN disables both spline and SiLU contributions throughout training and inference. Dense-only S2KAN retains only this jointly gated branch, with an empty symbolic library and $P=Q=0$; its dense gates remain trainable.
The no-dense variant yields a model related to equation-learner architectures~\citep{martius2016extrapolation,pmlr-v80-sahoo18a}, although this is an internal ablation of S2KAN rather than a reproduction of EQL.

\subsubsection{Real-world datasets}
\label{sec:dense-only-control}

We compare both controls with the full S2KAN models in Table~\ref{tab:real-world-both}, using all four datasets, ten data splits and three initialization seeds per split (120 models per method). All three methods use the same architecture, log-MSE objective, $\beta=1$ after 500 warmup epochs, constant learning rate $10^{-3}$, batch size 64, and 5,000 training epochs. Checkpoints minimize validation MSE evaluated every ten epochs; fitting, validation and test samples and preprocessing are identical between methods. The controls use the same EPYC 9654 central processing unit (CPU) class as the archived full models but were trained separately. Table~\ref{tab:realworld-eql-control} includes all fits and uses the same SF procedure and evaluation dictionary for every method.

Full S2KAN retains dense terms in all 30 Airfoil and all 30 Superconductor models, and achieves lower median test RMSE than the no-dense control on both datasets (2.022 versus 2.087 and 19.514 versus 20.068, respectively). Its RMSE is lower in nine of ten splits for each dataset after averaging over the three initialization seeds. These gains accompany larger models: median $k$ is 103.8 versus 64.4 for Airfoil and 61.3 versus 41.6 for Superconductor. On Concrete and Power Plant, all full-S2KAN models close their dense gates; the control has lower median error on Concrete, with larger $k$, while Power Plant differs little in either error or size. Across the six correlated SF settings and four datasets, 18 median comparisons favor the control, four favor full S2KAN, and two tie. Across all 24 comparisons, including ties, the median SF difference (full S2KAN minus no-dense S2KAN) is $-1.9$ percentage points. This comparison shows a dataset-dependent tradeoff among predictive accuracy, representation size and symbolic fidelity.

Full S2KAN has lower median test RMSE than dense-only S2KAN on all four datasets. After averaging over the three seeds within each split, its RMSE is lower in all ten splits for Airfoil, Concrete and Power Plant, and six of ten for Superconductor. Its median $\ktot$ is also smaller on three datasets; on Concrete, dense-only is smaller (64.4 versus 76.2) but less accurate (RMSE 7.366 versus 5.268 MPa). Full S2KAN has higher median SF in all 24 dataset--setting comparisons. Thus sparsifying dense branches alone does not reproduce the accuracy and symbolic fidelity of the full dictionary under this protocol. One dense-only Concrete fit selects an epoch-10 checkpoint during warmup; it remains included, following the common validation rule. 

\begin{table}[t]%
\centering\singlespacing{}
\caption{Activation-family ablations on four real-world datasets. Entries are medians over 30 fits per dataset and method, with IQRs in brackets (omitted for constant values).}
\label{tab:realworld-eql-control}
\begingroup\small\setlength{\tabcolsep}{3pt}\renewcommand{\arraystretch}{1.13}
\textit{(a) Prediction and model size}\par\smallskip
\begin{tabular}{@{}llrrrr@{}}
\toprule
Dataset & Model & RMSE & $R^2$ & Active & $k$ \\
\midrule
Airfoil & Full S2KAN & \shortstack[r]{2.022\\[-1pt]{\fontsize{6}{7}\selectfont [1.954, 2.048]}} & \shortstack[r]{0.912\\[-1pt]{\fontsize{6}{7}\selectfont [0.909, 0.920]}} & \shortstack[r]{26\\[-1pt]{\fontsize{6}{7}\selectfont [24.2, 26.0]}} & \shortstack[r]{103.8\\[-1pt]{\fontsize{6}{7}\selectfont [94.7, 114.1]}} \\
 & No-dense S2KAN & \shortstack[r]{2.087\\[-1pt]{\fontsize{6}{7}\selectfont [2.021, 2.148]}} & \shortstack[r]{0.908\\[-1pt]{\fontsize{6}{7}\selectfont [0.902, 0.915]}} & \shortstack[r]{26\\[-1pt]{\fontsize{6}{7}\selectfont [25.0, 26.0]}} & \shortstack[r]{64.4\\[-1pt]{\fontsize{6}{7}\selectfont [62.1, 69.8]}} \\
 & Dense-only S2KAN & \shortstack[r]{2.334\\[-1pt]{\fontsize{6}{7}\selectfont [2.168, 2.414]}} & \shortstack[r]{0.886\\[-1pt]{\fontsize{6}{7}\selectfont [0.875, 0.898]}} & \shortstack[r]{14\\[-1pt]{\fontsize{6}{7}\selectfont [13.0, 16.0]}} & \shortstack[r]{211.9\\[-1pt]{\fontsize{6}{7}\selectfont [197.3, 240.0]}} \\
\addlinespace
Concrete & Full S2KAN & \shortstack[r]{5.268\\[-1pt]{\fontsize{6}{7}\selectfont [4.996, 5.612]}} & \shortstack[r]{0.894\\[-1pt]{\fontsize{6}{7}\selectfont [0.885, 0.906]}} & \shortstack[r]{61\\[-1pt]{\fontsize{6}{7}\selectfont [58.0, 63.8]}} & \shortstack[r]{76.2\\[-1pt]{\fontsize{6}{7}\selectfont [71.7, 80.4]}} \\
 & No-dense S2KAN & \shortstack[r]{5.165\\[-1pt]{\fontsize{6}{7}\selectfont [4.876, 5.404]}} & \shortstack[r]{0.898\\[-1pt]{\fontsize{6}{7}\selectfont [0.891, 0.909]}} & \shortstack[r]{71\\[-1pt]{\fontsize{6}{7}\selectfont [63.5, 75.8]}} & \shortstack[r]{88.1\\[-1pt]{\fontsize{6}{7}\selectfont [79.5, 94.6]}} \\
 & Dense-only S2KAN & \shortstack[r]{7.366\\[-1pt]{\fontsize{6}{7}\selectfont [7.016, 7.878]}} & \shortstack[r]{0.793\\[-1pt]{\fontsize{6}{7}\selectfont [0.781, 0.809]}} & \shortstack[r]{4\\[-1pt]{\fontsize{6}{7}\selectfont [4.0, 5.0]}} & \shortstack[r]{64.4\\[-1pt]{\fontsize{6}{7}\selectfont [60.9, 71.4]}} \\
\addlinespace
Power plant & Full S2KAN & \shortstack[r]{4.186\\[-1pt]{\fontsize{6}{7}\selectfont [4.104, 4.468]}} & \shortstack[r]{0.937\\[-1pt]{\fontsize{6}{7}\selectfont [0.931, 0.941]}} & \shortstack[r]{16\\[-1pt]{\fontsize{6}{7}\selectfont [15.0, 17.0]}} & \shortstack[r]{36.7\\[-1pt]{\fontsize{6}{7}\selectfont [34.3, 39.4]}} \\
 & No-dense S2KAN & \shortstack[r]{4.178\\[-1pt]{\fontsize{6}{7}\selectfont [4.114, 4.466]}} & \shortstack[r]{0.937\\[-1pt]{\fontsize{6}{7}\selectfont [0.931, 0.942]}} & \shortstack[r]{15\\[-1pt]{\fontsize{6}{7}\selectfont [15.0, 16.0]}} & \shortstack[r]{36.8\\[-1pt]{\fontsize{6}{7}\selectfont [33.8, 39.0]}} \\
 & Dense-only S2KAN & \shortstack[r]{4.622\\[-1pt]{\fontsize{6}{7}\selectfont [4.496, 4.844]}} & \shortstack[r]{0.924\\[-1pt]{\fontsize{6}{7}\selectfont [0.920, 0.930]}} & \shortstack[r]{7\\[-1pt]{\fontsize{6}{7}\selectfont [6.0, 8.0]}} & \shortstack[r]{107.2\\[-1pt]{\fontsize{6}{7}\selectfont [95.1, 122.6]}} \\
\addlinespace
Superconductor & Full S2KAN & \shortstack[r]{19.514\\[-1pt]{\fontsize{6}{7}\selectfont [18.853, 20.273]}} & \shortstack[r]{0.682\\[-1pt]{\fontsize{6}{7}\selectfont [0.637, 0.698]}} & \shortstack[r]{16\\[-1pt]{\fontsize{6}{7}\selectfont [14.2, 19.0]}} & \shortstack[r]{61.3\\[-1pt]{\fontsize{6}{7}\selectfont [51.8, 74.8]}} \\
 & No-dense S2KAN & \shortstack[r]{20.068\\[-1pt]{\fontsize{6}{7}\selectfont [19.541, 20.639]}} & \shortstack[r]{0.665\\[-1pt]{\fontsize{6}{7}\selectfont [0.642, 0.672]}} & \shortstack[r]{20.5\\[-1pt]{\fontsize{6}{7}\selectfont [18.0, 22.0]}} & \shortstack[r]{41.6\\[-1pt]{\fontsize{6}{7}\selectfont [33.2, 45.6]}} \\
 & Dense-only S2KAN & \shortstack[r]{19.952\\[-1pt]{\fontsize{6}{7}\selectfont [18.987, 20.521]}} & \shortstack[r]{0.658\\[-1pt]{\fontsize{6}{7}\selectfont [0.641, 0.687]}} & \shortstack[r]{6.5\\[-1pt]{\fontsize{6}{7}\selectfont [5.2, 8.0]}} & \shortstack[r]{100.8\\[-1pt]{\fontsize{6}{7}\selectfont [82.8, 130.8]}} \\
\addlinespace
\bottomrule
\end{tabular}
 \par\medskip
\textit{(b) Symbolic fidelity (\%)}\par\smallskip
\begin{tabular}{@{}llrrrrrr@{}}
\toprule
& & \multicolumn{3}{c}{$\epsilon=0.01$} & \multicolumn{3}{c}{$\epsilon=0.1$} \\
Dataset & Model & $J=1$ & $J=2$ & $J=3$ & $J=1$ & $J=2$ & $J=3$ \\
\midrule
Airfoil & Full S2KAN & \shortstack[r]{22.4\\[-1pt]{\fontsize{6}{7}\selectfont [20.0, 25.7]}} & \shortstack[r]{50.0\\[-1pt]{\fontsize{6}{7}\selectfont [46.3, 52.2]}} & \shortstack[r]{64.0\\[-1pt]{\fontsize{6}{7}\selectfont [60.0, 69.2]}} & \shortstack[r]{33.3\\[-1pt]{\fontsize{6}{7}\selectfont [29.4, 39.8]}} & \shortstack[r]{84.3\\[-1pt]{\fontsize{6}{7}\selectfont [82.6, 87.5]}} & \shortstack[r]{91.3\\[-1pt]{\fontsize{6}{7}\selectfont [87.5, 92.0]}} \\
 & No-dense S2KAN & \shortstack[r]{20.9\\[-1pt]{\fontsize{6}{7}\selectfont [17.6, 26.7]}} & \shortstack[r]{54.2\\[-1pt]{\fontsize{6}{7}\selectfont [48.9, 59.1]}} & \shortstack[r]{71.4\\[-1pt]{\fontsize{6}{7}\selectfont [65.2, 73.6]}} & \shortstack[r]{38.6\\[-1pt]{\fontsize{6}{7}\selectfont [34.8, 41.5]}} & \shortstack[r]{88.2\\[-1pt]{\fontsize{6}{7}\selectfont [85.1, 91.8]}} & \shortstack[r]{96.0\\[-1pt]{\fontsize{6}{7}\selectfont [95.7, 100.0]}} \\
 & Dense-only S2KAN & \shortstack[r]{6.5\\[-1pt]{\fontsize{6}{7}\selectfont [0.0, 9.6]}} & \shortstack[r]{25.0\\[-1pt]{\fontsize{6}{7}\selectfont [19.1, 27.1]}} & \shortstack[r]{25.8\\[-1pt]{\fontsize{6}{7}\selectfont [25.0, 30.8]}} & \shortstack[r]{8.0\\[-1pt]{\fontsize{6}{7}\selectfont [6.2, 13.8]}} & \shortstack[r]{33.3\\[-1pt]{\fontsize{6}{7}\selectfont [31.2, 41.7]}} & \shortstack[r]{50.0\\[-1pt]{\fontsize{6}{7}\selectfont [40.4, 55.8]}} \\
\addlinespace
Concrete & Full S2KAN & \shortstack[r]{77.8\\[-1pt]{\fontsize{6}{7}\selectfont [73.7, 79.5]}} & \shortstack[r]{85.4\\[-1pt]{\fontsize{6}{7}\selectfont [83.5, 88.6]}} & \shortstack[r]{86.9\\[-1pt]{\fontsize{6}{7}\selectfont [85.1, 90.4]}} & \shortstack[r]{82.1\\[-1pt]{\fontsize{6}{7}\selectfont [78.5, 85.6]}} & \shortstack[r]{90.3\\[-1pt]{\fontsize{6}{7}\selectfont [89.1, 94.2]}} & \shortstack[r]{94.4\\[-1pt]{\fontsize{6}{7}\selectfont [91.9, 96.1]}} \\
 & No-dense S2KAN & \shortstack[r]{76.4\\[-1pt]{\fontsize{6}{7}\selectfont [73.0, 81.7]}} & \shortstack[r]{86.5\\[-1pt]{\fontsize{6}{7}\selectfont [84.7, 91.5]}} & \shortstack[r]{88.0\\[-1pt]{\fontsize{6}{7}\selectfont [85.1, 92.3]}} & \shortstack[r]{82.2\\[-1pt]{\fontsize{6}{7}\selectfont [78.5, 85.4]}} & \shortstack[r]{92.0\\[-1pt]{\fontsize{6}{7}\selectfont [89.9, 95.0]}} & \shortstack[r]{94.3\\[-1pt]{\fontsize{6}{7}\selectfont [92.7, 95.9]}} \\
 & Dense-only S2KAN & \shortstack[r]{0.0\\[-1pt]{\fontsize{6}{7}\selectfont \vphantom{[0.0, 0.0]}}} & \shortstack[r]{0.0\\[-1pt]{\fontsize{6}{7}\selectfont \vphantom{[0.0, 0.0]}}} & \shortstack[r]{0.0\\[-1pt]{\fontsize{6}{7}\selectfont [0.0, 23.8]}} & \shortstack[r]{0.0\\[-1pt]{\fontsize{6}{7}\selectfont \vphantom{[0.0, 0.0]}}} & \shortstack[r]{25.0\\[-1pt]{\fontsize{6}{7}\selectfont [21.2, 50.0]}} & \shortstack[r]{75.0\\[-1pt]{\fontsize{6}{7}\selectfont [52.5, 78.8]}} \\
\addlinespace
Power plant & Full S2KAN & \shortstack[r]{35.5\\[-1pt]{\fontsize{6}{7}\selectfont [30.2, 39.6]}} & \shortstack[r]{64.3\\[-1pt]{\fontsize{6}{7}\selectfont [59.1, 71.4]}} & \shortstack[r]{69.9\\[-1pt]{\fontsize{6}{7}\selectfont [64.3, 79.6]}} & \shortstack[r]{50.0\\[-1pt]{\fontsize{6}{7}\selectfont [46.2, 57.1]}} & \shortstack[r]{92.6\\[-1pt]{\fontsize{6}{7}\selectfont [85.7, 93.9]}} & \shortstack[r]{92.9\\[-1pt]{\fontsize{6}{7}\selectfont [92.3, 100.0]}} \\
 & No-dense S2KAN & \shortstack[r]{27.6\\[-1pt]{\fontsize{6}{7}\selectfont [22.9, 40.9]}} & \shortstack[r]{65.5\\[-1pt]{\fontsize{6}{7}\selectfont [60.4, 74.1]}} & \shortstack[r]{73.2\\[-1pt]{\fontsize{6}{7}\selectfont [64.9, 80.9]}} & \shortstack[r]{50.0\\[-1pt]{\fontsize{6}{7}\selectfont [39.1, 56.5]}} & \shortstack[r]{92.6\\[-1pt]{\fontsize{6}{7}\selectfont [87.5, 98.6]}} & \shortstack[r]{93.8\\[-1pt]{\fontsize{6}{7}\selectfont [92.3, 100.0]}} \\
 & Dense-only S2KAN & \shortstack[r]{0.0\\[-1pt]{\fontsize{6}{7}\selectfont \vphantom{[0.0, 0.0]}}} & \shortstack[r]{0.0\\[-1pt]{\fontsize{6}{7}\selectfont \vphantom{[0.0, 0.0]}}} & \shortstack[r]{0.0\\[-1pt]{\fontsize{6}{7}\selectfont \vphantom{[0.0, 0.0]}}} & \shortstack[r]{0.0\\[-1pt]{\fontsize{6}{7}\selectfont \vphantom{[0.0, 0.0]}}} & \shortstack[r]{29.2\\[-1pt]{\fontsize{6}{7}\selectfont [16.7, 42.1]}} & \shortstack[r]{63.3\\[-1pt]{\fontsize{6}{7}\selectfont [50.0, 78.8]}} \\
\addlinespace
Superconductor & Full S2KAN & \shortstack[r]{19.4\\[-1pt]{\fontsize{6}{7}\selectfont [13.4, 27.5]}} & \shortstack[r]{39.4\\[-1pt]{\fontsize{6}{7}\selectfont [30.1, 49.3]}} & \shortstack[r]{53.3\\[-1pt]{\fontsize{6}{7}\selectfont [44.2, 60.2]}} & \shortstack[r]{25.0\\[-1pt]{\fontsize{6}{7}\selectfont [15.7, 33.3]}} & \shortstack[r]{55.6\\[-1pt]{\fontsize{6}{7}\selectfont [50.0, 65.6]}} & \shortstack[r]{65.9\\[-1pt]{\fontsize{6}{7}\selectfont [57.1, 73.6]}} \\
 & No-dense S2KAN & \shortstack[r]{21.5\\[-1pt]{\fontsize{6}{7}\selectfont [15.8, 29.4]}} & \shortstack[r]{46.4\\[-1pt]{\fontsize{6}{7}\selectfont [35.7, 57.7]}} & \shortstack[r]{58.6\\[-1pt]{\fontsize{6}{7}\selectfont [47.4, 66.7]}} & \shortstack[r]{27.6\\[-1pt]{\fontsize{6}{7}\selectfont [19.0, 33.3]}} & \shortstack[r]{63.2\\[-1pt]{\fontsize{6}{7}\selectfont [55.6, 68.0]}} & \shortstack[r]{71.4\\[-1pt]{\fontsize{6}{7}\selectfont [63.5, 77.4]}} \\
 & Dense-only S2KAN & \shortstack[r]{0.0\\[-1pt]{\fontsize{6}{7}\selectfont \vphantom{[0.0, 0.0]}}} & \shortstack[r]{0.0\\[-1pt]{\fontsize{6}{7}\selectfont \vphantom{[0.0, 0.0]}}} & \shortstack[r]{0.0\\[-1pt]{\fontsize{6}{7}\selectfont \vphantom{[0.0, 0.0]}}} & \shortstack[r]{0.0\\[-1pt]{\fontsize{6}{7}\selectfont \vphantom{[0.0, 0.0]}}} & \shortstack[r]{0.0\\[-1pt]{\fontsize{6}{7}\selectfont [0.0, 13.3]}} & \shortstack[r]{16.7\\[-1pt]{\fontsize{6}{7}\selectfont [12.9, 20.0]}} \\
\addlinespace
\bottomrule
\end{tabular}
 \endgroup
\end{table}

\subsubsection{Nguyen benchmarks}

We compare full and no-dense S2KAN, retaining the symbolic, Chebyshev and Fourier terms in both models. Both methods use the L architecture $[n_0,3,1]$, $\beta=1$, the same dictionary, training/test sample and ten initialization seeds, a log-MSE data term, constant learning rate $10^{-3}$, batch size 128, and 200 warmup epochs. Their grid-update and stopping rules match; all 200 fits reached 10,000 epochs. Table~\ref{tab:nguyen-eql-control} retains every fit and uses the same symbolic-fidelity procedure as Table~\ref{tab:nguyen-full}.

The no-dense control achieves strong predictive performance on the Nguyen benchmarks, demonstrating the approximation capacity of the symbolic and finite-basis dictionary alone. Full S2KAN often selects a representation with no dense contribution: 71 of its 100 fitted models close all dense gates at the final checkpoint, while keeping these components available during training. Full S2KAN also achieves lower median test RMSE on six of ten problems, with the control achieving lower error on the remaining four. Across the ten problems and six correlated fidelity settings, 39 median comparisons tie, 11 favor full S2KAN and 10 favor the control; these descriptive counts indicate no consistent fidelity advantage for either configuration. Across all 60 comparisons, including ties, the median SF difference (full S2KAN minus no-dense S2KAN) is zero percentage points. Together with the library-misspecification experiment, these results support the flexibility of joint symbolic and dense selection: the model can arrive at a restricted representation when the dictionary suffices while retaining access to a dense fallback\@.

\begin{table}[!t]%
\centering\singlespacing{}
\caption{Dense-branch ablation on Nguyen benchmarks, using architecture L and $\beta=1$ for both methods. Entries are medians over ten initialization seeds on the same training/test sample; brackets give the first and third quartiles (omitted for constant values). ``No-dense S2KAN'' disables both spline and SiLU contributions while retaining the symbolic, Chebyshev and Fourier terms. RMSE is test error; Active counts gate-active edges; $k$ is weighted expected support. SF is the percentage of functionally active edges approximated within relative RMSE tolerance $\epsilon$ by at most $J$ gauged terms from $\{x,x^2,\sin x\}$. All fits are included; none has an empty functional-activity set.}
\label{tab:nguyen-eql-control}
\begingroup\small\setlength{\tabcolsep}{4pt}\renewcommand{\arraystretch}{1.13}
\begin{tabular}{@{}llrrrrrrrrr@{}}
\toprule
 & & & \multicolumn{2}{c}{Model size} & \multicolumn{3}{c}{SF (\%), $\epsilon=0.01$} & \multicolumn{3}{c}{SF (\%), $\epsilon=0.1$} \\
\cmidrule(lr){4-5}\cmidrule(lr){6-8}\cmidrule(lr){9-11}
Problem & Model & RMSE & Active & $k$ & $J=1$ & $J=2$ & $J=3$ & $J=1$ & $J=2$ & $J=3$ \\
\midrule
F1 & S2KAN & \shortstack[r]{0.0055\\[-1pt]{\fontsize{6}{7}\selectfont [0.0033, 0.0082]}} & \shortstack[r]{6\\[-1pt]{\fontsize{6}{7}\selectfont [6, 6]}} & \shortstack[r]{42.5\\[-1pt]{\fontsize{6}{7}\selectfont [38.4, 51.3]}} & \shortstack[r]{40.0\\[-1pt]{\fontsize{6}{7}\selectfont [21.3, 40.0]}} & \shortstack[r]{80.0\\[-1pt]{\fontsize{6}{7}\selectfont [76.2, 95.0]}} & \shortstack[r]{100.0\\[-1pt]{\fontsize{6}{7}\selectfont \vphantom{[0.0, 0.0]}}} & \shortstack[r]{60.0\\[-1pt]{\fontsize{6}{7}\selectfont [50.0, 60.0]}} & \shortstack[r]{100.0\\[-1pt]{\fontsize{6}{7}\selectfont [100.0, 100.0]}} & \shortstack[r]{100.0\\[-1pt]{\fontsize{6}{7}\selectfont \vphantom{[0.0, 0.0]}}} \\
 & No-dense S2KAN & \shortstack[r]{0.0041\\[-1pt]{\fontsize{6}{7}\selectfont [0.0030, 0.0083]}} & \shortstack[r]{6\\[-1pt]{\fontsize{6}{7}\selectfont \vphantom{[0.0, 0.0]}}} & \shortstack[r]{40.3\\[-1pt]{\fontsize{6}{7}\selectfont [39.0, 42.2]}} & \shortstack[r]{40.0\\[-1pt]{\fontsize{6}{7}\selectfont [40.0, 47.5]}} & \shortstack[r]{100.0\\[-1pt]{\fontsize{6}{7}\selectfont [100.0, 100.0]}} & \shortstack[r]{100.0\\[-1pt]{\fontsize{6}{7}\selectfont \vphantom{[0.0, 0.0]}}} & \shortstack[r]{60.0\\[-1pt]{\fontsize{6}{7}\selectfont [60.0, 60.0]}} & \shortstack[r]{100.0\\[-1pt]{\fontsize{6}{7}\selectfont \vphantom{[0.0, 0.0]}}} & \shortstack[r]{100.0\\[-1pt]{\fontsize{6}{7}\selectfont \vphantom{[0.0, 0.0]}}} \\
\addlinespace[3pt]
F2 & S2KAN & \shortstack[r]{0.0063\\[-1pt]{\fontsize{6}{7}\selectfont [0.0045, 0.0100]}} & \shortstack[r]{6\\[-1pt]{\fontsize{6}{7}\selectfont [6, 6]}} & \shortstack[r]{41.4\\[-1pt]{\fontsize{6}{7}\selectfont [40.0, 45.4]}} & \shortstack[r]{60.0\\[-1pt]{\fontsize{6}{7}\selectfont [52.5, 72.9]}} & \shortstack[r]{100.0\\[-1pt]{\fontsize{6}{7}\selectfont [80.0, 100.0]}} & \shortstack[r]{100.0\\[-1pt]{\fontsize{6}{7}\selectfont [100.0, 100.0]}} & \shortstack[r]{100.0\\[-1pt]{\fontsize{6}{7}\selectfont [76.2, 100.0]}} & \shortstack[r]{100.0\\[-1pt]{\fontsize{6}{7}\selectfont [100.0, 100.0]}} & \shortstack[r]{100.0\\[-1pt]{\fontsize{6}{7}\selectfont \vphantom{[0.0, 0.0]}}} \\
 & No-dense S2KAN & \shortstack[r]{0.0065\\[-1pt]{\fontsize{6}{7}\selectfont [0.0035, 0.0076]}} & \shortstack[r]{6\\[-1pt]{\fontsize{6}{7}\selectfont \vphantom{[0.0, 0.0]}}} & \shortstack[r]{41.3\\[-1pt]{\fontsize{6}{7}\selectfont [38.6, 45.3]}} & \shortstack[r]{55.0\\[-1pt]{\fontsize{6}{7}\selectfont [40.0, 60.0]}} & \shortstack[r]{100.0\\[-1pt]{\fontsize{6}{7}\selectfont \vphantom{[0.0, 0.0]}}} & \shortstack[r]{100.0\\[-1pt]{\fontsize{6}{7}\selectfont \vphantom{[0.0, 0.0]}}} & \shortstack[r]{90.0\\[-1pt]{\fontsize{6}{7}\selectfont [80.0, 100.0]}} & \shortstack[r]{100.0\\[-1pt]{\fontsize{6}{7}\selectfont \vphantom{[0.0, 0.0]}}} & \shortstack[r]{100.0\\[-1pt]{\fontsize{6}{7}\selectfont \vphantom{[0.0, 0.0]}}} \\
\addlinespace[3pt]
F3 & S2KAN & \shortstack[r]{0.0062\\[-1pt]{\fontsize{6}{7}\selectfont [0.0043, 0.0120]}} & \shortstack[r]{6\\[-1pt]{\fontsize{6}{7}\selectfont [6, 6]}} & \shortstack[r]{55.9\\[-1pt]{\fontsize{6}{7}\selectfont [53.2, 59.5]}} & \shortstack[r]{40.0\\[-1pt]{\fontsize{6}{7}\selectfont [28.7, 40.0]}} & \shortstack[r]{80.0\\[-1pt]{\fontsize{6}{7}\selectfont [75.0, 80.0]}} & \shortstack[r]{80.0\\[-1pt]{\fontsize{6}{7}\selectfont [80.0, 95.8]}} & \shortstack[r]{60.0\\[-1pt]{\fontsize{6}{7}\selectfont [42.5, 60.0]}} & \shortstack[r]{100.0\\[-1pt]{\fontsize{6}{7}\selectfont [85.0, 100.0]}} & \shortstack[r]{100.0\\[-1pt]{\fontsize{6}{7}\selectfont [100.0, 100.0]}} \\
 & No-dense S2KAN & \shortstack[r]{0.0089\\[-1pt]{\fontsize{6}{7}\selectfont [0.0041, 0.0098]}} & \shortstack[r]{6\\[-1pt]{\fontsize{6}{7}\selectfont \vphantom{[0.0, 0.0]}}} & \shortstack[r]{40.1\\[-1pt]{\fontsize{6}{7}\selectfont [39.2, 42.0]}} & \shortstack[r]{45.0\\[-1pt]{\fontsize{6}{7}\selectfont [28.7, 50.0]}} & \shortstack[r]{87.5\\[-1pt]{\fontsize{6}{7}\selectfont [75.0, 100.0]}} & \shortstack[r]{100.0\\[-1pt]{\fontsize{6}{7}\selectfont \vphantom{[0.0, 0.0]}}} & \shortstack[r]{60.0\\[-1pt]{\fontsize{6}{7}\selectfont [50.0, 60.0]}} & \shortstack[r]{100.0\\[-1pt]{\fontsize{6}{7}\selectfont \vphantom{[0.0, 0.0]}}} & \shortstack[r]{100.0\\[-1pt]{\fontsize{6}{7}\selectfont \vphantom{[0.0, 0.0]}}} \\
\addlinespace[3pt]
F4 & S2KAN & \shortstack[r]{0.0078\\[-1pt]{\fontsize{6}{7}\selectfont [0.0043, 0.0091]}} & \shortstack[r]{6\\[-1pt]{\fontsize{6}{7}\selectfont [5.25, 6]}} & \shortstack[r]{50.5\\[-1pt]{\fontsize{6}{7}\selectfont [46.6, 59.0]}} & \shortstack[r]{50.0\\[-1pt]{\fontsize{6}{7}\selectfont [50.0, 60.0]}} & \shortstack[r]{80.0\\[-1pt]{\fontsize{6}{7}\selectfont [76.2, 80.0]}} & \shortstack[r]{91.7\\[-1pt]{\fontsize{6}{7}\selectfont [80.0, 100.0]}} & \shortstack[r]{77.5\\[-1pt]{\fontsize{6}{7}\selectfont [75.0, 80.0]}} & \shortstack[r]{100.0\\[-1pt]{\fontsize{6}{7}\selectfont \vphantom{[0.0, 0.0]}}} & \shortstack[r]{100.0\\[-1pt]{\fontsize{6}{7}\selectfont \vphantom{[0.0, 0.0]}}} \\
 & No-dense S2KAN & \shortstack[r]{0.0050\\[-1pt]{\fontsize{6}{7}\selectfont [0.0039, 0.0074]}} & \shortstack[r]{6\\[-1pt]{\fontsize{6}{7}\selectfont \vphantom{[0.0, 0.0]}}} & \shortstack[r]{40.4\\[-1pt]{\fontsize{6}{7}\selectfont [37.8, 43.2]}} & \shortstack[r]{40.0\\[-1pt]{\fontsize{6}{7}\selectfont [40.0, 55.0]}} & \shortstack[r]{100.0\\[-1pt]{\fontsize{6}{7}\selectfont [100.0, 100.0]}} & \shortstack[r]{100.0\\[-1pt]{\fontsize{6}{7}\selectfont \vphantom{[0.0, 0.0]}}} & \shortstack[r]{60.0\\[-1pt]{\fontsize{6}{7}\selectfont [60.0, 80.0]}} & \shortstack[r]{100.0\\[-1pt]{\fontsize{6}{7}\selectfont \vphantom{[0.0, 0.0]}}} & \shortstack[r]{100.0\\[-1pt]{\fontsize{6}{7}\selectfont \vphantom{[0.0, 0.0]}}} \\
\addlinespace[3pt]
F5 & S2KAN & \shortstack[r]{0.0352\\[-1pt]{\fontsize{6}{7}\selectfont [0.0199, 0.0602]}} & \shortstack[r]{6\\[-1pt]{\fontsize{6}{7}\selectfont \vphantom{[0.0, 0.0]}}} & \shortstack[r]{58.2\\[-1pt]{\fontsize{6}{7}\selectfont [57.5, 58.7]}} & \shortstack[r]{75.0\\[-1pt]{\fontsize{6}{7}\selectfont [52.5, 75.0]}} & \shortstack[r]{100.0\\[-1pt]{\fontsize{6}{7}\selectfont \vphantom{[0.0, 0.0]}}} & \shortstack[r]{100.0\\[-1pt]{\fontsize{6}{7}\selectfont \vphantom{[0.0, 0.0]}}} & \shortstack[r]{100.0\\[-1pt]{\fontsize{6}{7}\selectfont [75.0, 100.0]}} & \shortstack[r]{100.0\\[-1pt]{\fontsize{6}{7}\selectfont \vphantom{[0.0, 0.0]}}} & \shortstack[r]{100.0\\[-1pt]{\fontsize{6}{7}\selectfont \vphantom{[0.0, 0.0]}}} \\
 & No-dense S2KAN & \shortstack[r]{0.0627\\[-1pt]{\fontsize{6}{7}\selectfont [0.0322, 0.0944]}} & \shortstack[r]{6\\[-1pt]{\fontsize{6}{7}\selectfont \vphantom{[0.0, 0.0]}}} & \shortstack[r]{57.4\\[-1pt]{\fontsize{6}{7}\selectfont [55.5, 58.4]}} & \shortstack[r]{75.0\\[-1pt]{\fontsize{6}{7}\selectfont [52.5, 75.0]}} & \shortstack[r]{100.0\\[-1pt]{\fontsize{6}{7}\selectfont \vphantom{[0.0, 0.0]}}} & \shortstack[r]{100.0\\[-1pt]{\fontsize{6}{7}\selectfont \vphantom{[0.0, 0.0]}}} & \shortstack[r]{87.5\\[-1pt]{\fontsize{6}{7}\selectfont [75.0, 100.0]}} & \shortstack[r]{100.0\\[-1pt]{\fontsize{6}{7}\selectfont \vphantom{[0.0, 0.0]}}} & \shortstack[r]{100.0\\[-1pt]{\fontsize{6}{7}\selectfont \vphantom{[0.0, 0.0]}}} \\
\addlinespace[3pt]
F6 & S2KAN & \shortstack[r]{0.0030\\[-1pt]{\fontsize{6}{7}\selectfont [0.0021, 0.0041]}} & \shortstack[r]{6\\[-1pt]{\fontsize{6}{7}\selectfont \vphantom{[0.0, 0.0]}}} & \shortstack[r]{40.9\\[-1pt]{\fontsize{6}{7}\selectfont [37.1, 44.2]}} & \shortstack[r]{80.0\\[-1pt]{\fontsize{6}{7}\selectfont [65.0, 95.0]}} & \shortstack[r]{100.0\\[-1pt]{\fontsize{6}{7}\selectfont [100.0, 100.0]}} & \shortstack[r]{100.0\\[-1pt]{\fontsize{6}{7}\selectfont \vphantom{[0.0, 0.0]}}} & \shortstack[r]{100.0\\[-1pt]{\fontsize{6}{7}\selectfont [85.0, 100.0]}} & \shortstack[r]{100.0\\[-1pt]{\fontsize{6}{7}\selectfont \vphantom{[0.0, 0.0]}}} & \shortstack[r]{100.0\\[-1pt]{\fontsize{6}{7}\selectfont \vphantom{[0.0, 0.0]}}} \\
 & No-dense S2KAN & \shortstack[r]{0.0048\\[-1pt]{\fontsize{6}{7}\selectfont [0.0029, 0.0070]}} & \shortstack[r]{6\\[-1pt]{\fontsize{6}{7}\selectfont \vphantom{[0.0, 0.0]}}} & \shortstack[r]{44.8\\[-1pt]{\fontsize{6}{7}\selectfont [44.3, 45.7]}} & \shortstack[r]{63.3\\[-1pt]{\fontsize{6}{7}\selectfont [45.0, 82.5]}} & \shortstack[r]{100.0\\[-1pt]{\fontsize{6}{7}\selectfont \vphantom{[0.0, 0.0]}}} & \shortstack[r]{100.0\\[-1pt]{\fontsize{6}{7}\selectfont \vphantom{[0.0, 0.0]}}} & \shortstack[r]{100.0\\[-1pt]{\fontsize{6}{7}\selectfont [80.8, 100.0]}} & \shortstack[r]{100.0\\[-1pt]{\fontsize{6}{7}\selectfont \vphantom{[0.0, 0.0]}}} & \shortstack[r]{100.0\\[-1pt]{\fontsize{6}{7}\selectfont \vphantom{[0.0, 0.0]}}} \\
\addlinespace[3pt]
F7 & S2KAN & \shortstack[r]{0.0196\\[-1pt]{\fontsize{6}{7}\selectfont [0.0088, 0.0287]}} & \shortstack[r]{6\\[-1pt]{\fontsize{6}{7}\selectfont \vphantom{[0.0, 0.0]}}} & \shortstack[r]{56.5\\[-1pt]{\fontsize{6}{7}\selectfont [55.5, 57.4]}} & \shortstack[r]{55.0\\[-1pt]{\fontsize{6}{7}\selectfont [50.0, 60.0]}} & \shortstack[r]{100.0\\[-1pt]{\fontsize{6}{7}\selectfont [100.0, 100.0]}} & \shortstack[r]{100.0\\[-1pt]{\fontsize{6}{7}\selectfont \vphantom{[0.0, 0.0]}}} & \shortstack[r]{80.0\\[-1pt]{\fontsize{6}{7}\selectfont [63.7, 100.0]}} & \shortstack[r]{100.0\\[-1pt]{\fontsize{6}{7}\selectfont \vphantom{[0.0, 0.0]}}} & \shortstack[r]{100.0\\[-1pt]{\fontsize{6}{7}\selectfont \vphantom{[0.0, 0.0]}}} \\
 & No-dense S2KAN & \shortstack[r]{0.0176\\[-1pt]{\fontsize{6}{7}\selectfont [0.0153, 0.0225]}} & \shortstack[r]{6\\[-1pt]{\fontsize{6}{7}\selectfont \vphantom{[0.0, 0.0]}}} & \shortstack[r]{59.5\\[-1pt]{\fontsize{6}{7}\selectfont [57.0, 60.2]}} & \shortstack[r]{60.0\\[-1pt]{\fontsize{6}{7}\selectfont [52.5, 71.2]}} & \shortstack[r]{100.0\\[-1pt]{\fontsize{6}{7}\selectfont \vphantom{[0.0, 0.0]}}} & \shortstack[r]{100.0\\[-1pt]{\fontsize{6}{7}\selectfont \vphantom{[0.0, 0.0]}}} & \shortstack[r]{100.0\\[-1pt]{\fontsize{6}{7}\selectfont [100.0, 100.0]}} & \shortstack[r]{100.0\\[-1pt]{\fontsize{6}{7}\selectfont \vphantom{[0.0, 0.0]}}} & \shortstack[r]{100.0\\[-1pt]{\fontsize{6}{7}\selectfont \vphantom{[0.0, 0.0]}}} \\
\addlinespace[3pt]
F8 & S2KAN & \shortstack[r]{0.0141\\[-1pt]{\fontsize{6}{7}\selectfont [0.0074, 0.0203]}} & \shortstack[r]{6\\[-1pt]{\fontsize{6}{7}\selectfont \vphantom{[0.0, 0.0]}}} & \shortstack[r]{51.8\\[-1pt]{\fontsize{6}{7}\selectfont [48.6, 54.2]}} & \shortstack[r]{41.7\\[-1pt]{\fontsize{6}{7}\selectfont [20.0, 57.5]}} & \shortstack[r]{77.5\\[-1pt]{\fontsize{6}{7}\selectfont [63.7, 80.0]}} & \shortstack[r]{100.0\\[-1pt]{\fontsize{6}{7}\selectfont [100.0, 100.0]}} & \shortstack[r]{63.3\\[-1pt]{\fontsize{6}{7}\selectfont [52.5, 95.0]}} & \shortstack[r]{100.0\\[-1pt]{\fontsize{6}{7}\selectfont \vphantom{[0.0, 0.0]}}} & \shortstack[r]{100.0\\[-1pt]{\fontsize{6}{7}\selectfont \vphantom{[0.0, 0.0]}}} \\
 & No-dense S2KAN & \shortstack[r]{0.0079\\[-1pt]{\fontsize{6}{7}\selectfont [0.0070, 0.0082]}} & \shortstack[r]{6\\[-1pt]{\fontsize{6}{7}\selectfont \vphantom{[0.0, 0.0]}}} & \shortstack[r]{56.8\\[-1pt]{\fontsize{6}{7}\selectfont [54.5, 58.1]}} & \shortstack[r]{10.0\\[-1pt]{\fontsize{6}{7}\selectfont [0.0, 25.0]}} & \shortstack[r]{75.0\\[-1pt]{\fontsize{6}{7}\selectfont [75.0, 100.0]}} & \shortstack[r]{100.0\\[-1pt]{\fontsize{6}{7}\selectfont [100.0, 100.0]}} & \shortstack[r]{50.0\\[-1pt]{\fontsize{6}{7}\selectfont [31.2, 57.5]}} & \shortstack[r]{100.0\\[-1pt]{\fontsize{6}{7}\selectfont [81.2, 100.0]}} & \shortstack[r]{100.0\\[-1pt]{\fontsize{6}{7}\selectfont \vphantom{[0.0, 0.0]}}} \\
\addlinespace[3pt]
F9 & S2KAN & \shortstack[r]{0.0037\\[-1pt]{\fontsize{6}{7}\selectfont [0.0029, 0.0044]}} & \shortstack[r]{9\\[-1pt]{\fontsize{6}{7}\selectfont \vphantom{[0.0, 0.0]}}} & \shortstack[r]{41.9\\[-1pt]{\fontsize{6}{7}\selectfont [39.8, 43.7]}} & \shortstack[r]{93.8\\[-1pt]{\fontsize{6}{7}\selectfont [70.8, 100.0]}} & \shortstack[r]{100.0\\[-1pt]{\fontsize{6}{7}\selectfont \vphantom{[0.0, 0.0]}}} & \shortstack[r]{100.0\\[-1pt]{\fontsize{6}{7}\selectfont \vphantom{[0.0, 0.0]}}} & \shortstack[r]{100.0\\[-1pt]{\fontsize{6}{7}\selectfont [87.5, 100.0]}} & \shortstack[r]{100.0\\[-1pt]{\fontsize{6}{7}\selectfont \vphantom{[0.0, 0.0]}}} & \shortstack[r]{100.0\\[-1pt]{\fontsize{6}{7}\selectfont \vphantom{[0.0, 0.0]}}} \\
 & No-dense S2KAN & \shortstack[r]{0.0037\\[-1pt]{\fontsize{6}{7}\selectfont [0.0020, 0.0046]}} & \shortstack[r]{9\\[-1pt]{\fontsize{6}{7}\selectfont [8.25, 9]}} & \shortstack[r]{38.2\\[-1pt]{\fontsize{6}{7}\selectfont [32.9, 42.4]}} & \shortstack[r]{85.7\\[-1pt]{\fontsize{6}{7}\selectfont [80.8, 96.4]}} & \shortstack[r]{100.0\\[-1pt]{\fontsize{6}{7}\selectfont \vphantom{[0.0, 0.0]}}} & \shortstack[r]{100.0\\[-1pt]{\fontsize{6}{7}\selectfont \vphantom{[0.0, 0.0]}}} & \shortstack[r]{85.7\\[-1pt]{\fontsize{6}{7}\selectfont [83.9, 100.0]}} & \shortstack[r]{100.0\\[-1pt]{\fontsize{6}{7}\selectfont \vphantom{[0.0, 0.0]}}} & \shortstack[r]{100.0\\[-1pt]{\fontsize{6}{7}\selectfont \vphantom{[0.0, 0.0]}}} \\
\addlinespace[3pt]
F10 & S2KAN & \shortstack[r]{0.0037\\[-1pt]{\fontsize{6}{7}\selectfont [0.0031, 0.0070]}} & \shortstack[r]{9\\[-1pt]{\fontsize{6}{7}\selectfont [9, 9]}} & \shortstack[r]{22.0\\[-1pt]{\fontsize{6}{7}\selectfont [20.1, 24.1]}} & \shortstack[r]{86.6\\[-1pt]{\fontsize{6}{7}\selectfont [55.6, 96.9]}} & \shortstack[r]{100.0\\[-1pt]{\fontsize{6}{7}\selectfont [100.0, 100.0]}} & \shortstack[r]{100.0\\[-1pt]{\fontsize{6}{7}\selectfont \vphantom{[0.0, 0.0]}}} & \shortstack[r]{94.4\\[-1pt]{\fontsize{6}{7}\selectfont [87.5, 100.0]}} & \shortstack[r]{100.0\\[-1pt]{\fontsize{6}{7}\selectfont \vphantom{[0.0, 0.0]}}} & \shortstack[r]{100.0\\[-1pt]{\fontsize{6}{7}\selectfont \vphantom{[0.0, 0.0]}}} \\
 & No-dense S2KAN & \shortstack[r]{0.0041\\[-1pt]{\fontsize{6}{7}\selectfont [0.0028, 0.0074]}} & \shortstack[r]{9\\[-1pt]{\fontsize{6}{7}\selectfont [8.25, 9]}} & \shortstack[r]{22.6\\[-1pt]{\fontsize{6}{7}\selectfont [17.8, 25.2]}} & \shortstack[r]{93.8\\[-1pt]{\fontsize{6}{7}\selectfont [85.7, 100.0]}} & \shortstack[r]{100.0\\[-1pt]{\fontsize{6}{7}\selectfont [100.0, 100.0]}} & \shortstack[r]{100.0\\[-1pt]{\fontsize{6}{7}\selectfont \vphantom{[0.0, 0.0]}}} & \shortstack[r]{100.0\\[-1pt]{\fontsize{6}{7}\selectfont [91.7, 100.0]}} & \shortstack[r]{100.0\\[-1pt]{\fontsize{6}{7}\selectfont \vphantom{[0.0, 0.0]}}} & \shortstack[r]{100.0\\[-1pt]{\fontsize{6}{7}\selectfont \vphantom{[0.0, 0.0]}}} \\
\bottomrule
\end{tabular}
 \endgroup
\end{table}

\subsection{Regularization-strength sensitivity}
\label{sec:beta-sensitivity}

\subsubsection{Real-world datasets}

\begin{figure}[t]
\centering
\includegraphics[width=\textwidth]{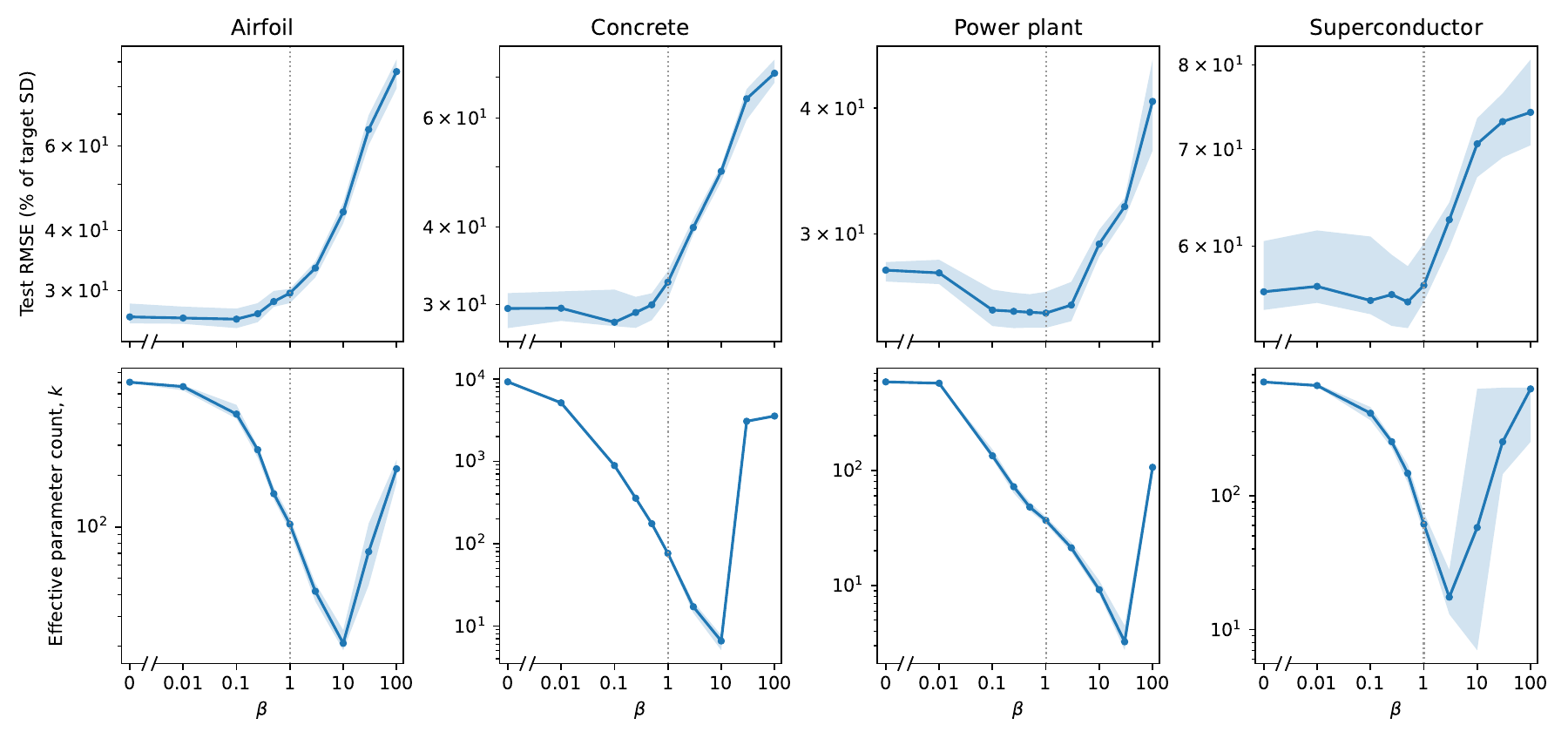}
\caption{Regularization trades model size against predictive accuracy on real-world datasets. Curves and shading show medians and IQRs over 30 validation-selected fits at each $\beta$. RMSE is expressed as a percentage of the test-target standard deviation (SD), calculated separately for each fit before aggregation. The dotted line marks $\beta=1$. 
}
\label{fig:rw-beta-selected}
\end{figure}

With the remaining real-world training settings held fixed, we vary $\beta$ over $\{0,\allowbreak 0.01,\allowbreak 0.1,\allowbreak 0.25,\allowbreak 0.5,\allowbreak 1,\allowbreak 3,\allowbreak 10,\allowbreak 30,\allowbreak 100\}$ on all ten data splits and three initialization seeds per split (1,200 fits). Each split--seed block uses identical initialization and warmup on the same CPU host. Figure~\ref{fig:rw-beta-selected} uses the main validation-selection rule. Relative to $\beta=0$, $\beta=1$ reduces median $\ktot$ by 85--99\%; median test RMSE improves for Power Plant, changes little for Superconductor, and increases for Airfoil and Concrete. Stronger regularization eventually degrades accuracy on every dataset. This sensitivity analysis does not select a new $\beta$ using test performance.

\begin{figure}[t]
\centering
\includegraphics[width=\textwidth]{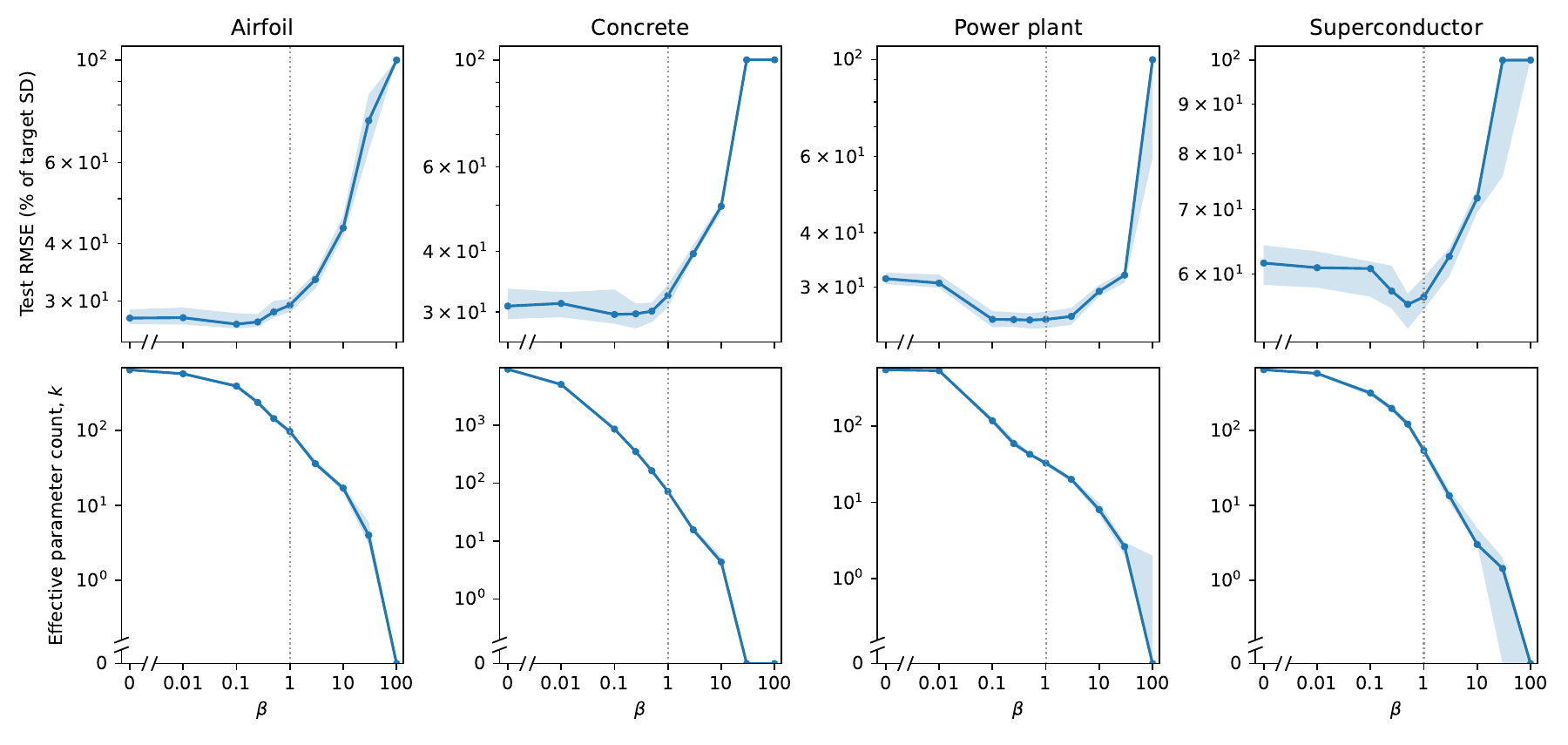}
\caption{The same real-world regularization sweep shown in Fig.~\ref{fig:rw-beta-selected} but for fits evaluated at the end of training. 
Unlike the validation-selected fits, median size decreases throughout the sweep; excessive regularization produces small but inaccurate models.}
\label{fig:rw-beta-final}
\end{figure}

The increase in selected model size at large $\beta$ reflects selection of earlier checkpoints, before further sparsification worsens validation error. Selection occurs within the 500-epoch unregularized warmup for 5/30 Airfoil fits at $\beta=100$ and 9/30, 14/30 and 16/30 Superconductor fits at $\beta=10,30,100$, respectively; no other settings select warmup checkpoints. For Concrete at $\beta=30$ and 100, every selected checkpoint is between epochs 580 and 610. At the common 5,000-epoch endpoint (Fig.~\ref{fig:rw-beta-final}), median $\ktot$ instead decreases monotonically with $\beta$, approaching zero at $\beta=100$ for every dataset, with poor predictive accuracy.

\subsubsection{Nguyen benchmarks}
\label{sec:nguyen-beta-sensitivity}

\begin{figure}[t]
\centering
\includegraphics[width=\textwidth]{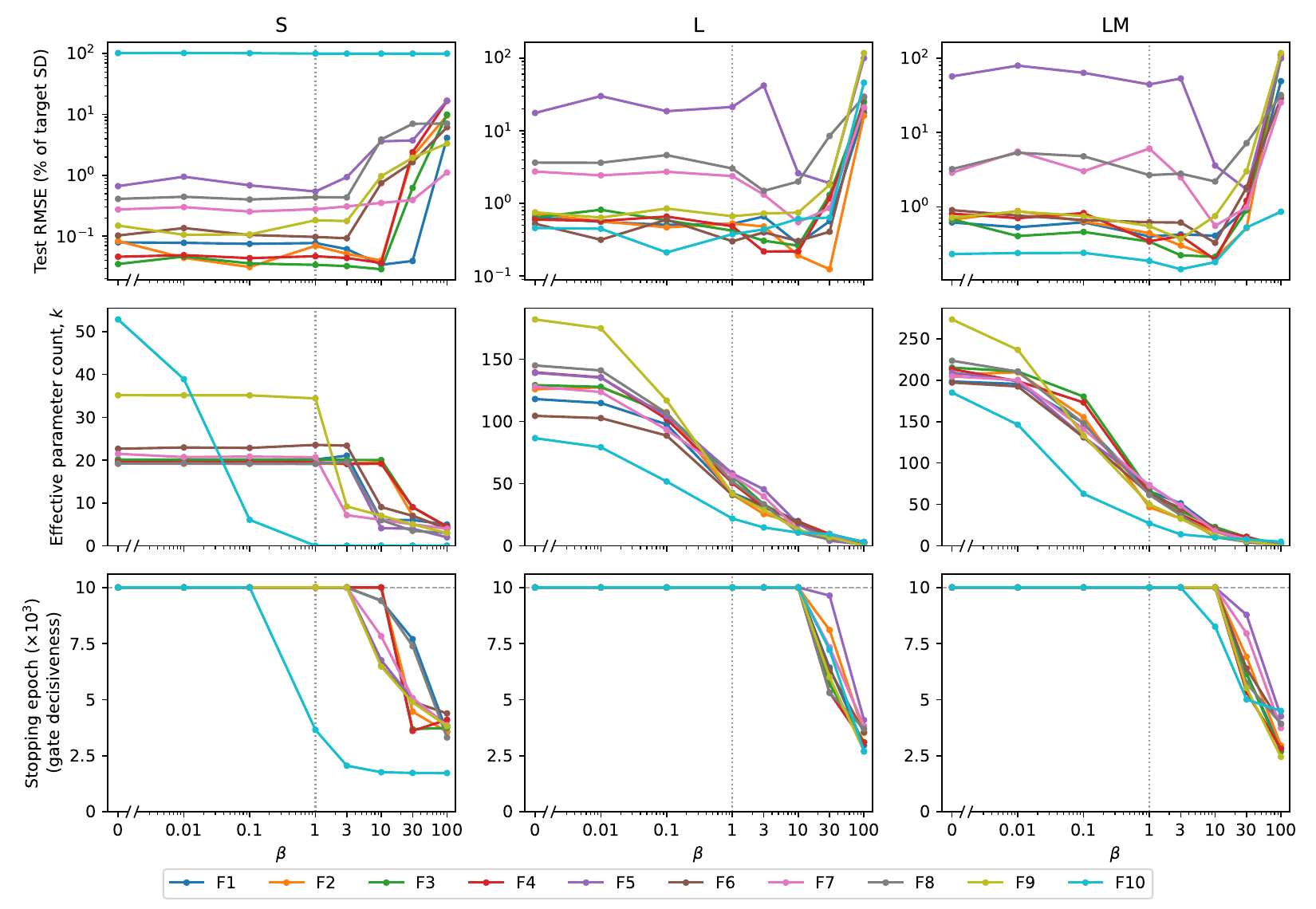}
\caption{Accuracy, model size, and stopping epoch versus regularization strength on Nguyen benchmarks. Each curve shows one problem's median over ten initialization seeds; IQRs are omitted for readability. Test RMSE is expressed as a percentage of the test-target standard deviation, calculated per fit before aggregation; $k$ follows Table~\ref{tab:nguyen-full}. The bottom row shows median stopping epochs under the gate-decisiveness rule, including runs reaching the 10,000-epoch cap (dashed line). The dotted line marks $\beta=1$. Positive $\beta$ values are log spaced, with zero shown separately; the RMSE axis is logarithmic and the other axes linear.}
\label{fig:nguyen-beta-sensitivity}
\end{figure}

Figure~\ref{fig:nguyen-beta-sensitivity} reports a sweep over $\beta\in\{0,0.01,0.1,1,3,10,30,100\}$ for all ten Nguyen problems and architectures S, L and LM, retaining all 2,400 fits through $\beta=100$. Unlike the real-world sweep, each problem uses a fixed training/test sample with ten initialization seeds per configuration. Except for the small F10 models at $\beta=1$, gate-based early stopping occurs only at $\beta\geq3$ and becomes widespread at $\beta=30$ and $100$ (bottom row).

\subsection{Behavior under library misspecification}
\label{subsec:misspecification}
We examine how S2KAN approximates targets whose generating primitives are absent from the symbolic library. The Nguyen benchmarks illustrate approximation using the remaining dictionary, while a controlled sinc ablation examines recovery through dense components.

The Nguyen benchmark (Table~\ref{tab:nguyen-full}) provides several examples of misspecification: F7 requires $\log(\cdot)$ and F8 requires $\sqrt{\cdot}$, neither of which appear in the symbolic library (see \methods{}). 
Despite this, S2KAN achieved median $R^2 > 0.99$ on both problems for every architecture. Accuracy was nonetheless lower than for baseline KAN: on the two larger architectures, the median S2KAN RMSE exceeds 1\% of the target standard deviation for both problems, whereas KAN stays within it (Table~\ref{tab:nguyen-full}).
In fact, S2KAN retains high symbolic fidelity on both problems: on the larger architectures, $J=2$, $\epsilon=0.01$ fidelity is 78--100\% for S2KAN versus 6--17\% for KAN, indicating that the available library terms serve as effective approximations even without the ground-truth functions.
The no-dense control also approximates F7 and F8, achieving lower median RMSE than full S2KAN for architecture L (Table~\ref{tab:nguyen-eql-control}). Thus, these examples demonstrate approximation using available symbolic and finite-basis terms despite missing generating primitives, rather than isolating a benefit from dense fallback.

Given the tradeoff observed in these Nguyen problems, we next test whether the dense branch can recover a symbolic component deliberately omitted from the training dictionary.
We therefore repeated the sinc experiment (Fig.~\ref{fig:sincx}) with the reciprocal removed from the symbolic library, $\mathcal{S}=\{x\}$, for five seeds; all other settings, including $\beta=1$, match the figure (see \methods{}).
Without access to this term, S2KAN cannot symbolically represent the factor $1/x$ in $\mathrm{sinc}(x) = \sin(x) \cdot (1/x)$.
All five ablated models still fit the data (test MSE $< 10^{-5}$; median $2.1\times10^{-6}$, versus $2.9\times10^{-8}$ for baseline KAN).
In three of the five, one activation is a pure $\sin(x)$ and the other falls back to a spline (in one seed with a small additional Fourier term).
That activation is numerically the missing primitive: a single affinely gauged $1/x$ fits it with $R^2 > 0.9999$, whereas the best alternative primitive we tested, $1/x^2$, reaches only $R^2 \approx 0.998$.
The two remaining ablated S2KAN fits retain more complex decompositions, with splines on both activations.
The spline-only baseline KAN, trained with the same data, architecture, and training schedule, never learns such a decomposition: its five fits are accurate (test MSE $\le 1.2\times10^{-6}$), but none of their ten spline activations is well approximated by a single gauged $\sin(x)$ or $1/x$ ($R^2 \le 0.89$ in every case).
The same three seeds keep this sine-times-reciprocal structure at $\beta = 2$, $10$ and $30$, but at $\beta = 30$ the other two close both spline branches and fail ($R^2 \approx 0.31$).
In this controlled product architecture, with sine terms supplied by the Fourier basis, the dense branch recovers the omitted reciprocal over the tested interval $[1,15]$ in three of five seeds. This demonstrates recovery of a missing symbolic component when part of the compositional structure is supplied; recovery is not guaranteed, and strong regularization can suppress the fallback.

\subsection{Sensitivity of symbolic fidelity to an amplitude constraint}
\label{sec:sf-conditioning-control}

All reported SF values use unconstrained symbolic approximation. This sensitivity analysis covers the main results, the Nguyen no-dense control, and the real-world dense-only control. Here we test whether the conclusions depend on allowing large, mutually cancelling fitted terms. For an edge response $y_e$ approximated by $d+\sum_{m=1}^{j}a_m p_m(b_m x+c_m)$, the constrained extractor additionally requires
\begin{equation}
 \operatorname{sd}\!\left[a_m p_m(b_m x+c_m)\right]\leq C\operatorname{sd}(y_e),\qquad C=3,
 \label{eq:sf-amplitude-control}
\end{equation}
for each fitted term, with standard deviations evaluated on the same empirical input sample used for SF. The additive offset is unrestricted. This is a post-hoc extraction control: model weights, predictions, functional-activity criteria, primitive dictionaries, input samples, and the six combinations of $J\in\{1,2,3\}$ and $\epsilon\in\{0.01,0.1\}$ are held fixed. The one-term search is unchanged. Multi-term fits use a penalty for violating Eq.~\eqref{eq:sf-amplitude-control}, followed by further optimization and rejection of candidates exceeding the bound by more than a relative tolerance of $10^{-4}$. Since the constraint also changes the nonlinear search path, we report the actual score from each search; occasional increases in constrained SF are retained.

For the main results and the Nguyen no-dense control, the inventory contains 1,020 distinct models contributing numerical SF results: 600 Nguyen models, 240 validation-selected real-world models, 80 dynamical-system models, and 100 additional no-dense Nguyen models. The full S2KAN models reused in the no-dense comparison are counted only once. We also evaluate the two illustrative sinc checkpoints in Fig.~\ref{fig:sincx}. Each pair uses the same archived checkpoint and extractor settings. Recent score records identify checkpoints by hash; historical Nguyen S2KAN records are matched through their checkpoint directories and verified against their functional-activity sets, edge variances, and one-term fits. All six SF scores are independently reconstructed from the saved edge fits.

\begin{figure}%
\centering
\includegraphics[width=0.8\textwidth]{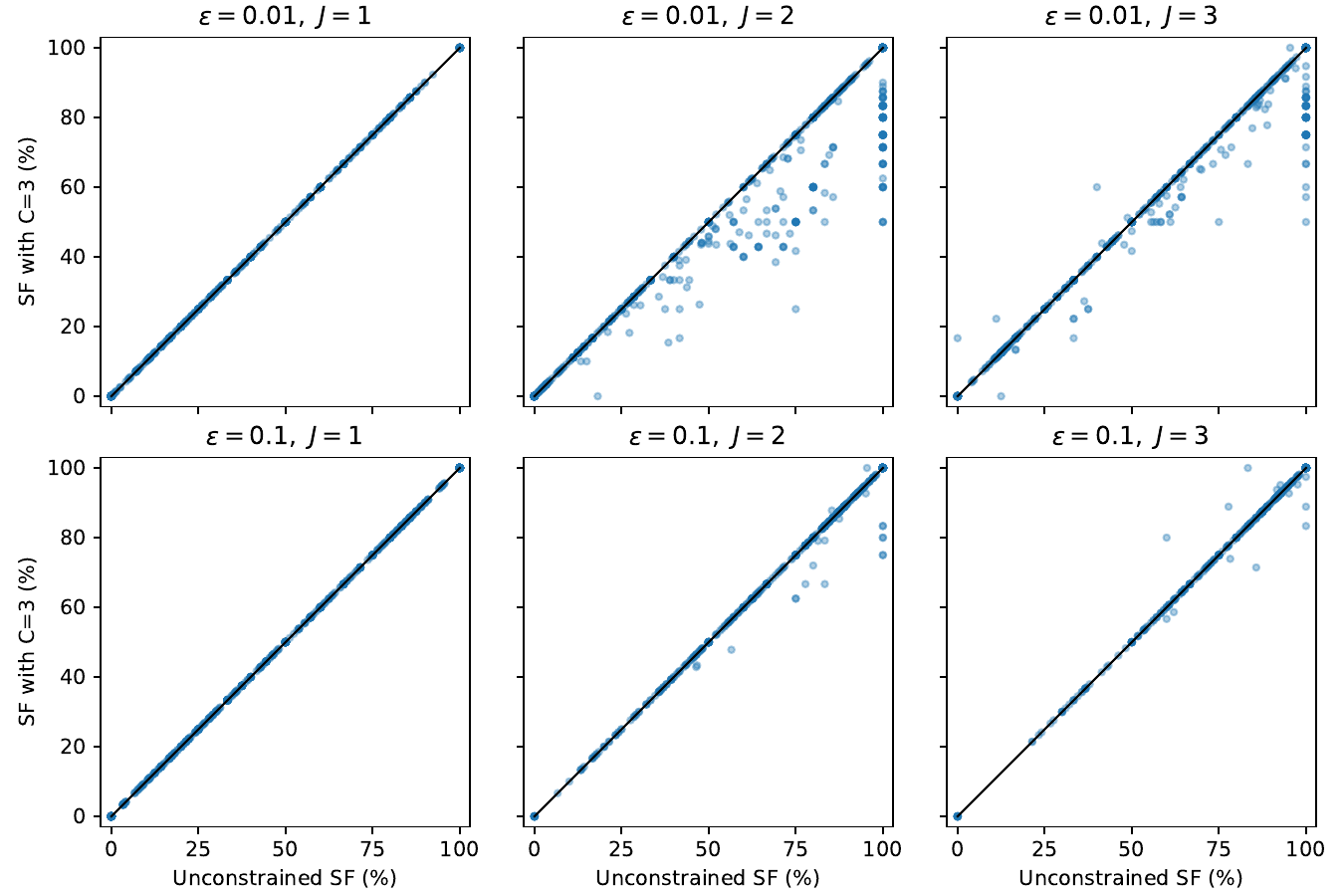}
\caption{Unconstrained versus amplitude-constrained ($C=3$) symbolic fidelity for the 1,022 models in the main-results and Nguyen no-dense inventory, including both sinc illustrations. Each panel contains one point per model; coincident points overlap. The line is equality. The $J=1$ search is unchanged by construction. Occasional points above the line reflect differences in the nonlinear search path rather than an improvement in the globally optimal constrained approximation.}
\label{fig:sf-conditioning-control}
\end{figure}

All 1,022 pairs passed the matching checks. Across the six settings, 5,804 of 6,132 scores (94.7\%) are identical; restricting the comparison to $J=2,3$, where the constraint can affect the search, 3,760 of 4,088 scores (92.0\%) are identical. Figure~\ref{fig:sf-conditioning-control} shows all pairs, and Table~\ref{tab:sf-conditioning-control} summarizes the changes. Both illustrated sinc models have identical scores at all six settings. Models with empty functional-activity sets retain the same SF convention in both versions.

For the additional 120 real-world dense-only models, 712 of 720 scores (98.9\%) are unchanged. The largest change in a dataset median is 3.33 percentage points. Full S2KAN retains higher median SF than dense-only S2KAN in all 24 comparisons with or without the constraint.

The main comparative findings persist (Table~\ref{tab:sf-conditioning-comparisons}). For Nguyen, unconstrained SF favors S2KAN in 114 of 180 median comparisons, with 66 ties; $C=3$ gives 113 wins and 67 ties, with no losses in either case. S2KAN retains higher median SF than baseline KAN in all 24 real-world comparisons. The dynamical-system comparison gives 28 wins, six ties and two losses with or without the amplitude constraint. The full-versus-no-dense comparison remains mixed: 11 wins, 39 ties and ten losses without the bound, versus 12 wins, 36 ties and 12 losses with it. These are descriptive counts across correlated settings, not independent statistical tests.

The constraint does affect some numerical SF estimates appreciably. The largest change in a Nguyen table median is 28.6 percentage points (F4, architecture LM, $\epsilon=0.01$, $J=2$); the largest real-world change is 17.9 points for power-plant S2KAN at the same setting, from 64.3\% to 46.4\%. The largest changes in configuration medians are 2.27 points for dynamical systems and 27.5 points for the no-dense control. Thus the broad comparisons are robust to this amplitude constraint, while individual fidelity values need not be invariant. Unconstrained SF measures approximation by a short symbolic expression; it does not itself guarantee well-conditioned coefficients.

\begin{table}[t]%
\centering\singlespacing{}
\caption{Amplitude-constraint sensitivity of SF. Changes are constrained minus unconstrained SF, in percentage points. ``Equal scores'' counts identical values across all six settings for the paired models; settings are correlated. ``Largest median shift'' compares medians within the same dataset/problem, architecture, method and regularization group at each setting, matching the main tables' aggregation. All expected fits are included in each group. The sinc illustration contains one checkpoint per method.}
\label{tab:sf-conditioning-control}
\begingroup\small\setlength{\tabcolsep}{4pt}
\begin{tabular}{@{}lrrrrr@{}}
\toprule
Cohort & Paired / total & Equal scores & Median $|\Delta|$ & Largest $|\Delta|$ & Largest median shift \\
\midrule
Dynamical systems & 80/80 & 440/480 & 0.00 & 5.88 & 2.27 \\
Nguyen & 600/600 & 3472/3600 & 0.00 & 50.00 & 28.57 \\
No-dense control & 100/100 & 547/600 & 0.00 & 50.00 & 27.50 \\
Real world & 240/240 & 1333/1440 & 0.00 & 33.33 & 17.88 \\
Sinc illustration & 2/2 & 12/12 & 0.00 & 0.00 & 0.00 \\
\bottomrule
\end{tabular}
 \endgroup
\end{table}

\begin{table}[t]%
\centering\singlespacing{}
\caption{Method comparisons using median SF at each of the six settings. Entries are wins/ties/losses for S2KAN versus baseline KAN, except the final row, which compares full S2KAN with its no-dense control. Nguyen comparisons span ten problems and three architectures; real-world comparisons span four datasets; dynamical-system comparisons span six S2KAN configurations, each compared with its system's baseline. Each median uses all fits in the corresponding paper table.}
\label{tab:sf-conditioning-comparisons}
\begingroup\small\setlength{\tabcolsep}{6pt}
\begin{tabular}{@{}lrr@{}}
\toprule
Comparison & Unconstrained & $C=3$ \\
\midrule
Nguyen & 114/66/0 & 113/67/0 \\
Real world & 24/0/0 & 24/0/0 \\
Dynamical systems & 28/6/2 & 28/6/2 \\
Full vs no dense & 11/39/10 & 12/36/12 \\
\bottomrule
\end{tabular}
 \endgroup
\end{table}

\section*{Statements and Declarations}

\subsection*{Funding}
No funding was received for this work.

\subsection*{Competing interests}
The authors declare no competing interests.

\subsection*{Data availability}
The real-world datasets used in this study are publicly available from the UCI Machine Learning Repository (\url{https://archive.ics.uci.edu/}). Dataset identifiers and descriptions are provided in \methods{}.

\subsection*{Code availability}
Code supporting this study will be made publicly available upon acceptance. The repository URL and version identifier will be included in the final manuscript.

\subsection*{Author contributions}
J.~Bagrow led the research and manuscript preparation, with contributions from J.~Bongard.

\bibliographystyle{paper-author-year}
\bibliography{references}

\end{document}